\documentclass[]{xiaomiev}
\microtypesetup{expansion=false} % TinyTeX/pdfTeX cannot expand the bundled Xiaomi font.
\usepackage[page,header]{appendix}
\usepackage{booktabs}
\usepackage{multirow}
\usepackage{graphicx}
\usepackage{tabularx}
\usepackage{hyperref}
\usepackage{amsmath}
\usepackage{amssymb}
\usepackage{makecell} 
\usepackage[colorinlistoftodos]{todonotes}
\usepackage{float}
\usepackage{enumitem}
\usepackage{listings}
\usepackage[ruled,linesnumbered]{algorithm2e}
\usepackage{tikz}
\usepackage{pgfplots}
\usepackage[table]{xcolor}
\pgfplotsset{compat=1.18}
\usetikzlibrary{arrows.meta, positioning, fit, backgrounds, calc, patterns}
\setlist[itemize]{leftmargin=15pt}
\definecolor{hxOrange}{HTML}{ED722E}
\definecolor{hxNavy}{HTML}{1B262C}
\definecolor{hxBlue}{HTML}{0F4C75}
\definecolor{hxTeal}{HTML}{3282B8}
\definecolor{hxSky}{HTML}{BBE1FA}
\definecolor{hxKey}{HTML}{0F4C75}
\definecolor{hxStr}{HTML}{8A5A00}
\definecolor{hxComment}{HTML}{6B7B8C}
\lstdefinelanguage{yaml}{
  sensitive=true,
  morecomment=[l]{\#},
  morestring=[b]",
  morestring=[b]',
  commentstyle=\color{hxComment}\itshape,
  stringstyle=\color{hxStr},
  keywords={true,false,null},
  keywordstyle=\color{hxTeal}\bfseries,
  moredelim=**[il][\color{hxKey}\bfseries]{:},
  moredelim=[l][\color{hxTeal}]{-\ },
}
\newtcblisting{promptfile}[2][]{%
  breakable, listing only,
  colback=hxSky!10, colframe=hxNavy, colbacktitle=hxNavy, coltitle=white,
  boxrule=0.6pt, left=4pt, right=4pt, top=3pt, bottom=3pt,
  title={\scriptsize\textbf{\texttt{#2}}}, #1}
\newtcblisting{manifestfile}[2][]{%
  breakable, listing only,
  listing options={language=yaml},
  colback=hxTeal!8, colframe=hxBlue, colbacktitle=hxBlue, coltitle=white,
  boxrule=0.6pt, left=4pt, right=4pt, top=3pt, bottom=3pt,
  title={\scriptsize\textbf{\texttt{#2}}}, #1}

\newtcolorbox{manifestcard}[1]{%
  colback=hxTeal!8, colframe=hxBlue, colbacktitle=hxBlue, coltitle=white,
  fonttitle=\scriptsize\bfseries, fontupper=\footnotesize,
  boxrule=0.6pt, left=5pt, right=5pt, top=4pt, bottom=5pt,
  before upper={\setlength{\parindent}{0pt}\raggedright},
  title={#1}}
\newtcolorbox{seedbox}[2][]{%
  enhanced, breakable,
  colback=hxSky!5, colframe=hxBlue!65, colbacktitle=hxSky!18, coltitle=hxBlue,
  fonttitle=\small\bfseries, fontupper=\small,
  boxrule=0.35pt, leftrule=1.6pt, arc=1mm,
  left=5pt, right=5pt, top=4pt, bottom=4pt,
  before skip=6pt, after skip=6pt,
  before upper={\setlength{\parindent}{0pt}\raggedright},
  title={#2}, #1}

\newlength{\ataglancegap}
\newcommand{\ataglance}[4]{%
\begin{seedbox}[breakable=false]{At a glance}
{\setlength{\parskip}{0pt}%
\setlength{\tabcolsep}{0pt}%
\renewcommand{\arraystretch}{1.03}%
\begin{tabularx}{\linewidth}{@{}>{\raggedright\arraybackslash\bfseries\color{hxBlue}}l@{\hspace{\ataglancegap}}>{\raggedright\arraybackslash}X@{}}%
Role & #1 \\[1.5pt]
System role & #2 \\[1.5pt]
Contract & #3 \\[1.5pt]
Evidence & #4
\end{tabularx}}
\end{seedbox}}
\newtcolorbox{seednote}[2][]{%
  enhanced, breakable,
  colback=hxOrange!5, colframe=hxOrange!70, colbacktitle=hxOrange!16, coltitle=hxStr,
  fonttitle=\small\bfseries, fontupper=\small,
  boxrule=0.35pt, leftrule=1.6pt, arc=1mm,
  left=5pt, right=5pt, top=4pt, bottom=4pt,
  before skip=6pt, after skip=6pt,
  before upper={\setlength{\parindent}{0pt}\raggedright},
  title={#2}, #1}

\RequirePackage{xspace}
\makeatletter
\DeclareRobustCommand\onedot{\futurelet\@let@token\@onedot}
\def\@onedot{\ifx\@let@token.\else.\null\fi\xspace}

\makeatother

\newcommand{\MiMemory}{Mi-Memory\xspace}
\newcommand{\MemStack}{MemStack\xspace}
\newcommand{\MemSense}{MemSense\xspace}
\newcommand{\LiteMem}{LiteMem\xspace}
\newcommand{\EMEND}{E\textsuperscript{2}MEND\xspace}
\newcommand{\DACCI}{D\textsuperscript{2}ACCI\xspace}
\newcommand{\MemFuse}{MemFuse\xspace}

\newcommand{\compacttable}{\footnotesize\setlength{\tabcolsep}{5pt}\renewcommand{\arraystretch}{1.05}}
\title{{\fontsize{14.5}{18}\selectfont \textcolor{hxOrange}{Mi-Memory}: A Lifecycle Memory Framework for Personal AI}}

\author{\textbf{Darwin Agent Team}\\
{\small See \hyperref[sec:contributions-acknowledgments]{Contributions and Acknowledgments} section for a full author list.}}

\renewcommand{\abstractinfont}{\fontsize{9}{10.5}\selectfont}

\abstract{
Personal AI is moving beyond chat-only interaction toward continuous services that span phones, cars, homes, wearables, cameras, and tools. In this setting, memory cannot remain a cache of prior conversations. It should serve as a continuity and governance substrate: preserving durable user state, grounding answers in multimodal and device evidence, supporting correction and forgetting, bounding policy evolution, and remaining deployable under latency, cost, privacy, and edge--cloud constraints. This technical report presents \textcolor{hxOrange}{\textbf{Mi-Memory}}, a lifecycle memory framework for Personal AI organized around four roles: \textbf{Structure}, \textbf{Expansion}, \textbf{Evolution}, and \textbf{Deployment}. A shared audit contract links these roles through four recurring artifact families: typed evidence payloads preserve source identity and provenance, diagnostic traces localize evidence loss across the serving pipeline, strategy artifacts make memory-policy changes explicit, and gate/rollback records bound accepted evolution. \MiMemory instantiates the roles through \MemStack, \MemSense/\MemFuse, \DACCI/\EMEND, and \LiteMem. In controlled-reference Structure evaluations, \MemStack reaches 93.59\%, 57.24\%, and 87.47\% on LoCoMo, PersonaMem-V2, and LongMemEval, respectively; other tracks report module-level, preliminary/internal, transfer-feasibility, or design-only evidence with explicit boundaries. \MiMemory is a step toward auditable, evidence-gated, and deployment-aware memory systems for Personal AI. \textcolor{hxOrange}{Project homepage: \href{https://darwin-agent.github.io/Mi-Memory/}{https://darwin-agent.github.io/Mi-Memory/}.}
  
}

\begin{document}
\maketitle

\enlargethispage{22mm}
\vspace*{-8mm}
\begin{center}
\includegraphics[width=0.89\textwidth]{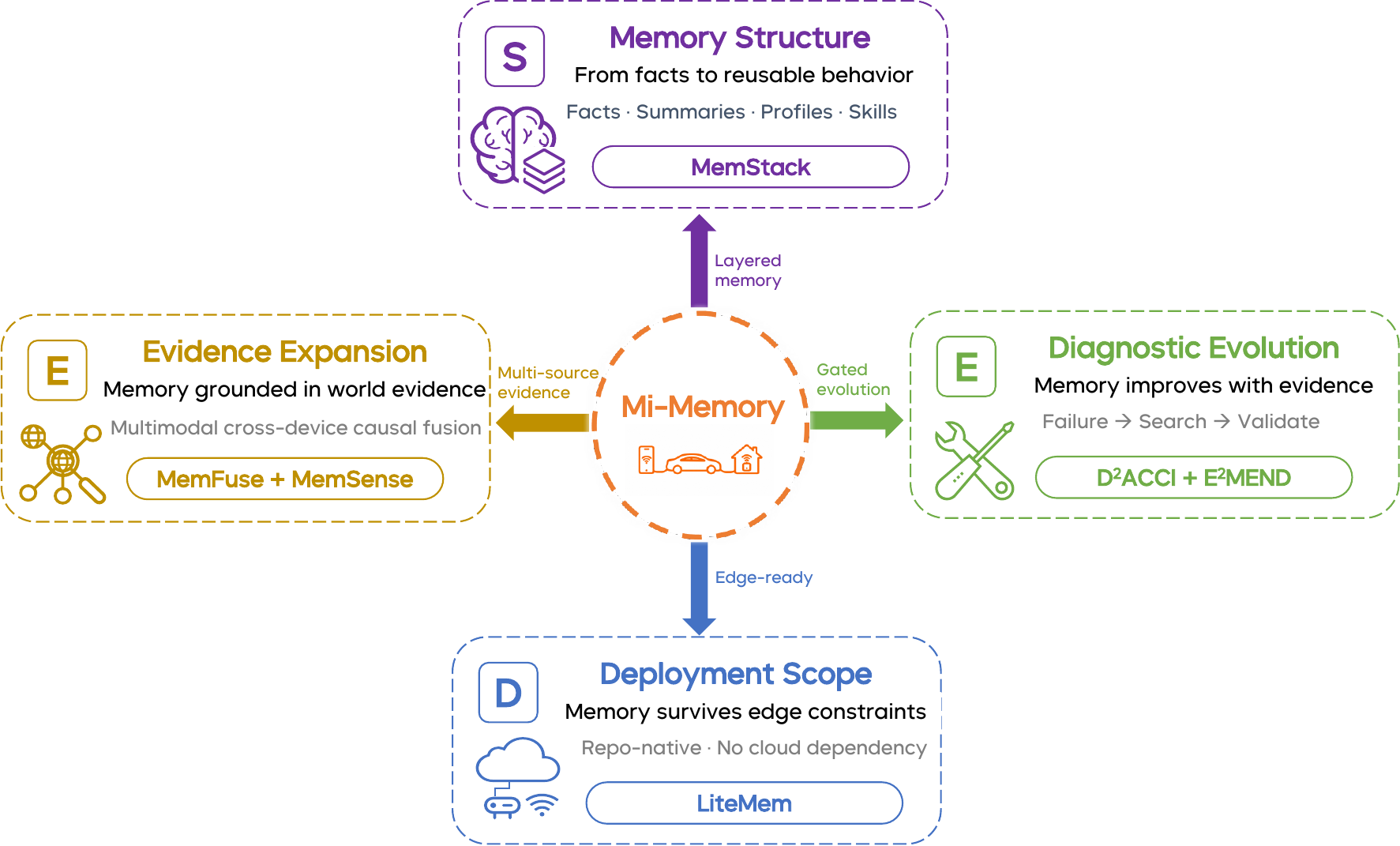}
{\captionsetup{hypcap=false}
\captionof{figure}{\MiMemory lifecycle.}
%The four SEED roles share one audit contract: typed evidence preserves source identity, diagnostic traces localize evidence loss, strategy artifacts constrain evolution, and gate/rollback records make updates reviewable.}
\label{fig:overview}}
\end{center}

\clearpage
% Keep the table of contents scannable: list sections and subsections only,
% not the many \paragraph-level headings used inside subsections.
\setcounter{tocdepth}{2}
\tableofcontents
\setcounter{tocdepth}{2}
\clearpage
%% ═══════════════════════════════════════════════════════════════════════════════
\section{Introduction}
\label{sec:introduction}

Personal AI is moving from a chat-only interface toward continuous, cross-device services. Its memory layer must maintain user state over time, connect evidence across phones, cars, homes, wearables, cameras, and tools, and revise or forget records as user intent and context change. This layer functions as both an architectural substrate and a governance surface: it determines what the assistant can know, explain, revise, and reuse under stated controls while making visible which observations are active, which sources are linked, and which updates are permitted.

These requirements make the systems problem broader than long-term recall. First, there is a \textbf{state gap}: durable facts, summaries, profiles, sessions, and procedures are often maintained at incompatible granularities. Second, there is a \textbf{source gap}: dialogue, images, tool traces, and device events are indexed by different subsystems, so cross-device evidence can disappear before it reaches generation. Third, there is a \textbf{governance gap}: final answer accuracy rarely reveals whether a failure began in ingestion, fusion, retrieval, filtering, context assembly, or generation, and unchecked policy updates can silently regress hidden slices. Finally, there is a \textbf{deployment gap}: memory design must account for cost, latency, privacy, and cloud/edge/local substrates instead of assuming a single vector-store runtime.

These gaps share a common cause: memory is treated as a static store, not as a \emph{lifecycle}. In practice, assistant-relevant evidence passes through stage transitions---observation, structuring, fusion, retrieval, assembly, response, correction, and deployment---and each transition can lose, distort, or silently regress information. A camera frame may be captured but never linked to its causal context; a fused episode may exist in storage but fail retrieval filtering; a promising policy change may be deployed but silently degrade a hidden question category. The lifecycle perspective therefore shifts the target from optimizing one isolated memory store to defining an end-to-end contract that makes evidence movement traceable and stage transitions auditable.

\textbf{\MiMemory} operationalizes this perspective through four lifecycle roles. \textbf{Structure} organizes admitted observations at the appropriate granularity and keeps them retrievable across time. \textbf{Expansion} admits new evidence sources---images, device events, cross-device links---without losing identity or provenance. \textbf{Evolution} ties each memory-system change to a falsifiable hypothesis, fixed evaluation conditions, and reversible update records. \textbf{Deployment} preserves audit obligations when memory moves across serving substrates---from cloud to edge to local files. Later sections ground these roles in concrete modules and explicit evidence levels.

For example, in a multi-device consumer ecosystem such as Human-Car-Home, a restaurant destination selected on the phone may influence car navigation; a wearable sleep signal may contextualize a family routine; and an in-car reminder before a child's training session may depend on a home-camera observation, a calendar event, route timing, and prior user corrections, rather than any single chat turn. The lifecycle framing matters because this cross-device evidence must survive every stage transition from observation to response, and failures can surface at any point along the way.

\subsection{Research Questions}

\MiMemory{} is organized around four research questions that turn the gaps above into the report structure. Table~\ref{tab:rq-map} is a roadmap, not a leaderboard: each row names the lifecycle role being tested, the artifact family that carries the audit obligation, the section where the mechanism is instantiated, and the evidence boundary for interpreting the claim. The table also separates two questions that are often conflated in memory-system papers: whether the runtime preserves task-relevant state, and whether later lifecycle functions---evidence admission, policy evolution, and deployment transfer---preserve the same traceability obligations.

\begin{table}[htbp]
\centering
\caption{Research-question map.}
\label{tab:rq-map}
\compacttable
\begin{tabularx}{\textwidth}{p{0.32\textwidth}p{0.10\textwidth}p{0.28\textwidth}p{0.07\textwidth}X}
\toprule
\textbf{Research question} & \textbf{Lifecycle} & \textbf{Main artifact} & \textbf{Section} & \textbf{Evidence level} \\
\midrule
\textbf{RQ1:} How can long-term assistant memory preserve task-relevant user state without collapsing into unstructured chunks? & Structure & Layered memory records, retrieval traces, adapter contracts & Sec.~\ref{sec:memorystack} & Controlled reference \\
\textbf{RQ2:} How can memory incorporate evidence from cameras, wearables, cars, homes, and other devices without losing provenance? & Expansion & Typed source observations, MemoryPacks, FusionSession graphs & Sec.~\ref{sec:source} & Module-level and preliminary/internal \\
\textbf{RQ3:} How can memory systems improve without silent regressions or hidden framework edits? & Evolution & \DACCI traces, \EMEND strategy artifacts, gates, rollback records & Sec.~\ref{sec:emend} & Controlled offline benchmark / descriptive \\
\textbf{RQ4:} How can the same memory contract transfer to privacy-sensitive or edge-oriented deployment substrates? & Deployment & Markdown/Git memory files, repository traces, file-tool retrieval & Sec.~\ref{sec:light} & Transfer-feasibility \\
\bottomrule
\end{tabularx}
\end{table}

Together, the rows emphasize lifecycle inspectability over any single end-to-end benchmark. RQ1 anchors the runtime substrate: the \textbf{Structure} role asks whether structured memory records, retrieval traces, and adapter contracts can support controlled memory QA (instantiated as \MemStack, Section~\ref{sec:memorystack}). RQ2 widens the evidence universe: the \textbf{Expansion} role tests whether visual and cross-device observations can enter the system as typed, provenance-preserving payloads (via \MemSense and \MemFuse, Section~\ref{sec:source}). RQ3 introduces governed change: the \textbf{Evolution} role asks whether diagnostic traces can become reversible strategy updates instead of silent prompt or retrieval edits (via \DACCI and \EMEND, Section~\ref{sec:emend}). RQ4 tests transfer: the \textbf{Deployment} role asks which parts of the same contract survive when the serving substrate becomes repository-native and lightweight (via \LiteMem, Section~\ref{sec:light}).

The central systems question is whether these roles preserve the shared artifacts needed for bounded, auditable iteration across the memory lifecycle. This organization motivates the later pairing of accuracy or qualitative examples with the artifact that makes each claim inspectable: typed payloads, diagnostic traces, strategy artifacts, gate decisions, rollback records, or repository history.

\subsection{Running Example: Human-Car-Home Training Handoff}
\label{sec:running-example}

The following Human-Car-Home trace makes the lifecycle concrete. On Tuesday afternoon, a parent is driving Ethan to basketball training. The car has left the residential community; the phone calendar records a 6:00 p.m. training session; the in-car navigation system estimates whether a short return home would still allow on-time arrival; and a home camera frame suggests that Ethan's training bag may still be near the entrance. Before entering the expressway, the parent asks the in-car assistant: ``Is there anything we should check before heading to training?''

A dialogue-only memory may have access only to the current question and return a general checklist. A lifecycle-aware memory system should instead preserve the evidence chain that makes a specific answer possible:
\begin{itemize}[nosep]
    \item \textbf{Observation}: the camera frame, calendar event, and vehicle route state are each captured as typed evidence with device identity and timestamp.
    \item \textbf{Structuring}: each observation is stored at an appropriate granularity---an atomic event, a session summary, or a stable profile entry---so it can be distinguished from other evidence over time.
    \item \textbf{Fusion}: related observations are linked into a coherent episode (training bag + scheduled practice + return-time estimate) despite arriving from different devices without shared keywords.
    \item \textbf{Retrieval and assembly}: when the in-car query arrives, the system recovers the fused episode and packs it into a bounded answer context alongside relevant prior corrections and family preferences.
    \item \textbf{Response}: the answer can cite the evidence it used---the bag observation, the time estimate, the correction about spare equipment---making its reasoning inspectable.
    \item \textbf{Diagnosis}: if the answer is wrong, the system can localize whether the failure came from missed observation, broken fusion, retrieval filtering, or generation error.
    \item \textbf{Governed update}: if the parent corrects the assistant (``Ethan keeps spare shoes at school; only remind me when the jersey is missing''), the correction enters memory through a gated policy change, not a silent edit.
    \item \textbf{Deployment transfer}: the same episode can be stored as local repository artifacts (Markdown files with Git provenance) when privacy or offline constraints require it, preserving inspectability under a lighter retrieval policy.
\end{itemize}

This is a lifecycle trace, not an end-to-end product claim. The requirement is to preserve evidence identity, provenance, temporal order, device constraints, and causal links across every stage transition, while keeping each transition auditable. Section~\ref{sec:overview} maps these stages to concrete modules; later technical sections (Sections~\ref{sec:memorystack}--\ref{sec:light}) describe how each stage is implemented and where evidence can still be lost.

\paragraph{Contributions.}
The report's primary contribution is the lifecycle audit contract itself. It unifies algorithmic, systems, and governance aspects of assistant memory by placing multi-granularity state organization, hybrid retrieval composition, evidence admission, and bounded policy evolution under one diagnostic framework. The concrete contributions are:
\begin{itemize}
    \item \textbf{Lifecycle formulation.} It formulates Personal AI memory as a traceable lifecycle problem and makes typed evidence, stage-local diagnostics, strategy artifacts, and gate/rollback records the organizing contract.
    \item \textbf{Structured runtime.} It presents \MemStack as a typed, multi-granularity reference runtime with hybrid retrieval, bounded context assembly, optional procedural hooks, and controlled-reference memory-QA evidence.
    \item \textbf{Evidence expansion.} It studies Evidence Admission through \MemSense and \MemFuse, demonstrating that visual and cross-device observations can become provenance-preserving payloads while marking \MemSense results as module-level evidence and \MemFuse results as preliminary/internal evidence.
    \item \textbf{Governed evolution.} It connects \DACCI and \EMEND to the same audit contract, demonstrating that diagnostic traces, schema-bounded strategy search, gate decisions, and rollback records can make memory-policy improvement explicit and reversible.
    \item \textbf{Deployment transfer.} It presents \LiteMem as a repository-native transfer path for selected \MiMemory contract elements, demonstrating that Markdown/Git artifacts and file-tool retrieval can preserve provenance, editability, and auditability under lightweight deployment constraints.
\end{itemize}
Section~\ref{sec:evaluation} centralizes the evidence-level taxonomy, statistical qualifiers, and follow-up validation plan.

%% ═══════════════════════════════════════════════════════════════════════════════
\section{Related Work}
\label{sec:related}

Assistant memory research spans several lines of work that are often studied separately. The systems closest to \MiMemory cover the lifecycle's complementary slices: durable state, source grounding, policy evolution, and deployable substrates.

\paragraph{General-purpose assistant memory systems.}
Agent memory is often framed as long-term dialogue recall, but Personal AI requires a broader substrate: durable state, multimodal and device-grounded evidence, managed update and forgetting, and deployment under cost, latency, privacy, and edge--cloud constraints. Prior systems relevant to this broader substrate fall into several families. Fact- and profile-memory systems such as Mem0~\cite{chhikara2025mem0,mem0token2026} emphasize extraction, update, and retrieval. Hierarchical and agent-runtime systems organize profiles, episodes, summaries, memory-operation policies, and tool-facing memory operations~\cite{hmo2026,memmachine2026,xia2026memora}; OS-inspired systems such as MemGPT manage long-lived agent state through virtual context and memory tiers~\cite{packer2023memgpt}.

Managed assistant products such as Apple Intelligence's on-device context and Google Gemini's memory features indicate demand for persistent personalization, although their internal provenance, diagnostic, and evolution mechanisms are rarely public. Cross-scenario memory providers and newer benchmarks underscore the challenge of remaining reliable across heterogeneous tasks~\cite{chen2026agentic,longmemevalv2_2026}. The unresolved systems problem is how to connect these memory families through a shared audit contract: what evidence is created, how it crosses module boundaries, where it can be lost, how updates are governed, and which deployment substrate can serve it within stated constraints. \MiMemory treats this contract as the connective layer across memory stores, source modules, evolution controls, and deployment substrates.

\paragraph{Memory safety, privacy, and governance.}
Memory also expands the safety surface beyond standard prompt context: stored facts can become stale, sensitive, over-generalized, or retrieved in the wrong situation. Recent memory-security work studies lifecycle attacks and defenses~\cite{memsecurity2026}, trustworthy retrieval gates~\cite{memgate2026}, and stale or harmful memories that remain frequently accessed~\cite{stale2026,whentoforgot2026}. Industry systems increasingly emphasize user-visible control, local context, and editability, but public documentation rarely exposes the diagnostic pipeline behind those controls. \MiMemory focuses on the audit-enabling layer of memory governance: it makes provenance, update decisions, forget constraints, and strategy changes explicit, so product-level consent mechanisms, access-control policies, and privacy policies can operate on inspectable artifacts, not hidden retrieval state.

\paragraph{Memory evaluation and diagnostic iteration.}
Benchmarks such as LoCoMo~\cite{maharana2024locomo}, PersonaMem-V2~\cite{jiang2025personamem}, and LongMemEval~\cite{wu2024longmemeval} evaluate whether a memory system can answer questions over long histories. Newer representation and benchmark settings emphasize abstraction--specificity balance, forgetting-aware evaluation, and long-horizon agent behavior~\cite{xia2026memora,uddin2026memorafama,longmemevalv2_2026}. Multimodal memory benchmarks such as M$^3$Exam~\cite{m3exam2026} and MemLens~\cite{memlens2026} further test whether visual observations and cross-modal context persist across sessions, while arena-style evaluations such as EvoArena~\cite{evoarena2026} track evolution robustness in dynamic environments. These benchmarks are necessary, but accuracy alone rarely explains why an improvement happened or why a regression occurred. Evaluation-driven iteration work shows that seemingly improved prompts or pipeline changes can hurt hidden categories~\cite{mves2026}, and data-centric AI highlights the value of systematic evidence over ad hoc model changes~\cite{zha2023datacentric}. \DACCI builds on this direction by making each iteration falsifiable: a change must be tied to a hypothesis, per-question diagnostic traces, paired comparisons, and a category-level non-regression decision.

\paragraph{Self-evolving memory and agent optimization.}
Self-evolving memory systems aim to reduce manual tuning by searching architectures, prompts, or memory-operation policies. EvolveMem explores automated memory architecture search~\cite{evolvemem2026}, while Memory-R1 and Memento represent RL-style or case-based learning of memory-using agent policies~\cite{yan2026memoryr1,zhou2025memento}. MIA couples non-parametric trajectory memory with a parametric planner through test-time learning in deep research agents~\cite{qiao2026mia}. Broader agent memory work studies adaptive collaboration, dual-process cognitive memory, and memory skills for fine-grained update decisions~\cite{ama2026,dcpm2026,memskill2026}. These approaches motivate automated improvement, but broad end-to-end objectives can complicate production debugging because indexing, extraction, retrieval, and presentation may all change at once. \EMEND adopts a narrower and more auditable stance: it automates only schema-constrained strategy changes inside a locked framework, while preserving candidate traces, gate decisions, and rollback records for later inspection.

\paragraph{Beyond dialogue and deployable memory.}
Many memory benchmarks still center on dialogue or text histories, while deployed assistants increasingly observe visual scenes, device events, tool traces, and user routines. Multimodal memory agents with dual-layer hybrid memory~\cite{m2a2026} and lifelong multimodal memory~\cite{memverse2026} demonstrate that visual, auditory, and contextual observations can be organized as persistent agent knowledge. This direction echoes the memory-stream idea from generative-agent simulations, where observations, reflections, and plans persist beyond one prompt. Other work frames memory as externalized agent context or as a metabolic process of continual update and forgetting~\cite{externalization2026,memmetabolism2026}. \MiMemory incorporates these directions by treating non-dialogue inputs as typed source observations whose identity, time, device provenance, and causal links must survive into downstream memory and retrieval.

A second axis is where memory is stored. Work on direct corpus interaction, repository- or file-native agent memory, version-controlled reasoning traces, and lightweight memory-augmented generation indicates that assistant memory can reside in file-native and versioned substrates alongside vector/database runtimes and conversational transcripts~\cite{dci2026,byterover2026,gitofthoughts2026,lightmem2026,superlocalmemory2026}. These systems vary in scope: some change the retrieval interface, some externalize agent state into files or repositories, and others reduce memory-augmented generation cost. \LiteMem follows this deployment-oriented line by testing whether selected \MiMemory principles can transfer to Markdown/Git memory under lightweight local constraints, while optional procedural hooks remain a design-only interface for user-specific interaction procedures inside \MemStack.

\paragraph{System-level positioning.}
Table~\ref{tab:positioning-matrix} maps related work to its \emph{primary contribution} within the lifecycle framing. The column heading uses ``primary contribution'' because many systems span multiple concerns---Memora covers both state organization and evolution, EvolveMem bridges evolution and deployment---but each is listed under the slice it most advances. \MiMemory connects these slices under one audit contract across the Personal AI memory lifecycle.

\begin{table}[htbp]
\centering
\caption{Related-work positioning by primary lifecycle contribution.}
\label{tab:positioning-matrix}
\compacttable
\begin{tabularx}{\textwidth}{@{}p{0.25\textwidth}X p{0.22\textwidth}@{}}
\toprule
\textbf{Primary contribution} &
\textbf{Representative prior work} &
\textbf{\MiMemory module} \\
\midrule
State organization & Mem0 / MemGPT / HMO / MemMachine / Memora$^\ast$ & \MemStack \\
Evidence grounding & MemVerse / Mem-Gallery / M$^3$Exam / M2A / MemLens & \MemSense + \MemFuse \\
Policy evolution & EvolveMem$^\ast$ / Memory-R1 / MemSkill / SSGM & \DACCI / \EMEND \\
Deployment substrate & LightMem / Git-of-Thoughts / Git Context Controller & \LiteMem \\
\bottomrule
\end{tabularx}
\par\vspace{2pt}
{\footnotesize $^\ast$Also contributes to an adjacent slice (Memora: evolution; EvolveMem: deployment).}
\end{table}

% \paragraph{Positioning and gap.}
The remaining gap is therefore operational, not purely algorithmic: Personal AI memory needs a contract through which provenance, typed interfaces, update governance, and deployment budgets can survive module boundaries. The following sections instantiate that contract through \MemStack, Evidence Admission, \DACCI/\EMEND, and \LiteMem, while Section~\ref{sec:evaluation} specifies each claim's maturity and boundary.

%% ═══════════════════════════════════════════════════════════════════════════════
\section{Problem Formulation and System Overview}
\label{sec:overview}

\subsection{Formulation: Lifecycle Audit Contract}

To make the lifecycle view operational, this section names the objects that must remain traceable as evidence moves between stages. At turn $t$, let $q_t$ denote the user request, $O_t$ the typed source-observation pool available at serving time, $M_t$ the memory state before serving, $E_{q_t}$ the selected evidence, $C_{q_t}$ the bounded generation context, and $a_t$ the generated answer.

A general assistant memory system is modeled as two coupled processes: a serving runtime that produces the current answer and an improvement loop that changes memory state or policy across turns. The memory policy governs ingestion, retrieval, filtering, context construction, and update.

During serving, the runtime selects evidence $E_{q_t} \subseteq M_t \cup O_t$, assembles $C_{q_t}$, and generates an answer $a_t$ grounded in $E_{q_t}$. Across turns, the improvement loop writes the next memory state $M_{t+1}$ and may revise the memory policy only through governed evolution.

Unlike conventional RAG, $O_t$ is not a fixed text corpus. It may contain dialogue turns, device events, visual observations, user corrections, profile updates, and deployment-specific constraints. The memory policy therefore governs retrieval, admission, fusion, invalidation, and the substrate that carries evidence.

The audit contract treats answer accuracy as necessary but insufficient because a correct answer can still hide an unsupported memory write or policy change. It separates \emph{serving correctness} (whether a single answer is grounded) from \emph{update correctness} (whether a memory or policy change can be trusted). A valid memory system must satisfy three operational requirements:
\begin{enumerate}
    \item \textbf{Evidence preservation}: every memory-dependent answer should retain traceable links from output context back to source observations or stored memory items.
    \item \textbf{Stage-local diagnosis}: when an answer fails, the trace should identify whether the earliest loss occurred in ingestion, storage, retrieval, filtering, context packing, or generation.
    \item \textbf{Auditable evolution}: a policy change should be tied to a hypothesis, evaluated under a fixed harness, compared against a baseline, and either accepted or rejected with preserved artifacts.
\end{enumerate}

Operationally, an answer satisfies the audit contract only if $a_t$ can be traced back through the assembled context $C_{q_t}$, selected evidence $E_{q_t}$, and source or memory identifiers in $M_t \cup O_t$. A policy update satisfies the contract only if it carries a versioned strategy diff, a fixed harness, a gate decision, and a rollback point. With those records in place, violations become observable: missing source identity, unlocalized evidence loss, an unversioned policy change, or an accepted update without a restore record.

The formulation also separates three objects that are often conflated. A \emph{memory item} is stored runtime content; \emph{evidence} is the subset of source observations or memory items used to justify a specific answer or update; and a \emph{strategy} is an explicit configuration that changes how evidence is selected or assembled. The distinction matters because storing the correct memory item is not enough: retrieval, filtering, or context construction can still drop the evidence before generation.

Together, these distinctions define the audit contract used throughout the report. At each module boundary, the contract is carried by four artifact families: \textbf{typed evidence} preserves source identity and provenance; \textbf{diagnostic traces} record where selected evidence moved or disappeared; \textbf{strategy artifacts} make memory-policy changes explicit and versioned; and \textbf{gate/rollback records} capture whether those changes can be accepted.

A lifecycle role denotes an obligation in the audit contract, not a claim about empirical maturity. Structure preserves typed memory items; Evidence Admission preserves source identity across heterogeneous observations; Evolution makes policy changes falsifiable and reversible; and Deployment tests how much of the same contract survives under different serving constraints.

\begin{table}[htbp]
\centering
\caption{Audit artifacts.}
\label{tab:audit-contract}
\compacttable
\begin{tabularx}{\textwidth}{>{\raggedright\arraybackslash}p{0.21\textwidth}>{\raggedright\arraybackslash}p{0.29\textwidth}>{\raggedright\arraybackslash}p{0.20\textwidth}>{\raggedright\arraybackslash}X}
\toprule
\textbf{Artifact} & \textbf{What it preserves} & \textbf{Primary roles} & \textbf{Failure made visible} \\
\midrule
Typed evidence payload & Source id, time, device, confidence, and provenance & Structure; Evidence Admission & Missing, stale, or mis-linked evidence \\
Diagnostic trace & Stage-local movement from ingestion to answer context & Structure; Evolution & Earliest loss in retrieval, filtering, packing, or generation \\
Strategy artifact & Allowed memory-policy mutation and versioned parameters & Evolution & Hidden prompt or configuration drift \\
Gate/rollback record & Accept/reject decision, comparison baseline, and restore point & Evolution; Deployment & Silent regression or unsafe strategy adoption \\
\bottomrule
\end{tabularx}
\end{table}

Table~\ref{tab:audit-contract} turns the formulation into module responsibilities: \MemStack anchors typed state, \MemSense and \MemFuse admit heterogeneous source evidence, \DACCI and \EMEND govern strategy change, and \LiteMem tests substrate transfer.

Section~\ref{sec:evaluation} later separates \emph{evidence level} (harness maturity) from \emph{statistical strength} (repeated runs, confidence intervals, or paired tests). That separation keeps this overview focused on the audit contract while preserving explicit result boundaries for the empirical sections.

\paragraph{Assumptions, boundaries, and terms.}
\MiMemory focuses on memory-dependent assistant behavior within a trusted serving environment and explicit user-data governance. Its contribution here is auditable memory behavior: source observations, stored items, retrieved candidates, strategy changes, and rollback records remain visible to downstream policy and product controls. This boundary sets the evaluation scope: the report evaluates recall, grounding, bounded offline evolution, source expansion, and transfer feasibility, while production privacy enforcement, online abuse handling, and user-facing consent controls remain platform responsibilities that require additional validation. The detailed risk-gap summary is moved to Appendix~\ref{app:memorystack-extended} so the main formulation can stay focused on the lifecycle contract.

The terms used throughout the report are: a \emph{memory item} is a typed stored unit (L0 fact, L1 summary, L2 profile, optional procedural hook, or source payload); \emph{evidence} is the subset of items or source observations that justify a specific answer or update; a \emph{strategy artifact} is a declarative, versioned configuration that \EMEND can mutate; and a \emph{harness} is a fixed evaluation environment. Full definitions, including FusionSession, MemoryPack, diagnostic trace, and gate, are in Table~\ref{tab:terminology} (Appendix~\ref{app:memorystack-extended}).

The operational boundary separates the audited improvement process \DACCI from the automated strategy-search step \EMEND. \DACCI is a human-guided engineering methodology: its primary unit is a falsifiable human hypothesis plus evidence artifacts. \EMEND is an automated strategy-search engine inside that methodology: its primary unit is a candidate strategy artifact evaluated by a locked harness. Code, schema, benchmark, and acceptance-rule changes remain \DACCI decisions, not autonomous \EMEND acceptances.

\subsection{Framework: Lifecycle Architecture and Interfaces}

The framework is contract-first: each \MiMemory component is introduced by the lifecycle obligation it carries and by how it produces, consumes, or audits one of four shared artifacts: typed evidence payloads, diagnostic traces, strategy artifacts, and gate/rollback records. This contract separates the serving runtime, which handles source admission and memory processing, from the evolution governance plane, which diagnoses and updates those capabilities.

\begin{figure}[htbp]
    \centering
    \includegraphics[width=0.85\textwidth]{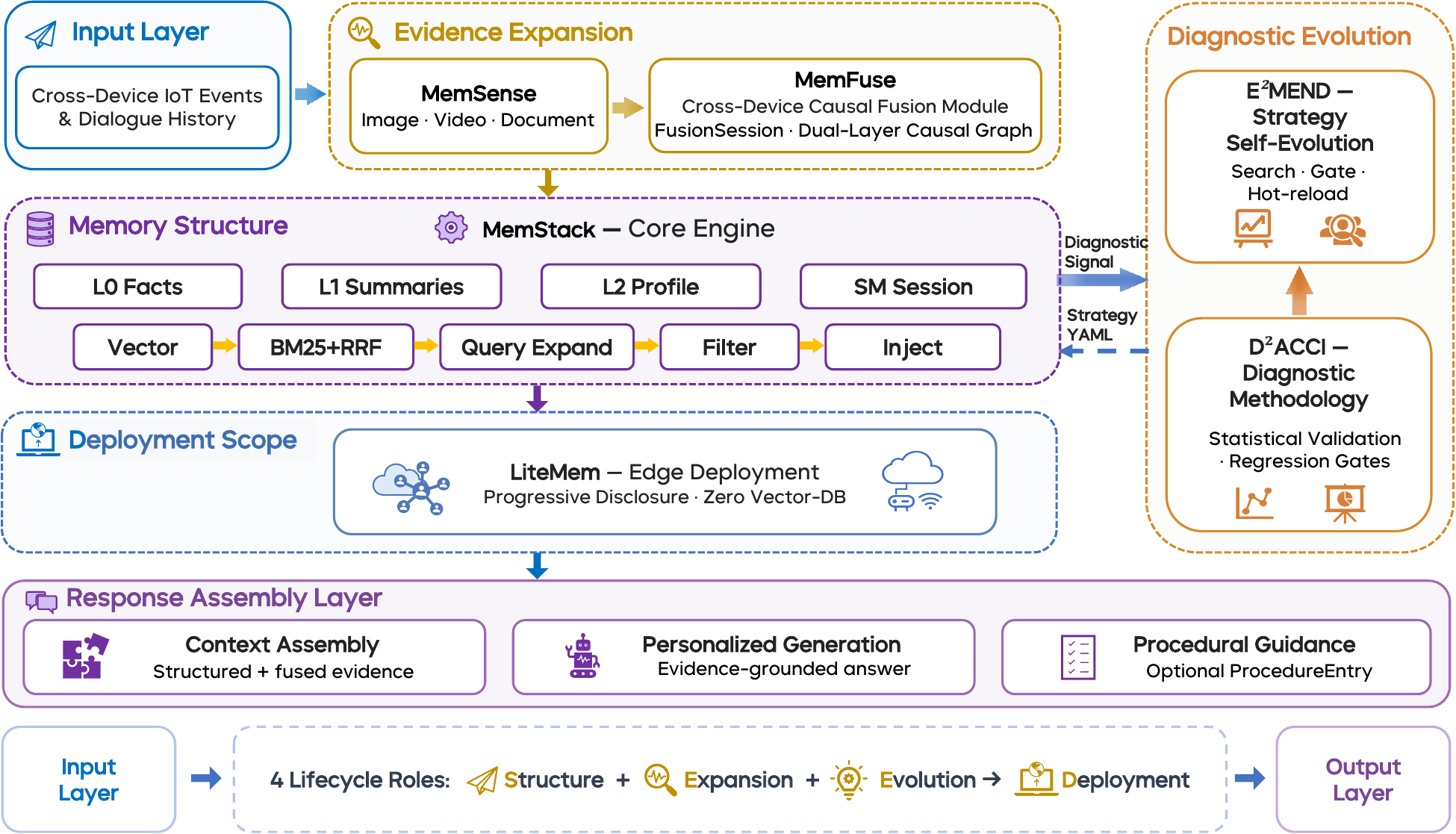}
    \caption{\MiMemory flow.}
    \label{fig:architecture}
\end{figure}

The four lifecycle roles defined in Section~\ref{sec:introduction} map the contract obligations to concrete modules. \MemStack maintains the runtime substrate, \MemSense and \MemFuse admit heterogeneous evidence, \DACCI and \EMEND govern bounded change, and \LiteMem exposes a transfer-feasibility substrate. Figure~\ref{fig:architecture} summarizes this module map and the typed information flow between roles.

\paragraph{Data-flow walkthrough.}
The running example in Section~\ref{sec:running-example} is a module-interface walkthrough, not a second scenario. Source observations first enter the Expansion role: \MemSense assigns stable identity to visual evidence, and \MemFuse links related device events into MemoryPack-style payloads when causal evidence is sufficient. Those payloads then enter the Structure role through \MemStack, where L0/L1/L2/SM records preserve atomic observations, session-level summaries, and stable family constraints.

At answer time, retrieval and context assembly select the relevant fused and structured evidence while preserving source identifiers in the assembled context. If the answer fails, the retained trace becomes an Evolution input for \DACCI/\EMEND: the failure can be attributed to missing admission, failed fusion, retrieval loss, filtering, context packing, or generation. When Deployment constraints apply, \LiteMem tests which parts of this path can be represented as repository-native files with Git provenance.

\paragraph{Lifecycle stages and interfaces.}
The framework also defines a stage sequence. Sources first produce dialogue, device, and visual observations; admission and ingestion normalize them; representation stores L0/L1/L2/SM records, fused nodes, and optional procedural hooks; retrieval and context assembly select bounded evidence; feedback updates or invalidates memory; governed evolution changes only accepted policies; and deployment tests which obligations survive under lighter substrates. Appendix~\ref{app:memorystack-extended} contains the compact lifecycle-stage map, while Section~\ref{sec:evaluation} reports evidence levels.

Five typed payload families make this stage sequence explicit across module boundaries. \texttt{FusedEvent} and \texttt{PerceptionFact} records move source evidence into \MemStack; optional \texttt{ProcedureEntry} records move from \MemStack to context assembly; \texttt{DiagnosticSignal} moves from \MemStack to \EMEND; and \texttt{StrategyArtifact} moves from \EMEND back to \MemStack. Table~\ref{tab:interface-schema} and the extended interface-contract table in Appendix~\ref{app:memorystack-extended} specify which fields must survive a boundary for provenance, diagnosis, and rollback to remain possible.

\paragraph{Design rationale.} The design principle is single-concern ownership. \MemFuse handles device-stream causality, \MemSense handles visual grounding, \MemStack handles storage/retrieval correctness, \EMEND handles bounded strategy search, and \DACCI handles evolution governance. Boundary crossings must pass through a typed payload. This separation enables two properties:
\begin{itemize}
    \item \textbf{Independent evolution}: \EMEND can define a search space with up to $10^6$ strategy combinations while limiting the risk of corrupting the retrieval framework (Section~\ref{sec:emend}).
    \item \textbf{Incremental integration}: \MemFuse and \MemSense connect through typed payloads, allowing new source modalities to integrate with \MemStack without changing the core storage and retrieval abstraction (Section~\ref{sec:source}).
\end{itemize}

\noindent Extended technical details---retrieval formulas, algorithm pseudocode, candidate traces, schema excerpts, error analyses, and protocol configuration---are collected in the appendix.

%% ---------------------------------------------------------------------------
\section{Structure: Structured Memory Substrate with \MemStack}
\label{sec:memorystack}

The \textbf{Continuity} requirement asks how later answers can reuse durable user state across turns without hiding which observations, memory items, and policy decisions justify that reuse. \MemStack makes this requirement concrete as the runtime substrate of \MiMemory: it turns histories and observations into typed records, preserves provenance across memory stages, and exposes retrieval, assembly, and invalidation traces for downstream governance. Later chapters build on this substrate by admitting richer evidence, evolving its policies, and testing whether the same audit obligations hold in lighter deployment settings. Optional procedural hooks share the same substrate, but they remain design-only and are not evaluated as standalone modules in this report.

\ataglance{RQ1 --- Continuity / Structure}{\MemStack runtime substrate}{typed L0/L1/L2/SM records plus retrieval, assembly, and invalidation traces}{controlled-reference memory QA; optional procedural hooks remain design-only}

%% ---------------------------------------------------------------------------
\subsection{Design Motivation: Continuity as an Audit Obligation}
\label{sec:memorystack-principles}

A personal AI assistant can lose continuity at several stages: it may fail to write the relevant event, retrieve a semantically adjacent but wrong memory, discard correct evidence during filtering, or answer incorrectly despite having the needed evidence. These failures motivate the central invariant of the Structure role:

\begin{quote}
\emph{Evidence should be written at the appropriate granularity, remain distinguishable across time, be recoverable through complementary search paths, survive filtering and invalidation, and fit within a bounded context budget.}
\end{quote}

This invariant is evaluated through traceable evidence flow, not a single universal threshold. Retrieval recall and context preservation are run-level validation checks interpreted within each benchmark slice; concrete values and formulas are reported as implementation details in Appendix~\ref{app:memorystack-extended}. The claim is architectural: continuity becomes diagnosable when memory stages emit inspectable artifacts.

%% ---------------------------------------------------------------------------
\subsection{Method: Trace-Producing Memory Runtime}
\label{sec:memorystack-runtime}

\begin{figure}[h]
\centering
\includegraphics[width=0.99\textwidth]{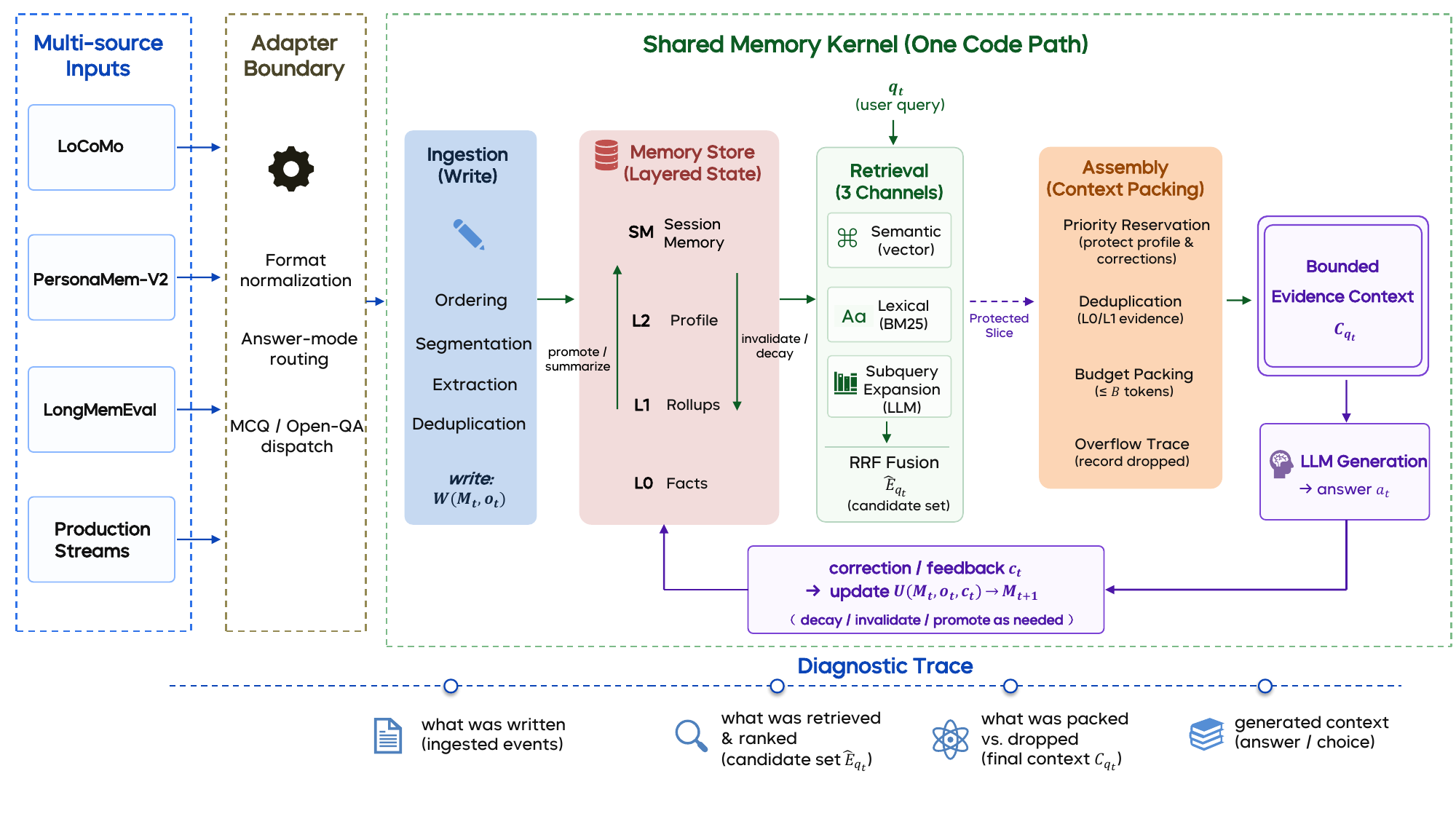}
\caption{\MemStack runtime.}
\label{fig:memorystack-runtime}
\end{figure}

The runtime is organized as an observable serving path, not a single opaque retrieval call. Its contribution is a joint algorithmic and systems formulation: memory organization, retrieval composition, and update are defined so downstream evaluation can attribute gains and failures to specific stages. Figure~\ref{fig:memorystack-runtime} illustrates the path: adapters normalize task inputs, layered storage maintains typed state, retrieval recovers candidate evidence, context assembly packs bounded evidence, and diagnostics record where evidence is retained, transformed, or dropped.

The shared serving path has two purposes. First, benchmark-specific behavior is encoded as configuration on top of a shared kernel instead of separate code paths. Second, the same stages produce comparable traces across LoCoMo, PersonaMem-V2, and LongMemEval, so later \DACCI/\EMEND analysis can attribute failures to stage-level causes rather than benchmark scripts.

\paragraph{Layered state.}
\label{sec:memorystack-storage}
The runtime uses a four-layer operational hierarchy so evidence loss stays diagnosable, not opaque. \textbf{L0} atomic facts support precise grounding; \textbf{L1} session or topic rollups preserve coverage; \textbf{L2} profile entries support stable personalization; and \textbf{SM} session memory maintains unresolved local context. The layers are not proposed as a cognitive model. They are an observability decomposition, so a failure can be localized to missing event capture, over-compression, stale profile updates, or session-boundary handling. Detailed field formats, retention knobs, and retrieval parameters are deferred to Appendix~\ref{app:memorystack-extended}.

\paragraph{Ingestion, retrieval, and assembly.}
\label{sec:memorystack-ingestion}
The serving contract has one rule: every admitted observation keeps temporal order, source identity, and a path back to the evidence that justified it. Long or topic-shifting histories may use supplemental extraction, but this chapter treats it as an implementation detail of the same admission contract; batching, segmentation, and concurrency details are reported in Appendix~\ref{app:memorystack-extended}. Retrieval combines semantic, lexical, and subquery-expanded channels because each channel recovers a different failure mode. Context assembly then reserves high-priority profile and correction constraints, deduplicates L0/L1 evidence, and records overflow traces rather than silently dropping constraints. Together these stages create the audit path needed to distinguish ingestion failure, retrieval failure, assembly failure, and generation error.

\begin{table}[!htbp]
\centering
\caption{Layer contract.}
\label{tab:memorystack-layers}
\compacttable
\begin{tabularx}{\textwidth}{l>{\hsize=0.7\hsize}X >{\hsize=1.15\hsize}X >{\hsize=1.15\hsize}X}
\toprule
\textbf{Layer} & \textbf{Write trigger} & \textbf{Primary role} & \textbf{Observable trace} \\
\midrule
L0 & Per-event capture & Atomic facts and temporal grounding & Event/source ids and provenance pointers \\
L1 & Session/topic boundary & Coverage-preserving local summary & Summary lineage and supersession links \\
L2 & Stable profile update & Long-horizon preferences and constraints & Profile diffs and correction history \\
SM & Current turn / open loop & Short-lived context and unresolved state & Session-window trace and overflow record \\
\bottomrule
\end{tabularx}
\end{table}
\FloatBarrier

The layer contract stays compact: the chapter includes only the detail needed to localize later failures, while the precise schema and retention values remain in the appendix.

\paragraph{Formalization: memory organization, retrieval, and update.}
\label{sec:memorystack-formal}
Let the operational memory state at turn $t$ be $M_t = (L_{0,t},\; L_{1,t},\; L_{2,t},\; SM_t)$, where each component is a layered memory store matching Figure~\ref{fig:memorystack-runtime}: $L_{0,t}$ holds atomic fact records with source links, $L_{1,t}$ holds session or topic rollups, $L_{2,t}$ holds stable profile and correction entries, and $SM_t$ (Session Memory) holds unresolved short-lived context for the current turn.

This tuple is the structured realization of the abstract $M_t$ in Section~\ref{sec:overview}. $q_t$ is the same user-request symbol and $o_t \in O_t$ an individual admitted observation. An observation enters through a layer-aware write operator $W$ that may create, summarize, promote, or invalidate records while preserving provenance:
\begin{equation}
M_t^{\mathrm{write}} = W(M_t, o_t).
\label{eq:memstack-write}
\end{equation}
For a query $q_t$, three retrieval operators $R$ (for \emph{Retrieval}) combine complementary channels---semantic ($R_{\mathrm{sem}}$, vector similarity), lexical ($R_{\mathrm{lex}}$, BM25), and expansion-based ($R_{\mathrm{exp}}$, LLM subquery rewriting):
\begin{equation}
\hat{E}_{q_t} = R_{\mathrm{sem}}(q_t, M_t) \cup R_{\mathrm{lex}}(q_t, M_t) \cup R_{\mathrm{exp}}(q_t, M_t).
\label{eq:memstack-retrieval}
\end{equation}
Here $\hat{E}_{q_t}$ is the candidate evidence set before budgeting, corresponding to the overview's $E_{q_t}$ after channel union but before assembly. The selected evidence is then budgeted and deduplicated into the assembled context:
\begin{equation}
C_{q_t} = A(\hat{E}_{q_t}, M_t; B),
\label{eq:memstack-assembly}
\end{equation}
where the assembly step keeps a reserved slice for profile corrections and session-memory constraints before packing the rest under budget $B$. The symbol $C_{q_t}$ is the same bounded generation context as in Section~\ref{sec:overview}. When correction or feedback $c_t$ arrives after generation, the update operator
\begin{equation}
M_{t+1} = U(M_t, o_t, c_t)
\label{eq:memstack-update}
\end{equation}
can decay, summarize, promote, or invalidate records. This formalization is operational, not cognitive: it describes audit-visible state transitions, not a psychological model.

\paragraph{Running trace: Ethan's training handoff.}
Continuing the scenario from Section~\ref{sec:running-example}, the current equipment reminder is recorded in L0, the unresolved trip context stays in SM, the repeated practice routine is summarized into L1, and the shoe-storage correction updates L2. At answer time, context assembly protects the correction record from being crowded out by lower-priority evidence, so the response can distinguish a current reminder from a stable profile rule. The trace exercises the full continuity path from writing through retrieval, packing, and update.

\paragraph{Optional procedural hooks.}
\label{sec:memorystack-hooks}
When a user-specific procedure is enabled, \MemStack can attach a compact trigger--procedure record to the same context-assembly path. The hook is admitted only after factual retrieval passes provenance and confidence gates, and it is injected as operational guidance, not factual evidence. In this report, these hooks remain design-only and are not benchmarked separately.

\paragraph{Kernel--adapter boundary.}
\label{sec:memorystack-adapters}
The shared kernel owns ingestion, storage, retrieval, assembly, and diagnostics; adapters normalize benchmark input shape and answer mode. This boundary keeps PersonaMem's MCQ logic, LoCoMo's temporal categories, and LongMemEval's per-question haystacks outside the memory kernel, so continuity improvements are not hidden inside benchmark-specific forks. Because this report does not provide separate benchmark evidence for optional procedural hooks, their representation details and diagnostic examples remain in Appendix~\ref{app:memorystack-hooks}.

%% ---------------------------------------------------------------------------
\subsection{Validation: Controlled Reference for Continuity}
\label{sec:memorystack-results}

\MemStack is evaluated as the controlled-reference runtime anchor for the lifecycle framework, not as a universal leaderboard claim. The evaluation asks whether the shared kernel can remain competitive with a reproduced reference while adding stage-local traces for diagnosis. External systems such as Mem0, AMA, MemMachine, DCPM, and HMO~\cite{chhikara2025mem0,ama2026,memmachine2026,dcpm2026,hmo2026} remain relevant context, but their public numbers use different models, judges, prompts, and sampling policies.

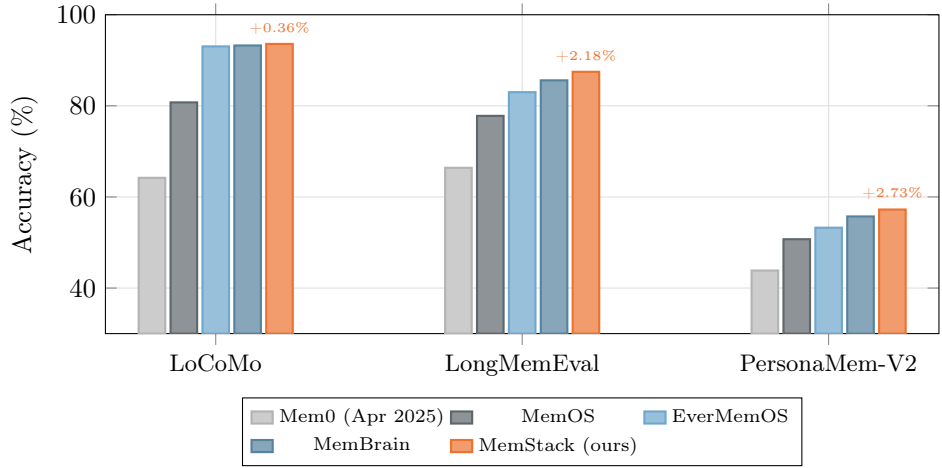
\begin{figure}[h]
\centering
\begin{tikzpicture}
\begin{axis}[
    width=0.76\textwidth,
    height=5.8cm,
    ybar,
    bar width=10pt,
    ylabel={Accuracy (\%)},
    symbolic x coords={LoCoMo,LongMemEval,PersonaMem-V2},
    xtick=data,
    x tick label style={font=\small},
    ymin=30, ymax=100,
    grid=major,
    grid style={draw=gray!25},
    legend style={font=\scriptsize, at={(0.5,-0.20)}, anchor=north, legend columns=3},
    legend image code/.code={\draw[#1] (0cm,-0.08cm) rectangle (0.28cm,0.08cm);},
    every axis plot/.append style={thick},
    enlarge x limits=0.18,
    nodes near coords style={font=\tiny, anchor=south, text=hxOrange},
]

\addplot[fill=gray!40, draw=gray!60] coordinates {(LoCoMo,64.20) (LongMemEval,66.40) (PersonaMem-V2,43.85)};
\addplot[fill=hxNavy!50, draw=hxNavy!70] coordinates {(LoCoMo,80.76) (LongMemEval,77.80) (PersonaMem-V2,50.72)};
\addplot[fill=hxTeal!50, draw=hxTeal!70] coordinates {(LoCoMo,93.05) (LongMemEval,83.00) (PersonaMem-V2,53.25)};
\addplot[fill=hxBlue!50, draw=hxBlue!70] coordinates {(LoCoMo,93.25) (LongMemEval,85.60) (PersonaMem-V2,55.72)};
\addplot[fill=hxOrange!70, draw=hxOrange, nodes near coords, point meta=explicit symbolic]
coordinates {(LoCoMo,93.59) [+0.36\%] (LongMemEval,87.47) [+2.18\%] (PersonaMem-V2,57.24) [+2.73\%]};

\legend{Mem0 (Apr 2025), MemOS, EverMemOS, MemBrain, \MemStack (ours)}
\end{axis}
\end{tikzpicture}
\caption{Controlled-reference comparison on three benchmarks (GPT-4.1-mini backbone). All baselines share the same evaluation harness from the EverMemOS evaluation; Mem0 is labeled by the Apr~2025 release of the original Mem0 paper. Orange labels report \MemStack's relative improvement over MemBrain on each benchmark.}
\label{fig:stack-results}
\end{figure}

Figure~\ref{fig:stack-results} provides the controlled-reference anchor: \MemStack and the reproduced MemBrain baseline share the same harness, model roles, data splits, and diagnostic vocabulary. The associated diagnostic surfaces are: BM25+RRF and agentic expansion (LoCoMo), supplement path and forget constraints (PersonaMem-V2), and agentic retrieval with session cleanup (LongMemEval). The narrow LoCoMo margin reflects parity-level competitiveness, while LongMemEval and PersonaMem-V2 report larger relative gains over the controlled MemBrain reference. Representative failure-fix trajectories and case-level traces are preserved in Appendix~\ref{app:dacci-extended}.

\paragraph{Main implication.}
\MemStack establishes the continuity stage of \MiMemory as both an algorithmic and systems substrate. User state can be represented, retrieved, assembled, invalidated, and audited through a shared memory kernel rather than an opaque RAG component. Its contribution is to expose multi-granularity state, hybrid retrieval, budgeted packing, and stage-local diagnostic traces within one controlled-reference setting. This role also explains the ordering of the later modules: Evidence Admission modules emit payloads into the same continuity substrate with provenance, \DACCI/\EMEND modify policies through auditable strategy artifacts, and \LiteMem is evaluated by how much of the same continuity contract survives in a repository-native substrate.

%% ═══════════════════════════════════════════════════════════════════════════════
\section{Expansion: Evidence Expansion with \MemSense and \MemFuse}
\label{sec:source}

The \textbf{Evidence Admission} requirement asks how observations beyond dialogue can become admissible memory evidence. For Personal AI, memory extends beyond a larger text corpus: evidence arrives through images, phones, cars, wearables, cameras, home devices, and tools as well as through dialogue. \MemSense and \MemFuse form the Evidence Admission layer for RQ2. \MemSense converts real-world scenes into structured, identity-preserving memory objects; \MemFuse aligns atomic events from different devices, sessions, and times into cross-device causal evidence for \MemStack. The shared requirement is provenance: additional sources are useful to the extent that their identity, time, device provenance, confidence, and relation to the user query remain inspectable after retrieval.

The admission path runs from real-world scenes and device observations $\rightarrow$ multimodal and event-level atomic memories $\rightarrow$ cross-device fused packs $\rightarrow$ structured assistant memory. As the source-side counterpart to \MemStack, this layer expands what can count as admissible evidence before storage and retrieval: multimodal memory grounds visual context, while \MemFuse connects evidence observed through different entry points.

\ataglance{RQ2 --- Evidence Admission}{\MemSense/\MemFuse bridge}{IKB entries and MemoryPacks preserve identity, time, provenance, confidence, and cross-event links}{\MemSense evidence plus preliminary \MemFuse evidence}

%% ---------------------------------------------------------------------------
\subsection{Design Motivation: World Evidence Beyond Dialogue}
\label{sec:source-problem}

Dialogue-only memory is insufficient in multi-device Personal AI ecosystems such as Human-Car-Home. User state is split across complementary sources: dialogue captures intent and self-reports, while perception and device events capture implicit state and environmental evidence. A restaurant destination selected on the phone may shape car navigation; a wearable sleep signal may contextualize a family routine; and an in-car reminder before a child's training session may depend on a home-camera observation, a calendar event, route timing, and prior user corrections, rather than any single dialogue turn.

This split creates two evidence-admission gaps. The first is a \textbf{multimodal identity gap}: images and visual contexts need stable representation as evidence objects, not disposable attachments or unstable captions. The second is a \textbf{cross-device fusion gap}: related observations may describe the same situation without sharing keywords. If a home camera sees a training bag near the doorway, a calendar records basketball practice, and the vehicle estimates that returning home would still preserve on-time arrival, no single event is sufficient; the assistant needs to recover the temporal-causal chain before it can issue an in-car reminder.

Let $O$ denote the typed observation universe used at serving time in Section~\ref{sec:overview}: text, images, and device events are all observations, but each keeps its own payload schema and provenance fields. We write $q_t$ for the current user request and $e$ for an individual candidate evidence item. The Evidence Admission objective therefore selects an evidence set $E_{q_t}$ by balancing two per-item terms: relevance to $q_t$ and provenance strength for auditability.
\begin{equation}
E_{q_t} = \operatorname*{arg\,max}_{E \subseteq O \cup G \cup K} \sum_{e \in E} \bigl(\operatorname{rel}(e,q_t) + \lambda \operatorname{prov}(e)\bigr)
\quad \text{s.t.} \quad \sum_{e \in E} \operatorname{tok}(e) \le B,
\label{eq:source-evidence}
\end{equation}
where $E_{q_t}$ matches the selected-evidence notation in Section~\ref{sec:overview}, $G$ is the cross-device memory graph, $K$ is the structured Image Knowledge Base (IKB), $B$ is the answer-time evidence budget, $\lambda$ balances relevance against provenance, $\operatorname{rel}(e,q_t)$ scores how well a candidate evidence item supports the current request, and $\operatorname{prov}(e)$ is the traceability term that rewards source identity, time/session metadata, device provenance, and causal links. Thus, the objective prefers evidence that is both on-topic and auditable, while still fitting within the budget. The exact scoring function is implementation-specific; the equation states the selection contract expected of this layer before downstream retrieval or generation. In Section~\ref{sec:overview} terms, this layer expands the admissible evidence pool inside $O$ while keeping the selected subset $E_{q_t}$ and the assembled context $C_{q_t}$ governed by the same audit contract.

%% ---------------------------------------------------------------------------
\subsection{Method: Evidence Admission Pipeline from Grounding to Fusion}
\label{sec:source-solution}

The pipeline is organized around one question: can a non-dialogue observation become assistant memory without losing traceability? \MemSense grounds individual scenes and image observations as stable, identity-preserving evidence objects. \MemFuse then decides whether related observations from different devices, sessions, or times should stay atomic or be fused into a MemoryPack with explicit provenance and causal edges. Both components feed \MemStack through typed payloads instead of bypassing the memory runtime.

The stages are deliberately sequential. Evidence that cannot be named or traced should not be fused into higher-level memory; once evidence is fused, the resulting pack must still expose the atomic observations that justified it. This design separates source construction from answer-time selection: source modules expand the evidence pool, while \MemStack and its diagnostic traces still decide which evidence reaches the final context.

\subsubsection{Grounding Multimodal Evidence as Atomic Memory}
\label{sec:multimodal}

\begin{figure}[H]
\centering
\includegraphics[width=0.95\textwidth]{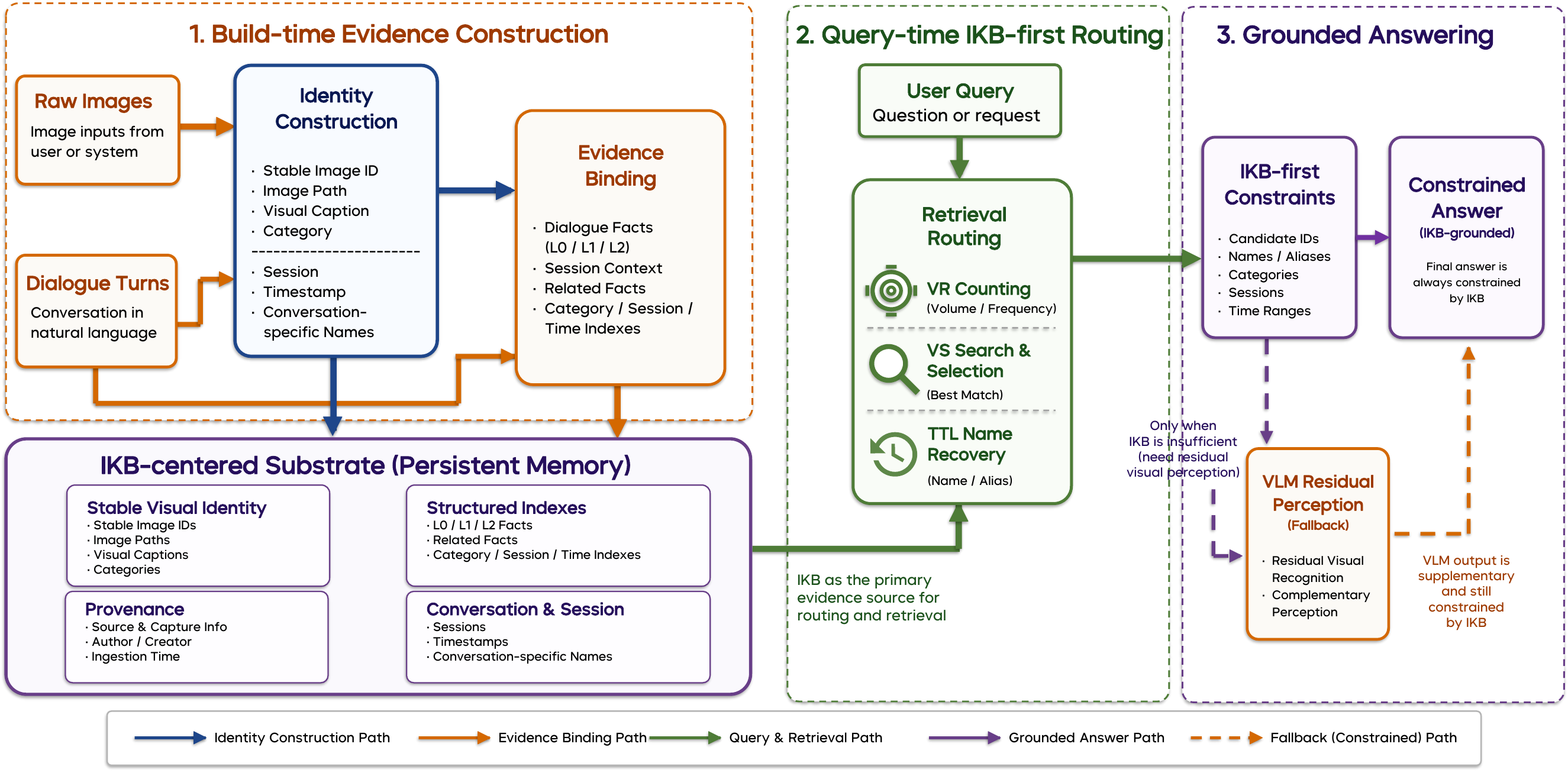}
\caption{IKB path.}
\label{fig:mm-architecture}
\end{figure}

For each image $x_j$, \MemSense assigns a stable conversation identifier such as \texttt{D3:IMG\_001}, a session $s_j$, a date $\mathrm{date}_j$, a visual caption $c_j$, and a conversation-specific name $n_j$ extracted from surrounding dialogue. The implemented \MemSense pipeline (Figure~\ref{fig:mm-architecture}) produces complementary stores inside \MemStack. L0/L1/L2 and Session Memory anchor images to dialogue facts and temporal context. The IKB records image ID, path, date, session, conversation name, category, and related facts for deterministic counting, filtering, and name recovery. Multimodal retrieval indexes then provide probabilistic text--text, image--image, and text--image search.

The IKB entry for an image is:
\begin{equation}
K(x_j) = (\mathrm{id}_j, \mathrm{path}_j, s_j, \mathrm{date}_j, n_j, \mathrm{cat}_j, c_j, \mathit{Rel}_j),
\label{eq:ikb-entry}
\end{equation}
where $\mathrm{cat}_j$ is produced by a staged VLM/dialogue normalization pass, not by a fixed ontology alone. The builder extracts a visual category, reconciles conversation names and synonyms, and splits over-merged groups when image similarity is low. $\mathit{Rel}_j$ is the set of related L0 facts selected by image-ID match, session/date overlap, and content-keyword match (written $\mathit{Rel}$ rather than $R$ to avoid overloading the retrieval operators $R_{\mathrm{sem}}, R_{\mathrm{lex}}, R_{\mathrm{exp}}$ in Section~\ref{sec:memorystack}). In the implemented prototype, IKB is a structured side table indexed by image id, session/date, category, and name, while L0/L1/L2 remain the general memory store. The two stores are linked by stable IDs instead of being collapsed into a single vector index. This binding matters because many multimodal memory questions ask what was shown, counted, or named, not what a standalone VLM would label in isolation.

At answer time, the intent router maps visual-memory questions to execution strategies such as high-recall retrieval, date-filtered retrieval, session-direct retrieval, or \texttt{ikb\_query}. The routing rule is:
\begin{quote}
\emph{When structured IKB records can resolve a visual identity question, they should constrain probabilistic visual recognition rather than only supplement it.}
\end{quote}
As a result, visual recall/counting (VR) uses IKB counts and image IDs, visual search/enumeration (VS) uses category/session/time-filtered candidate lists, and text-to-look naming (TTL) uses conversation-specific names before residual VLM perception.

\subsubsection{Fusing Cross-Device Evidence into MemoryPacks}
\label{sec:memfuse-fusion}

\MemFuse extends the same evidence-preservation principle to asynchronous device events. Perception modules and dialogue devices emit observations as \texttt{AtomicEvent} records with provenance, time, importance, urgency, and visibility metadata. For each incoming event $e_t$, \MemFuse retrieves candidate neighbors, expands them through graph links, and sends the neighborhood plus session context to an LLM judge. The judge chooses one of three actions:
\begin{equation}
\mathrm{Fuse}(e_t, \mathcal{N}_t) \;\in\; \{\texttt{CreatePack},\; \texttt{UpdatePack}(p_k),\; \texttt{Standalone}\},
\label{eq:fusion-decision}
\end{equation}
where $\mathcal{N}_t$ is the retrieved neighborhood and $p_k$ denotes an existing \texttt{FusedNode}. \texttt{CreatePack} forms a new activity-level node from the incoming event $e_t$ and its retrieved neighborhood; \texttt{UpdatePack} attaches $e_t$ to an existing pack; and \texttt{Standalone} keeps $e_t$ as an isolated atomic event when evidence for fusion is insufficient. This explicit three-way outcome keeps fusion conservative: ambiguous events remain recoverable as atomic evidence instead of being forced into a premature MemoryPack.

\begin{figure}[H]
\centering
\includegraphics[width=0.87\textwidth]{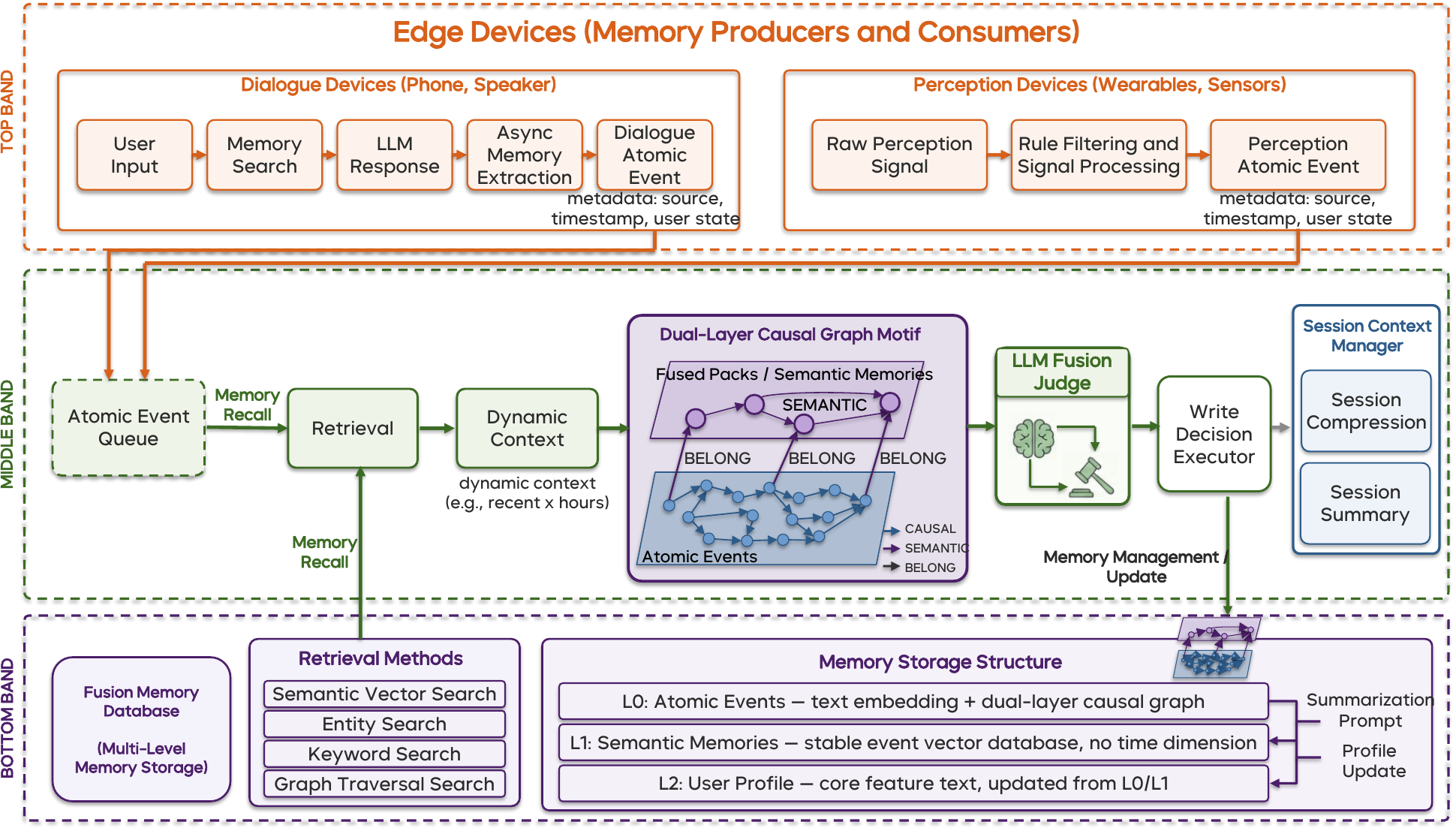}
\caption{\MemFuse pack path.}
\label{fig:memfuse-architecture}
\end{figure}

\MemFuse uses session-level fusion, not a stateless pairwise classifier. It maintains a three-zone \texttt{FusionSession}: recent daily summaries, same-day accumulated events, and the current event with retrieved candidates. These zones are configuration-level budgets: when a zone is full, lower-confidence or older candidates are summarized or evicted, while accepted graph edges remain stored.

Accepted judgments are stored as a dual-layer graph (Figure~\ref{fig:memfuse-architecture}). \texttt{AtomicEvent} nodes preserve fine-grained provenance; \texttt{FusedNode} nodes store activity-level packs, namely the MemoryPacks consumed by \MemStack. \texttt{CAUSAL}, \texttt{SEMANTIC}, and \texttt{BELONG} edges capture causal links, retrieval neighbors, and event-to-pack membership. The processing cycle is: initialize zone state, collect candidates, judge fusion, persist the graph and MemoryPack, then summarize or evict transient zone contents. Additional ingestion and session-graph diagrams are provided in Appendix~\ref{app:source-breadth-extended}, so the chapter can stay focused on source-to-memory logic.

At query time, \MemFuse searches atomic events and fused nodes with vector and BM25 retrieval, merges them with reciprocal-rank fusion, and treats the merged results as seed nodes. Seeds expand along \texttt{CAUSAL}, \texttt{SEMANTIC}, and \texttt{BELONG} edges, allowing ``anything to check before training'' to reach ``training bag near doorway'' despite limited lexical overlap.

In the running example (Section~\ref{sec:running-example}), graph expansion from the training-check intent reaches the fused pack that links bag, calendar, and route evidence through relation edges. Each fused node occupies one context slot but can expand into exact device and dialogue evidence when \MemStack needs provenance. Table~\ref{tab:rq2-evidence-invariants} summarizes how \MemSense and \MemFuse preserve the RQ2 invariants---identity, time, provenance, and causal relation---across the Evidence Admission layer.

\begin{table}[h]
\centering
\caption{Evidence invariants.}
\label{tab:rq2-evidence-invariants}
\compacttable
\begin{tabularx}{\textwidth}{>{\hsize=0.5\hsize}X >{\hsize=1.25\hsize}X >{\hsize=1.25\hsize}X}
\toprule
\textbf{Invariant} & \textbf{\MemSense} & \textbf{\MemFuse} \\
\midrule
Identity & Stable image ids, conversation names, IKB entries & AtomicEvent ids and FusedNode / MemoryPack ids \\
Time & Session/date fields linked to visual records & Event timestamps, daily/session zones, temporal graph edges \\
Provenance & Image path, dialogue context, related L0 facts & Device source, visibility metadata, source-event links \\
Causality / relation & Related-fact links for visual context & \texttt{CAUSAL}, \texttt{SEMANTIC}, and \texttt{BELONG} edges \\
Downstream audit & IKB-to-L0/L1/L2 links & MemoryPack payloads consumed by \MemStack \\
\bottomrule
\end{tabularx}
\end{table}

%% ---------------------------------------------------------------------------
\subsection{Validation: Evidence Coverage in Realistic Daily Scenarios}
\label{sec:source-effect}

The two components cover complementary sides of Evidence Admission. In this report, RQ2 covers construction and provenance preservation under module-specific harnesses: \MemSense tests the grounding half of the admission contract, while \MemFuse tests whether grounded observations remain connected across devices and time.

\MemSense is evaluated on Mem-Gallery~\cite{memgallery2026}, a public benchmark for multimodal long-term conversational memory in MLLM agents. Mem-Gallery contains 240 multi-session dialogues across 20 scenarios, 3,962 dialogue rounds, 1,003 contextual input images, and 1,711 QA pairs with annotated clues. The result is treated as module-level evidence for the current IKB-centered multimodal memory pipeline.

\begin{table}[H]
\centering
\caption{Mem-Gallery results (\MemSense: GPT-4.1-mini backbone, GPT-4o-mini judge).}
\label{tab:mm-status}
\compacttable
\begin{tabularx}{\textwidth}{lrlX}
\toprule
\textbf{Metric / Error Source} & \textbf{Value} & \textbf{Scope} & \textbf{Interpretation} \\
\midrule
Average judge accuracy & 89.15\% & 1,711 Qs & Primary metric; average fraction of correct votes across three judge runs \\
Strict binary accuracy & 88.19\% & 1,711 Qs & Counts questions with judge accuracy below 1.0 as wrong \\
Strict errors & 202 & 20 scenarios & Includes 34 partially correct borderline cases \\
Average latency & 35.3s & Offline module pipeline & Median 32.4s; P99 91.8s \\
Harness processing failure rate & 0\% & Offline module pipeline & No offline harness processing failures observed \\
Image-related error share & 58\% & VR + VS + TTL & Dominant localized residual error source \\
\bottomrule
\end{tabularx}
\end{table}

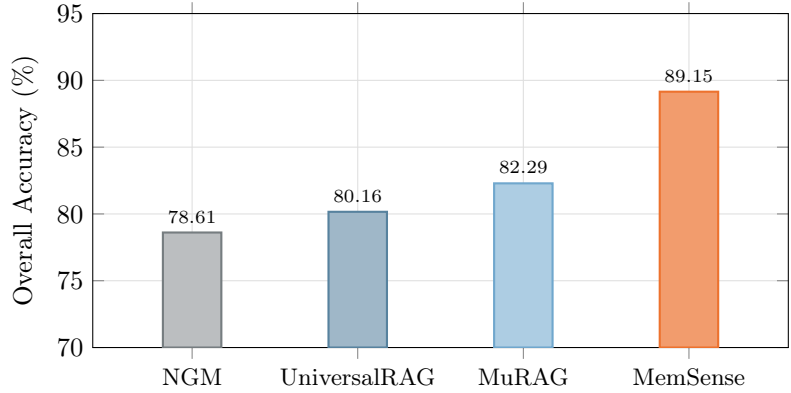
\begin{figure}[H]
\centering
\begin{tikzpicture}
\begin{axis}[
    width=0.65\textwidth,
    height=6cm,
    ybar,
    bar width=22pt,
    bar shift=0pt,
    ylabel={Overall Accuracy (\%)},
    xtick={0,1,2,3},
    xticklabels={NGM, UniversalRAG, MuRAG, MemSense},
    x tick label style={font=\small},
    xmin=-0.6, xmax=3.6,
    ymin=70, ymax=95,
    ytick={70,75,80,85,90,95},
    grid=major,
    grid style={draw=gray!25},
    every axis plot/.append style={thick},
    nodes near coords,
    nodes near coords style={font=\scriptsize, anchor=south, /pgf/number format/fixed},
]
\addplot[fill=hxNavy!30, draw=hxNavy!60] coordinates {(0,78.61)};
\addplot[fill=hxBlue!40, draw=hxBlue!70] coordinates {(1,80.16)};
\addplot[fill=hxTeal!40, draw=hxTeal!70] coordinates {(2,82.29)};
\addplot[fill=hxOrange!70, draw=hxOrange] coordinates {(3,89.15)};
\end{axis}
\end{tikzpicture}
\caption{Mem-Gallery overall accuracy comparison. Baselines use a unified Qwen-2.5-VL-7B backbone as reported in~\cite{memgallery2026}; \MemSense uses GPT-4.1-mini with GPT-4o-mini judge. The backbone difference precludes direct numerical ranking; the figure provides contextual positioning only.}
\label{fig:memsense-comparison}
\end{figure}

The 89.15\% and 88.19\% figures (Table~\ref{tab:mm-status}) differ only in aggregation: the former keeps the fractional three-vote judge score, while the latter applies a stricter binary rule and treats the 34 partially correct cases as errors. For reference, the Mem-Gallery paper~\cite{memgallery2026} reports the best-performing open-source baseline (MuRAG) at 82.29\% overall under a unified Qwen-2.5-VL-7B backbone, followed by UniversalRAG (80.16\%) and NGM (78.61\%). Figure~\ref{fig:memsense-comparison} shows these reference numbers. Direct numerical comparison is not claimed because the pipeline and backbone differ; the difference is treated as contextual support for the hypothesis that IKB-first structured routing may improve on end-to-end multimodal retrieval approaches. A controlled ablation isolating the IKB contribution under a shared backbone remains future work. The latency row reports offline pipeline latency under the evaluation harness, not production serving latency; environment details are not reported here.

The residual errors are stage-localized: image-category construction can over-merge names, cross-session visual recall can truncate candidate sets, and answer-time VLMs can override conversation-specific labels with general visual names. The development trajectory and representative VR/VS/TTL case traces are moved to Appendix~\ref{app:mm-extended}. The operational conclusion is that IKB should be the primary evidence source for structured visual questions, with VLM perception reserved for residual disambiguation. The latency row also sets a deployment boundary: the current multimodal path provides module-level support for evidence construction, but production use still requires caching, trigger narrowing, and fast-path routing for common visual queries.

Validation is organized around the evidence-admission contract, not a general multimodal leaderboard. \MemSense is judged by whether structured IKB records preserve identity, time, provenance, and downstream audit links; \MemFuse is judged by whether cross-device packs recover causal chains and conflict relations without erasing contradictory evidence. MemFuseBench therefore checks whether admissible evidence survives the transition from raw observation to typed memory.

MemFuseBench follows a Scene-to-Sensor paradigm: it first defines what happened, then generates what each device observed. It is designed to be system-agnostic in execution: the same question set and checklist judge apply to Direct LLM, Naive RAG, mem0, and \MemFuse, and scores reflect a common gpt-4o-mini judge pipeline.

Checklist scoring decomposes each answer into required evidence items: correct event identification, device-source attribution, causal link recovery, conflict handling, and final answer consistency. The judge marks each item before averaging into a dimension score; the rubric requires item-level yes/no decisions before the final score, not a free-form preference comparison. Because MemFuseBench is internal, these results are preliminary descriptive evidence for causal/fusion behavior, not an external-validity claim; external validation remains pending. The benchmark does not report a human upper bound, inter-annotator agreement, or oracle-fusion ceiling here, so absolute checklist scores indicate task-specific coverage instead of normalized task competence.

\begin{table}[h]
\centering
\caption{MemFuseBench scores.}
\label{tab:memfuse-dimension-results}
\compacttable
\resizebox{\textwidth}{!}{%
\begin{tabular}{lccccccc}
\toprule
% \textbf{System} & \textbf{Causal Reasoning} & \textbf{Information Fusion} & \textbf{Conflict Arbitration} & \textbf{Multi-User Synthesis} & \textbf{Cross-User Query} & \textbf{Perspective Difference} & \textbf{Overall} \\
\makecell{\textbf{System}\\\textbf{(4o-mini)}} & \makecell{\textbf{Causal}\\\textbf{Reasoning}} & \makecell{\textbf{Information}\\\textbf{Fusion}} & \makecell{\textbf{Conflict}\\\textbf{Arbitration}} & \makecell{\textbf{Multi-User}\\\textbf{Synthesis}} & \makecell{\textbf{Cross-User}\\\textbf{Query}} & \makecell{\textbf{Perspective}\\\textbf{Difference}} & \textbf{Overall} \\
\midrule
Direct LLM & 34.7\% & 31.6\% & 37.3\% & \textbf{32.3\%} & 25.6\% & 21.2\% & 30.2\% \\
Mem0 infer k=20 & 37.4\% & 31.8\% & \textbf{45.3\%} & 26.9\% & \textbf{34.6\%} & 15.0\% & 30.5\% \\
RAG k=20 & 33.3\% & 24.4\% & 41.3\% & 31.4\% & 28.8\% & 16.3\% & 27.9\% \\
\MemFuse{} k=20 & \textbf{41.4\%} & \textbf{42.2\%} & 33.3\% & 25.5\% & 34.4\% & \textbf{30.6\%} & \textbf{35.2\%} \\
\bottomrule
\end{tabular}%
}
\end{table}

Within that boundary, the dimension breakdown in Table~\ref{tab:memfuse-dimension-results} suggests better checklist coverage for \MemFuse{} k=20 on cross-device evidence construction. It reports the highest causal-reasoning, information-fusion, perspective-difference, and overall checklist scores among the compared systems, with a +4.7pp margin relative to the mem0 reference. Scores in this range (30--42\%) reflect granular sub-criterion coverage: each query is decomposed into multiple required evidence items, and partial credit is awarded per item.

The lower-scoring dimensions are also informative. \MemFuse scores lower on conflict arbitration than mem0 (33.3\% vs. 45.3\%) and lower on multi-user synthesis than Direct LLM (25.5\% vs. 32.3\%), suggesting that the current graph-fusion objective over-prioritizes causal/link coverage and under-models contradictory evidence ownership, recency, and user attribution. Cross-user query is near parity with mem0 (34.4\% vs. 34.6\%), while perspective difference shows the largest observed \MemFuse margin (30.6\% vs. the next-best 21.2\%).

\paragraph{Takeaway.}
\MemSense and \MemFuse expand the input surface of \MemStack while keeping its core abstraction unchanged. Visual scenes become atomic memory objects with stable IDs and conversation names; device streams become atomic events with provenance and visibility; fused packs connect these observations across devices and time. The contribution is to keep expanded evidence typed, traceable, and auditable.

%% ═══════════════════════════════════════════════════════════════════════════════
\section[Evolution: Governed Evolution from \DACCI to \EMEND]{Evolution: Governed Evolution from \DACCI to \EMEND}
\label{sec:emend}

A memory system cannot remain fixed as user behavior drifts, new sources appear, model endpoints change, and previously effective heuristics become stale. Once memory is treated as lifecycle infrastructure, improvement cannot be reduced to an untracked prompt tweak or retrieval edit. A policy change should be versioned, evaluated against fixed evidence, and accepted through gates and rollback records. This section defines the governance plane of \MiMemory: serving-time traces become evidence supporting or rejecting policy updates, \DACCI supplies the human-governed diagnostic loop, and \EMEND automates only the bounded strategy-search subset of that loop. The design draws on trace-driven multi-agent evolution patterns such as AEGIS in HarnessX~\cite{chen2026harnessx}, but the claim here is narrower: evidence-gated strategy evolution for memory systems.

\ataglance{RQ3 --- Evolution}{governance plane}{Diagnostic traces, strategy artifacts, gate reports, and rollback records bind each policy change}{fixed-harness \EMEND strategy search under the human-governed \DACCI loop}

\paragraph{Control boundary.}
The \EMEND gates and rollback records address a specific strategy-drift risk. They reject schema-illegal, unversioned, or degrading strategy artifacts within the fixed offline harness before the serving strategy is reloaded. Adversarial model safety, privacy enforcement, consent management, and abuse resistance remain platform-level validation layers outside the current evidence scope.

This separation clarifies authority: \DACCI decides what may change and how evidence is judged; \EMEND searches only the declared strategy space under those rules.

\subsection{Design Motivation: Manual Diagnostic Iteration Cannot Keep Pace}

The motivation starts with the bottleneck that makes bounded automation useful.

\MemStack and \DACCI establish the controlled memory runtime and diagnostic loop used in this report. The remaining bottleneck is scaling that diagnostic cycle. User behavior drift, new scenario onboarding, and model upgrades all require engineers to re-run the diagnostic-fix-verify cycle. As scenario coverage grows, the manual loop becomes more costly to sustain.

\EMEND automates the repeatable portion of \DACCI: diagnosis, strategy proposal, and gated verification under a locked framework and schema-constrained strategy space. It does not modify the framework, evaluation protocol, data schema, or governance criteria defined by \DACCI. The trade-off is deliberate: \EMEND can tune parameters, prompts, routing rules, and feature toggles, but any change that expands the capability boundary returns to human-governed \DACCI review.

\EMEND therefore differs from recent deep-research agents that learn from trajectory or case memory online~\cite{qiao2026mia,zhou2025memento}: evolution is treated as a governed strategy-artifact update, not unconstrained serving-time policy learning.

The next subsection defines the dual-loop boundary that separates human-governed capability changes from automated strategy search.

%% ═══════════════════════════════════════════════════════════════════════════════
\subsection{Method: Dual-Loop Framework for Bounded Evolution}

Bounded evolution starts by separating mutable strategy from locked framework. The two loops divide responsibility: \DACCI is the governance loop for framework, protocol, data schema, and acceptance criteria, while \EMEND is the automation loop for mutating declared strategy artifacts. This distinction supports regression control and reproducibility: when a candidate is accepted, the system can attribute the gain to a declarative strategy mutation rather than hidden framework drift.

\begin{table}[htbp]
\centering
\caption{Dual-loop division of labor.}
\label{tab:emend-search-boundary}
\compacttable
\begin{tabularx}{\textwidth}{l X X}
\toprule
\textbf{Scope} & \textbf{\DACCI governance loop} & \textbf{\EMEND automation loop} \\
\midrule
Declarative strategy space & Defines admissible fields, value ranges, invariants, and evolvable dimensions & Mutates extraction limits, retrieval weights, \texttt{top\_k}, rerank size, source windows, prompts, and feature toggles \\
Framework & Approves locked components: extraction code, retrieval pipeline, storage logic, context assembler & Reuses the locked framework without changing code or storage schema \\
Protocol & Defines benchmark data, judge policy, scoring rubric, gate criteria, and non-regression rules & Runs the fixed slice, applies gate checks, and reports paired deltas under those rules \\
Artifacts & Requires versioned artifacts, traces, gate reports, and rollback records for audit & Persists each candidate mutation, decision record, revert, and restore event \\
\bottomrule
\end{tabularx}
\end{table}

This boundary is motivated by an internal qualitative audit of 13 baseline memory providers. The reviewed systems typically package extraction, retrieval, pruning, and presentation into 200--500 lines of tightly coupled provider logic. As a result, changing one strategy often means rewriting the provider, which complicates A/B testing, rollback, and auditing. The capability matrix suggests that simultaneous support for serving-time memory updates, explicit failure learning, memory-attribution traces, and consolidation, not append-only growth, is still uncommon among the reviewed providers. The dual-loop design reframes this point as a division of labor: \DACCI governs capability and evidence contracts, while \EMEND executes bounded strategy automation against those contracts.

The division of labor in Table~\ref{tab:emend-search-boundary} also defines the handoff rule: if a candidate requires framework code, benchmark protocol, storage schema, or acceptance criteria to change, the issue is handed from \EMEND automation to \DACCI governance. Autonomous acceptance is limited to candidates that remain within the declared strategy space and pass bounded regression checks.

\begin{figure}[htbp]
\centering
\includegraphics[width=0.86\textwidth]{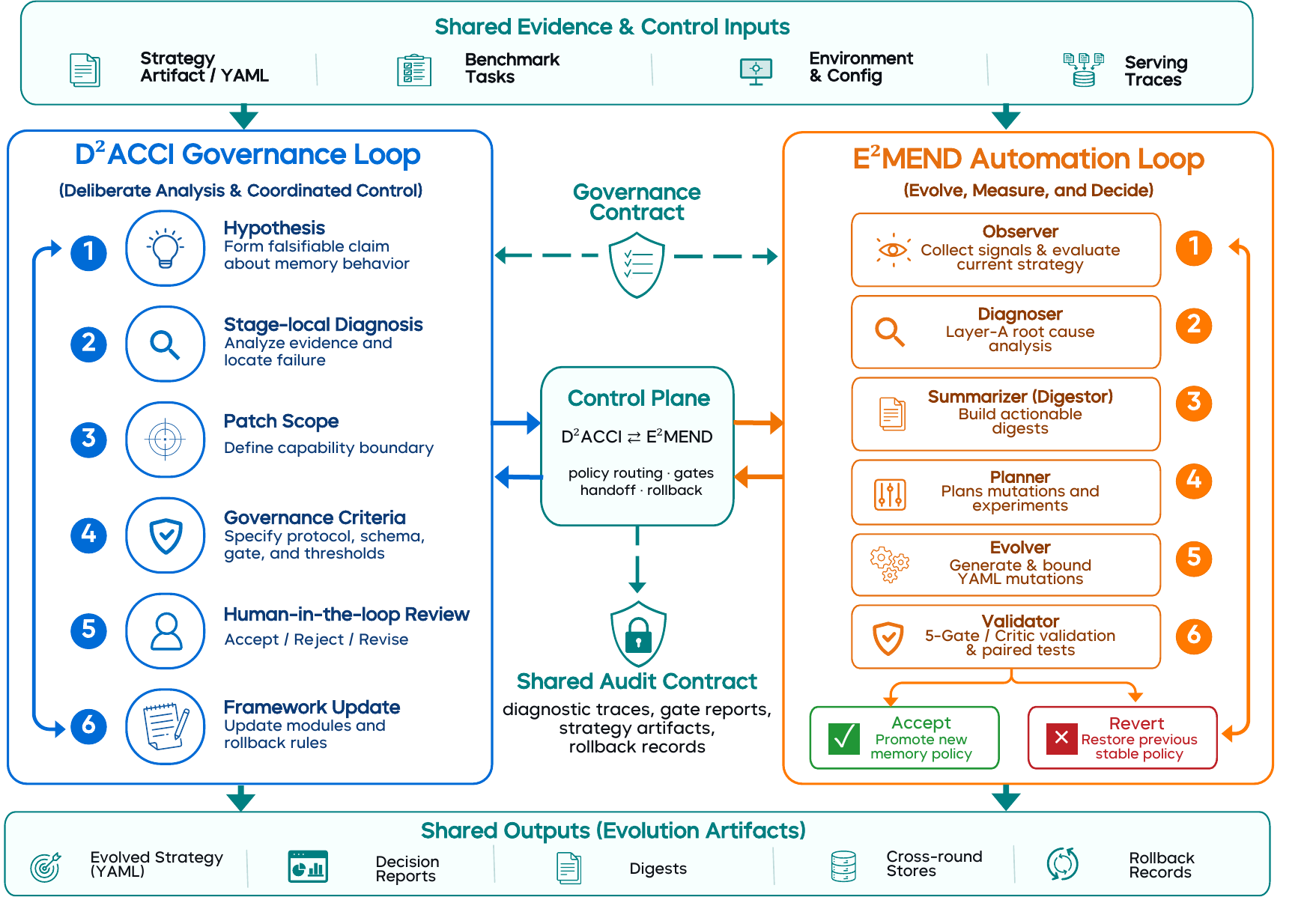}
\caption{Dual-loop framework for governed memory evolution.}
\label{fig:emend-evolution-loop}
\end{figure}

Figure~\ref{fig:emend-evolution-loop} makes the coordination explicit. \DACCI and \EMEND are two loops with distinct responsibilities and implementation paths. \DACCI decides what kinds of changes are legitimate and how evidence is judged. \EMEND repeatedly tests strategy mutations under those fixed rules, preserving the artifacts needed for audit and rollback.

\FloatBarrier

\subsubsection[DACCI: Human-in-the-Loop Evolution Governance]{\DACCI: Human-in-the-Loop Evolution Governance}
\label{sec:dacci}

Within this boundary, \DACCI (Diagnostic-Driven AI-Collaborative Iteration) is the human-in-the-loop governance loop for memory evolution. Humans define falsifiable hypotheses, AI agents help scale diagnosis and patch synthesis, and deterministic analyzers decide whether the resulting evidence justifies acceptance. This subsection formalizes the evidence contract that \EMEND later automates within a narrower strategy space.

The method rests on two invariants. First, \emph{stage-local attribution}: when a memory-dependent answer fails and sufficient evidence annotations are available, the diagnostic trace localizes the earliest evidence-loss stage among ingestion, retrieval, filtering, context assembly, and generation (formalized in Appendix~\ref{app:dacci-extended}). This avoids misdirecting retrieval fixes toward ingestion problems or generation fixes toward filtering problems. Second, \emph{constrained improvement}: a candidate configuration is accepted only when paired comparisons show positive evidence under the applicable gate and every question category satisfies its non-regression threshold. If a run lacks an explicitly reported p-value, confidence interval, or variance estimate, the comparison is treated as descriptive evidence only.

\paragraph{Iteration loop.}
A \DACCI round proceeds in four steps:
\begin{enumerate}
    \item \textbf{Hypothesis}: An engineer proposes a falsifiable claim (e.g., ``single-hop temporal questions fail because BM25 cannot resolve date expressions'') and predicts which failure category should decrease.
    \item \textbf{Diagnosis}: AI agents run the current harness, produce per-question diagnostic traces, and the Layer-A classifier routes each failure to the earliest responsible stage---ingestion, retrieval, filtering, or generation---when evidence annotations and source-id links are available.
    \item \textbf{Patch}: The engineer (or an AI agent) implements a bounded fix scoped to the hypothesized stage (e.g., adding date-aware BM25 tokenization), keeping the change within one module boundary.
    \item \textbf{Verification}: The analyzer aligns baseline and candidate outputs by question key, partitions samples into improved / regressed / both-wrong / both-correct, and applies the configured gate. If the hypothesis prediction matches the observed category movement and no category regresses beyond threshold, the patch is accepted.
\end{enumerate}

\paragraph{Concrete example.}
In one representative round on the LoCoMo development trace, the Layer-A classifier attributed 30\% of remaining failures to \texttt{ingestion\_gap}---cases where relevant dialogue turns were never extracted into L0 memory. The hypothesis was that long conversations ($>$50 turns) exceeded the single-pass extraction context budget, silently dropping later evidence. The fix---adding the supplement extraction path with topic segmentation and overlapping windows---reduced \texttt{ingestion\_gap} from 30\% to 2\% of residual failures and yielded a cumulative +4.48pp accuracy gain. The paired gate reported per-category non-regression, and rejected alternative directions (aggressive prompt expansion, threshold lowering) were archived as negative priors for future \EMEND search.

\paragraph{Verification infrastructure.}
\label{sec:dacci-infra}
Given baseline and candidate outputs aligned by question key, the analyzer partitions samples into improved, regressed, both-wrong, and both-correct sets. A candidate first passes a bounded probe; when the run reports sufficient paired evidence, the full gate can include McNemar-style checks, bootstrap intervals, and per-category non-regression gates. When those statistical artifacts are not reported, the result is used only as descriptive development evidence. Fixes should be implemented in shared modules whenever possible, enabling a unified framework without benchmark-specific forks. Extended formulas, trace schemas, and representative iteration rounds are provided in Appendix~\ref{app:dacci-extended}.

\paragraph{Empirical validation.}
The representative development trace (Table~\ref{tab:dacci-rounds} in Appendix~\ref{app:dacci-extended}) indicates controlled improvement: most accepted gains came from fixing a localized failure surface, not adding broad prompt complexity; rejected directions were archived as negative priors for later automated search.

\paragraph{Limitations.}
\label{sec:dacci-failures}
\DACCI is most reliable when failures are observable at pipeline stages and intermediate artifacts are recorded. Its claims are scoped to stage-local diagnosis and human-approved hypotheses; it does not claim convergence or optimality. These limits motivate \EMEND: bounded, reversible strategy search under the same evidence and acceptance constraints, while framework or protocol changes remain in the human-governed \DACCI loop.

\subsubsection{\EMEND: Automated Strategy Evolution}

\EMEND implements the automation side of the dual-loop framework. It does not treat the memory system as an unconstrained program-synthesis target. Its admissible search space is a \emph{schema-constrained declarative strategy space} with three parts: a schema-level contract for admissible fields, value ranges, evolvable dimensions, and cross-field invariants; strategy artifacts that instantiate concrete policies for ingestion, retrieval, lifecycle management, presentation, and feature toggles; and symptom-triggered candidate bundles that map observed failure modes to interpretable policy moves. Prompts are evolvable only when represented as versioned, rollback-capable strategy artifacts. Implicit prompt edits embedded in framework code are treated as framework changes governed by \DACCI, not strategy mutations handled by \EMEND.

\paragraph{Base harness and strategy space.}
The locked base harness contains components retained from repeated \DACCI iterations: cross-encoder rerank, intent parsing/query decomposition, layered retrieval, and raw-dialogue backfill. These components remain in the framework core, reducing uncontrolled regression risk while preserving a fixed development baseline. Within that boundary, \EMEND can evolve only declared strategy fields:
\begin{itemize}
    \item Extraction: versioned prompt templates, max facts per trajectory, dedup threshold.
    \item Retrieval: top\_k, time-decay half\_life, BM25/vector weight ratio, rerank top\_n, intent overrides.
    \item Lifecycle and presentation: pruning score, compression policy, source\_window size, output format.
    \item Advanced features: failure correction, conflict handling, conditional memory triggers.
\end{itemize}

\paragraph{Observe--Improve--Verify Loop.}

Each \EMEND evolution round uses a strategy artifact, benchmark tasks, and the environment configuration as inputs. A \textbf{Bootstrap Runtime} normalizes tasks, builds the evaluator, and initializes the EvolutionEngine. The loop then proceeds in three phases: Observe evaluates the current strategy and assigns stage-local root causes; Improve mutates only declared strategy dimensions; and Verify applies scope checks, Gate/Critic validation, targeted screens, paired comparisons, and rollback safeguards before accepting any serving update. The full per-round pseudocode and Layer~A decision tree are kept in Appendix~\ref{app:emend-artifacts}.

The three-phase design keeps deterministic checks and gate validation on the critical path while confining exploratory mutations to the bounded YAML strategy space. In the implementation, rollback is triggered when current accuracy falls more than 3pp below the all-time best accepted strategy, separating candidate rejection from serving-strategy drift. Layer~A routes failures into ingestion, retrieval, or generation gaps only when complete evidence annotations and source-id links are available; in deployment-like settings, the same logic is heuristic and audit-assisted, not a deterministic earliest-loss claim. The Critic adjusts strictness by actionability, and an optional UCB1 hypothesis tree can prioritize future mutation directions. These implementation-level rules are detailed in Appendix~\ref{app:emend-artifacts}.

\paragraph{Safeguards and rollback.}

\EMEND searches a textual strategy space, not a continuous parameter space. It faces risks analogous to RL policy optimization---for example optimizing a proxy metric, over-exploiting one dimension, or accumulating drift---but this report does not claim to prove RL-style reward hacking, policy collapse, or catastrophic forgetting in this setting. RL governance is therefore used only as an analogy for a three-layer defense (Table~\ref{tab:three-layer-defense}).

\begin{table}[htbp]
\centering
\caption{Evolution safeguards.}
\label{tab:three-layer-defense}
\compacttable
\begin{tabularx}{\textwidth}{l>{\hsize=0.9\hsize}X >{\hsize=0.9\hsize}X >{\hsize=1.2\hsize}X}
\toprule
\textbf{Layer} & \textbf{Mechanism} & \textbf{Analogy} & \textbf{Risk Addressed} \\
\midrule
Hard constraint & 5-Gate fail-fast pipeline & Action mask & Schema loopholes \\
Soft review & Reputation-driven Critic & Independent value estimate & Overuse of one mutable dimension \\
Cross-round safeguard & Best-ever rollback & Checkpoint restore & Accepted-strategy drift \\
\bottomrule
\end{tabularx}
\end{table}

Every candidate passes through a fail-fast gate pipeline followed by an independent Critic. The gates enforce schema legality, novelty, value ranges, prompt integrity, and replay stability; the Critic adds reputation-driven soft review before expensive evaluation. Rollback then operates at two levels:
\begin{itemize}
    \item \textbf{Per-round revert}: If the best candidate's delta falls below threshold, the strategy is not updated. The failure is logged to \path|dimension_effects|, increasing revert count and degrading that dimension's reputation for future Critic checks.
    \item \textbf{Best-ever auto-restore}: Every round persists the all-time best strategy and accuracy. If current accuracy degrades by $>3$pp below best-ever, the system restores from \texttt{best\_strategy.json}---regardless of whether the current round accepted or rejected a candidate. This bounds accumulated drift across many rounds.
\end{itemize}

\noindent Detailed gate definitions and Critic heuristics are provided in Appendix~\ref{app:emend-artifacts}. Eight cross-round persistent stores carry journals, dimension effects, progress scoreboards, question histories, candidate outcomes, stage audits, Layer-A logs, and hypothesis-tree state. Together, they provide cross-round state for the Replay gate, Critic reputation, optional UCB1 direction search, and rollback safeguard.

\subsection{Validation: Strategy Improvement {under} Bounded Automation}

The evaluated subset covers the implemented parts of the dual-loop design: Framework/Strategy separation, Layer~A diagnosis, Digester--Planner--Evolver proposal, Gate/Critic validation, targeted screening, accept/revert control flow, best-ever rollback, and cross-round persistent stores. Remaining roadmap items are deferred to Section~\ref{sec:conclusion}. The claim boundary is limited to \emph{bounded offline strategy automation} inside a locked harness; arbitrary memory-system evolution and online deployment are not evaluated here.

\paragraph{Experimental configuration.}
The run keeps the same role-separated offline configuration across stages: serving-side extraction/routing, evolution agents, and QA judging remain fixed, with full assignments listed in Appendix~\ref{app:emend-artifacts}. The \MemStack LoCoMo number in Section~\ref{sec:memorystack-results} is the earlier controlled-reference anchor for the base memory system. The \EMEND number reported here is a subsequent staged result under the same offline harness; it is evidence of governed strategy improvement over time, not a direct leaderboard comparison between independently timed runs.

On LoCoMo, \EMEND improves the initial memory-system baseline from \textbf{75.58\%} (1164/1540) to the reported best checkpoint of \textbf{94.74\%} (1459/1540), a +19.16pp gain and +295 net correct answers. This is descriptive full-benchmark evidence for the staged strategy loop, not a repeated-run statistical claim.

The improvement was staged rather than the result of a single prompt rewrite. Table~\ref{tab:emend-locomo-compressed} groups the accepted trajectory into external-facing strategy families; representative accepted, rejected, and no-effect candidate outcomes are preserved in Appendix~\ref{app:emend-artifacts}. Reproducibility is artifact-level: each \EMEND result is tied to a fixed benchmark, baseline, strategy artifact mutation, candidate trace, and gate record; exact replay may depend on model and judge endpoints.

\begin{table}[htbp]
\centering
\caption{Compressed \EMEND strategy checkpoints on LoCoMo.}
\label{tab:emend-locomo-compressed}
\compacttable
\begin{tabularx}{\textwidth}{lXrrr}
\toprule
\textbf{Strategy family} & \textbf{External-facing description} & \textbf{Stage} & \textbf{Acc.} & \textbf{Gain} \\
\midrule
Initial memory strategy & Baseline before staged \EMEND evolution & S0 & 75.58 & -- \\
Evidence-grounded context enrichment & Attach local dialogue evidence and recover inventory-style facts & S1 & 81.82 & +6.23 \\
Structured state abstraction & Convert evidence into typed temporal, location, plan, preference, slot, and ledger state & S2 & 87.08 & +11.49 \\
Verified candidate adoption & Adopt only candidates that pass support, shape, and non-regression checks & S3 & 91.49 & +15.91 \\
Deterministic support-cluster decision policy & Cluster equivalent verified candidates and choose the best-supported answer & S4 & 93.70 & +17.86 \\
Evidence-supported candidate consolidation & Consolidate supported candidate answers while excluding low-confidence candidate classes & S5 & 94.74 & +19.16 \\
\bottomrule
\end{tabularx}
\end{table}
\FloatBarrier

The largest absolute gains occur in single-hop (+139 correct), temporal (+69), and multi-hop (+70) questions. Inference remains the most difficult bucket, but it still improves by +17 correct. Beyond the score movement, the validation question is whether strategy updates are accepted, rejected, and archived under stable rules. The appendix trace indicates that \EMEND did not monotonically accept every proposal: self-correction and broad raw-dialogue fallback were rejected as regressive or unstable, while evidence admission, typed-state abstraction, verified candidate adoption, deterministic support clustering, and evidence-supported consolidation formed the accepted path.

\EMEND studies \emph{how} memory strategies can be changed under evidence constraints. Together with \DACCI, it instantiates the diagnostic-evolution category in Figure~\ref{fig:overview}: the evidence-governed \DACCI loop defines human-reviewed capability changes, while the bounded \EMEND loop searches the declarative strategy space under those fixed rules. This division of labor aligns with the independently proposed SSGM framework~\cite{ssgm2026}, which argues that memory evolution and governance should be decoupled to reduce drift and poisoning risk in deployed agents. In this section, ``decoupled'' means that the two loops have different responsibilities and implementation paths while sharing one audit contract. The resulting audit contract is explicit: a memory-system change must be represented as an inspectable artifact, evaluated under fixed governance, and then accepted, rejected, or rolled back with preserved evidence.

\paragraph{Takeaway.}
\DACCI keeps diagnosis human-auditable, while \EMEND automates only bounded strategy search under a locked framework. The objective is to make memory evolution reproducible, comparable, and reversible under one governance contract.

\paragraph{Future: Online Incremental Evolution.} This capability is not evaluated here. Production environments lack ground-truth labels, so online evolution should add narrowly scoped auxiliary prompts or conditional rules, trigger them only on matching data, track isolated impact, and merge or disable them independently.

%% ---------------------------------------------------------------------------
\section{Deployment: Lightweight Memory with \LiteMem}
\label{sec:light}

The \textbf{Deployment Transfer} requirement asks whether the same audit contract can survive when memory moves from service-specific stores into local repository-backed state. \LiteMem is used here as a substrate-transfer probe, not a complete replacement for the service-side stack. The substrate uses Markdown/YAML records, file-tool retrieval, and Git provenance, so continuity, editability, rollback, and bounded retrieval remain inspectable under local-first constraints. This positioning connects to repository-native and local-first memory tools, with related motivation from direct corpus interaction, Git/version-controlled context, and lightweight memory-augmented generation~\cite{dci2026,byterover2026,gitofthoughts2026,gcc2026,superlocalmemory2026}.

\ataglance{RQ4 --- Deployment Transfer}{\LiteMem repository-native substrate test}{Markdown/YAML state, file-tool retrieval, Git provenance, and file-level audit traces}{LoCoMo-aligned transfer-feasibility evidence}

%% ---------------------------------------------------------------------------
\subsection{Design Motivation: Memory {under} Deployment Constraints}
\label{sec:light-question}

Personal AI should not assume one cloud-hosted vector-store runtime. Cost, latency, privacy, offline operation, and user editability all push memory toward local or edge substrates. \LiteMem asks a narrower transfer question: can a repository-native implementation retain a substantial fraction of the measured memory benefit while preserving the lifecycle framework's audit obligations? The retained improvement is reported against the same no-memory baseline, with formulas and local decay settings in Appendix~\ref{app:light-extended}. This is transfer evidence within a LoCoMo-aligned setting, not a production-scale edge deployment claim.

%% ---------------------------------------------------------------------------

\subsection{Method: Repository-Native Audit Substrate}
\label{sec:light-schema}

\LiteMem models memory as repository state, not a specialized service. Its design target is to keep recall, capture, organization, edit history, and rollback visible through standard file and version-control operations. Markdown/YAML files provide persistent state, file search provides the initial retrieval path, and Git history provides the provenance and rollback layer. If Git is unavailable, the system still exposes inspectable files, but provenance and rollback support is more limited.

\begin{figure}[t]
\centering
\includegraphics[width=0.99\textwidth]{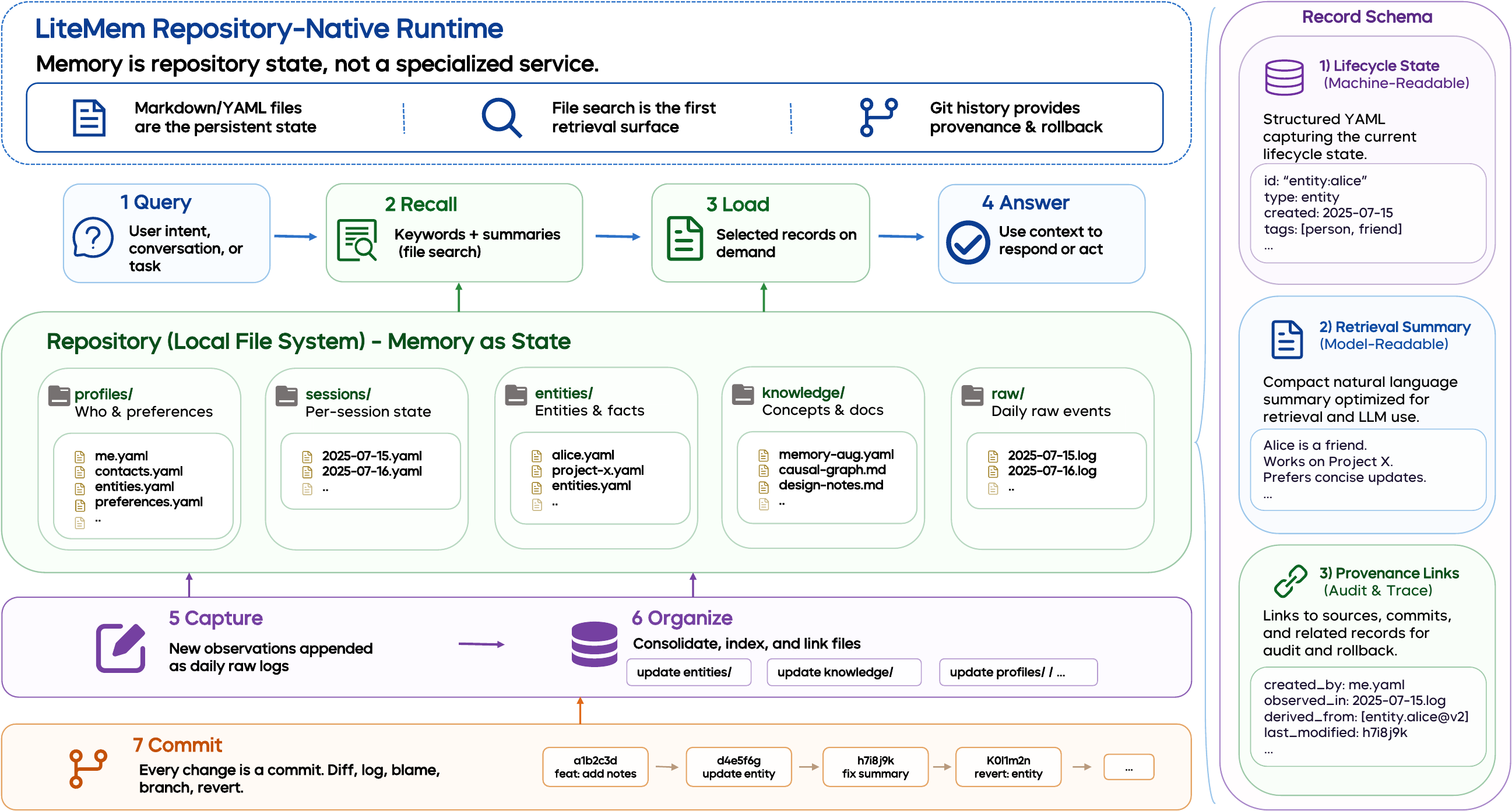}
\caption{\LiteMem recall path.}
\label{fig:litemem-runtime}
\end{figure}

Each record exposes three views: a machine-readable lifecycle state, a model-readable retrieval summary, and provenance links for audit. Figure~\ref{fig:litemem-runtime} illustrates how these views participate in a repository-native lifecycle: query, recall, lazy loading, answer or action, capture, organization, and Git-backed history. The concrete directory hierarchy and YAML fields are implementation details reported in Appendix~\ref{app:light-extended}. The claim tested here is that L0/L1/L2/SM-style information can map to local profile, session, entity, knowledge, and raw daily-event files without being hidden inside a database.

\begin{table}[H]
\centering
\caption{\LiteMem audit surfaces.}
\label{tab:light-surfaces}
\compacttable
\begin{tabularx}{\textwidth}{lXX}
\toprule
\textbf{Surface} & \textbf{Representation} & \textbf{Main role} \\
\midrule
Summary files & title, summary, keywords, and frontmatter metadata & first-pass retrieval and ranking \\
Raw daily logs & append-only Markdown event records & repair path for missed writes \\
Profile/entity/knowledge files & consolidated long-term state and procedures & stable user state, organization output, and task routines \\
Git history & diffs, commits, and rollback traces & provenance, supersession, and debuggability \\
\bottomrule
\end{tabularx}
\end{table}

\paragraph{Repository surfaces in one view.}
The lightweight substrate is structured, not flat. In the evaluated design, one memory often maps to one Markdown file or one section inside a consolidated file; the repository then organizes four broader audit surfaces: compact summary files for first-pass retrieval, append-only daily logs for repairable capture, consolidated profile/entity/knowledge files for stable state, and Git history for provenance and rollback. This separation makes the deployment-transfer claim inspectable: the system can search, inspect, edit, and audit memory with standard developer tools, not hidden service metadata.

Appendix~\ref{app:light-extended} provides the schema details. Conceptually, the repository carries lifecycle responsibilities that a service-side memory stack would normally distribute across a database, index, and audit service.

\paragraph{Retrieval scoring and temporal decay.}
\label{sec:light-scoring}
Candidates are ranked by lexical relevance, current importance, recency ($\Delta t$ since last access, timescale $\tau$), and a short-term access-feedback signal. The scoring function uses a memory record $m$, the current query $q_t$, and the current turn $t$ as explicit arguments:
\begin{equation}
\mathrm{Score}(m, q_t, t) = \lambda_1 \mathrm{Lex}(m,q_t) + \lambda_2 \mathrm{Imp}_t(m) + \lambda_3 \exp(-\Delta t / \tau) + \lambda_4 \mathrm{Access}(m).
\label{eq:light-rank}
\end{equation}
Here $\mathrm{Lex}(m,q_t)$ is the lexical-match score, $\mathrm{Imp}_t(m)$ is the current importance state, and $\mathrm{Access}(m)$ is a normalized access-feedback feature. Idle organization recomputes the importance state (Eq.~\ref{eq:light-decay}) and modulates it with user feedback ($\eta$, $\rho$ are boost/penalty coefficients):
\begin{equation}
\mathrm{Imp}_t(m) = \mathrm{Imp}_0(m) \cdot \exp(-\Delta t_m / \tau_m) + \eta \cdot \mathrm{access\_count}(m) - \rho \cdot \mathrm{skip\_count}(m).
\label{eq:light-decay}
\end{equation}
The two decay terms use different time bases: in Eq.~\ref{eq:light-rank}, $\Delta t$ is the interval since the record was last accessed (retrieval recency); in Eq.~\ref{eq:light-decay}, $\Delta t_m$ is the interval since the record's importance was last assigned or refreshed (organization-time salience decay). The corresponding timescales $\tau$ and $\tau_m$ reflect this difference: $\tau$ is short (query-time ranking) while $\tau_m$ is longer (background organization).

\paragraph{Progressive disclosure and lifecycle.}
\label{sec:light-progressive}
\LiteMem uses \emph{progressive disclosure}: retrieve compact summaries ranked by Eq.~\ref{eq:light-rank}, select top-$k$ candidates, and lazy-load full text only for selected records. The lifecycle has three flows: recall before answering, raw capture after interaction, and idle or session-boundary organization. The append-only daily log preserves a repair path when structured consolidation is incomplete. Details are in Appendix~\ref{app:light-extended}.

\paragraph{Running trace: Ethan's training handoff.}
In the running scenario (Section~\ref{sec:running-example}), \LiteMem scans compact summary files for ``training'', ``bag'', and ``equipment'', then lazily expands the matching daily log and entity file. A recent correction---spare shoes kept at school---is visible in Git history, which records when it entered the repository, while the diff identifies which earlier statement it superseded. After the interaction, the new outcome is appended to the daily log and later consolidated into the entity file, exercising all four deployment surfaces: summary recall, raw-event repair, consolidated state, and repository provenance.

%% ---------------------------------------------------------------------------
\subsection{Validation: Substrate Transfer}
\label{sec:light-implementation}

For validation, the repository-native design is instantiated in a local agent runtime and evaluated with a modular TypeScript harness. The harness uses matched prompts across full-history, keyword retrieval, deterministic extraction, LLM extraction, and agent-flow variants. Implementation details such as atomic writes, temporal simulation, speaker attribution, and organization actions are summarized in Appendix~\ref{app:light-extended}.

By separating retrieval, extraction, and flow-level choices under the same substrate, the harness keeps the result interpretable as audit-substrate transfer rather than a prompt-template or serialization artifact.

\paragraph{Experimental results.}
\label{sec:light-results}
Table~\ref{tab:light-results} reports transfer-feasibility evidence under a LoCoMo-aligned file-tool setting. \LiteMem reaches 90.81\%, retaining 90.0\% of the service-side improvement over the same no-memory baseline, with a 97.0\% absolute score ratio against the full \MemStack reference. The 90.0\% figure is computed relative to the same no-memory baseline: $(90.81-65.83)/(93.59-65.83)=90.0\%$. The claim is substrate-transfer evidence within the aligned setting, not a leaderboard submission or production cost/latency curve.

\begin{table}[H]
\centering
\caption{\LiteMem transfer results: multi-dimensional comparison.}
\label{tab:light-results}
\compacttable
\begin{tabularx}{\textwidth}{@{}l c c c c X@{}}
\toprule
\textbf{System} & \textbf{Accuracy} & \textbf{Infra. cost} & \textbf{Substrate} & \textbf{Latency} & \textbf{Trade-off summary} \\
\midrule
\MemStack (full) & \textbf{93.59\%} & High & Cloud service & Moderate & Vector DB + embedding + reranker; full diagnostic surface \\
\rowcolor{hxOrange!12}
\LiteMem & 90.81\% & None & Local Markdown/Git & Low & 90.0\% retention; no external dependency; fully offline \\
No-memory baseline & 65.83\% & None & Full-history prompt & High & Long-context only; no memory organization \\
Raw chunk-retrieval & 55.28\% & None & Naive file-split & Low & No semantic structure; lowest recall in this comparison \\
\bottomrule
\end{tabularx}
\end{table}

The category profile clarifies where transfer holds: direct recall and temporal localization remain relatively strong, while multi-hop relation traversal is the main remaining limitation. The result supports transfer of audit discipline---representation, retrieval discipline, provenance, and organization policy---not equivalence to a full service-side memory stack.

\paragraph{Takeaway.}
\LiteMem exposes two deployment boundaries. First, repository-native recall is limited by lexical search and lightweight organization unless optional local indexes or compact graph files are added. Second, long-horizon file-count scaling and latency remain unvalidated beyond the current transfer-feasibility setting. Within these boundaries, \LiteMem makes the deployment side of the audit contract concrete: in the evaluated repository-native setting, memory stays inspectable, editable, and provenance-aware even when specialized memory infrastructure is removed.

%% ---------------------------------------------------------------------------
\section{Evidence-Governed Claims Across the Lifecycle}
\label{sec:evaluation}

This section is a claim ledger rather than a benchmark summary. It asks whether the same evidence-governed audit contract remains coherent across Structure, Expansion, Evolution, and Deployment. Each role contributes one evidence anchor with a stated maturity level and claim boundary.

\begin{seednote}{Evidence-governed audit contract}
A memory result is admissible when it states what evidence entered the system, how that evidence was represented, which retrieval or governance decision used it, which trace makes the decision inspectable, and which claim boundary limits over-generalization.
\end{seednote}

\paragraph{Why these modules belong together.}
\MemStack tests durable state; \MemSense/\MemFuse test multimodal and cross-device evidence admission; \DACCI/\EMEND test governed memory-policy change; and \LiteMem tests substrate transfer. The modules are heterogeneous by design, but they expose the same artifact families: typed evidence, provenance, diagnostic traces, gated updates, and deployment boundaries.

%% ---------------------------------------------------------------------------
\subsection{Evidence-Governed Evaluation}

Because \MiMemory studies a lifecycle framework, the shared protocol evaluates audit obligations, not a universal metric. Each track states its run setting, admissible evidence and provenance, available traces, governed-change criterion, and explicit claim boundary. Missing artifacts or preliminary tracks remain marked at lower evidence levels.

\paragraph{Evidence-level taxonomy.}
The report uses \emph{evidence level} for evaluation maturity and \emph{statistical qualifier} for repeated-run, confidence-interval, or paired-test support. Unless such support is explicitly reported, numerical differences are descriptive rather than significance claims. The levels used here are: \emph{controlled reference} (\MemStack vs.\ reproduced MemBrain), \emph{module-level public or aligned} (\MemSense), \emph{controlled offline benchmark} (\EMEND), \emph{transfer-feasibility} (\LiteMem), \emph{preliminary/internal} (\MemFuse), and \emph{design-only} (optional procedural hooks). Run configurations and resource accounting are collected in Appendix~\ref{app:evaluation-protocol-details}.

%% ---------------------------------------------------------------------------
\paragraph{Consolidated evidence across lifecycle roles.}

Each lifecycle role contributes one bounded evidence anchor. For \textbf{Structure}, \MemStack reaches 93.59\% / 57.24\% / 87.47\% on LoCoMo / PersonaMem-V2 / LongMemEval under a controlled-reference comparison with a reproduced MemBrain baseline; the +0.34pp LoCoMo margin is descriptive parity. For \textbf{Expansion}, \MemSense reports 89.15\% Mem-Gallery accuracy, while \MemFuse reaches 35.2\% on the internal MemFuseBench (+4.7pp over mem0@20), framed as preliminary cross-device fusion evidence. For \textbf{Evolution}, \DACCI/\EMEND improve a subsequent locked LoCoMo offline run from 75.58\% to 94.74\% through traceable, gate-bounded strategy changes; this is not compared directly with the earlier \MemStack anchor. For \textbf{Deployment}, \LiteMem demonstrates 90.81\% transfer score and 90.0\% retention, establishing transfer feasibility rather than a scaling claim.

The results are role-specific, not a cross-role ranking. Their shared invariant is the audit contract: typed evidence, provenance, and an explicit boundary against over-generalization.

%% ---------------------------------------------------------------------------
\subsection{Cross-Module Findings and Limitations}
\label{sec:eval-process}

The lifecycle view surfaces interactions that would be hidden in a module-only report:
\begin{itemize}
    \item \textbf{\EMEND + \MemStack}: \EMEND can tune retrieval strategy inside a locked \MemStack framework because the runtime exposes strategy artifacts and diagnostic traces instead of hard-coded patches.
    \item \textbf{\MemFuse + continuity substrate}: \MemFuse outputs typed cross-device payloads that can be admitted into the same downstream memory contract used by \MemStack and optional procedural hooks.
    \item \textbf{Source-filtering tradeoff}: source-side filtering can improve dialogue QA while risking the loss of low-similarity causal edges, so causal payloads require separate retention and evaluation criteria.
\end{itemize}
These compatibility observations are not a substitute for joint ablation.

The current evidence supports contract compatibility and calibrated per-module progress; marginal contribution still requires a joint ablation benchmark.

\paragraph{Limitations and pending experiments.}
\label{sec:eval-cost}

\begin{seednote}{Safety and privacy boundary}
The audit contract makes memory behavior inspectable; privacy enforcement remains a platform-level layer. Deletion propagation, cross-user leakage, memory poisoning, Git-history exposure, and visual sensitive-information handling require platform-level access control, consent management, and adversarial validation beyond the module-level evidence reported here.
\end{seednote}

The benchmarks exercise recall fidelity, visual grounding, cross-device causality, bounded strategy evolution, and repository-native transfer in separate tracks. They do not yet provide a closed-loop personal AI benchmark where memory guides downstream actions~\cite{memoryarena2026,worldmemarena2026}, nor do they test extreme-scale context volumes~\cite{beam2026}. Full end-to-end ablation, year-scale deployment, and platform-level privacy validation remain roadmap items.

%% ═══════════════════════════════════════════════════════════════════════════════
\section{Conclusion and Outlook}
\label{sec:conclusion}

This report presents \MiMemory as a lifecycle memory framework for personal AI. The central conclusion is that assistant memory should be treated as inspectable infrastructure rather than a retrieval layer alone: durable state, multimodal evidence, governance, and deployment constraints must remain jointly auditable.

\MiMemory operationalizes this view through four roles. \MemStack anchors Structure, \MemSense/\MemFuse support Expansion, \DACCI/\EMEND govern Evolution, and \LiteMem tests Deployment under a repository-native substrate. Their claims are tied to traces and paired comparisons when available, and otherwise bounded by the evidence-level taxonomy in Section~\ref{sec:evaluation}.

%% ---------------------------------------------------------------------------
\subsection*{Evidence-Bounded Takeaways}

\begin{itemize}
    \item \textbf{Memory as infrastructure}: Assistant memory is treated as a governed substrate for continuity, personalization, and cross-device grounding rather than a retrieval layer alone.
    \item \textbf{Audit contract as integration primitive}: Typed evidence payloads, stage-local diagnostics, strategy artifacts, and rollback/gate records keep heterogeneous modules aligned while preserving each track's claim boundary.
    \item \textbf{Evidence-bounded anchors}: Structure, Evidence Admission, Evolution, and Deployment each contribute calibrated evidence rather than a single aggregate leaderboard.
    \item \textbf{Traceable improvement}: Representative diagnostics, rejected directions, and gate decisions are preserved as auditable development artifacts.
\end{itemize}

%% ---------------------------------------------------------------------------
The next practical step is to turn these artifact families into operational infrastructure for telemetry, privacy controls, editability, rollback, and reproducible ablation.

\subsection*{Research Outlook}

The four lifecycle gaps identified in Section~\ref{sec:introduction}---state, source, governance, and deployment---are partially addressed here, but none is closed by a single component. Closing them requires co-design across algorithms, shared evaluation contracts, and product-level validation. Five directions are most immediate.

\paragraph{Causal memory attribution.}
The open problem is determining the minimal evidence set that entails a given answer and verifying that no intermediate stage introduced unsupported content. This specializes faithfulness verification to multi-granularity stores where the same fact may exist as an L0 atomic fact, an L1 summary, and an L2 profile entry. The product target is user-facing provenance and operator debugging: when a user asks ``why did you say that?'', the system should trace the answer to specific memories and retrieval decisions.

\paragraph{Propagation-complete forgetting.}
Current forget-guard mechanisms do not guarantee that a revoked fact ceases to influence downstream summaries, profiles, or cached retrievals. The challenge is to invalidate or re-derive all dependent representations when evidence is deleted---this intersects with machine unlearning~\cite{bourtoule2021machine} but operates over structured memory, not parametric weights. The product requirement is demonstrable compliance through stage-local diagnostic traces.

\paragraph{Federated memory composition across devices.}
The research problem is maintaining a coherent user model when phones, cars, wearables, and homes contribute evidence under different retention, privacy, and sharing policies without one centralized store. This extends federated learning ideas to structured memory state and connects to policy-aware synchronization, conflict resolution, and trust-boundary enforcement.

\paragraph{Diagnostic-driven evolution across deployments.}
Scaling governed memory evolution from offline benchmark stages to production traffic requires online monitoring without ground-truth labels, automated hypothesis generation from failure clusters, and guarded incremental deployment with category-sensitive rollback safeguards. These problems connect to safe reinforcement learning and automated ML pipeline management, specialized to memory systems where improving one segment can silently regress another.

\paragraph{Standardized memory contracts.}
The community lacks a shared interface contract---analogous to function-calling schemas or tool-use protocols---for cross-system ablation, portable benchmarks, and composable evolution across heterogeneous memory architectures. The research question is what minimal contract can support diverse systems while staying constrained enough for meaningful comparison and governed interoperability.

\paragraph{Long-term convergence.}
The directions above converge on governed, multi-device, self-diagnosing assistant-memory infrastructure. The central open problem is making the contract \emph{expressive} across sources and modalities, \emph{auditable} from research through production, and \emph{transferable} across cloud, edge, and local substrates. No single module in this report achieves all three; the four lifecycle roles instead break the target into testable pieces. Long term, deployed observations should generate hypotheses, controlled tests should validate them under the shared contract, accepted changes should ship behind gates, and deployment outcomes should feed the next cycle of structure refinement, source expansion, and governance improvement.

\FloatBarrier
\bibliographystyle{plainnat}
\clearpage
\bibliography{references}
\clearpage
%% ═══════════════════════════════════════════════════════════════════════════════
\phantomsection
\section*{Contributions and Acknowledgments}
\label{sec:contributions-acknowledgments}
\addcontentsline{toc}{section}{Contributions and Acknowledgments}

\noindent
\begin{minipage}[t]{0.47\textwidth}
\paragraph{Core Contributors.}
\begin{itemize}
    \item Xule Liu$^{*}$
    \item Hanlin Teng$^{*}$
    \item Chao Li$^{*}$
    \item Yanan Ni
    \item Kun Shao$^{\dagger}$
    \item Jian Luan$^{\dagger}$
\end{itemize}
\end{minipage}\hfill
\begin{minipage}[t]{0.47\textwidth}
\paragraph{Contributors.}
\begin{itemize}
    \item Shuo Lu
    \item Audrey Wang
    \item Yijun Liu
    \item Yunfei Wang
    \item Xiaofeng Li
    \item Xian Yi
    \item Yuanfa Li
    \item Kang Zhao
    \item Jian Liang
    \item Yuxuan Chen
    \item Jinyuan Chen
    \item Heng Qu
\end{itemize}
\end{minipage}

\par\smallskip
{\footnotesize $^{*}$ Equal contribution. \textsuperscript{\textdagger} Corresponding author.}

\clearpage
%% Appendix material restored after main-text compression
\clearpage
\beginappendix
\appendix
% Keep the main table of contents focused on the pre-appendix narrative: show
% one Appendix entry, then suppress the detailed appendix section list.
\phantomsection
\addcontentsline{toc}{section}{Appendix}
\addtocontents{toc}{\protect\setcounter{tocdepth}{-1}}

% ── Appendix-wide: compact font and tighter float spacing ──
\small
\setlength{\textfloatsep}{6pt plus 2pt minus 2pt}
\setlength{\floatsep}{5pt plus 2pt minus 2pt}
\setlength{\intextsep}{5pt plus 2pt minus 2pt}

\noindent The appendix collects notation, extended module details, artifacts, and evaluation protocol details referenced in the main text. It is organized as follows:

\begin{itemize}[nosep,leftmargin=12pt]
\item \textbf{Section~A} --- Notation and symbols (principal symbols by lifecycle role)
\item \textbf{Section~B} --- \MemStack extended details (terminology, interfaces, retrieval, QA prompt, design principles, optional hooks)
\item \textbf{Section~C} --- Evidence Admission extended artifacts (\MemFuse ingestion and session graph)
\item \textbf{Section~D} --- \MemSense extended artifacts (IKB passes, answer algorithm, progress)
\item \textbf{Section~E} --- \DACCI extended artifacts (iteration rounds, verification infrastructure, case studies)
\item \textbf{Section~F} --- \EMEND evolution artifacts (loop algorithm, strategy schema, gate/critic rules, UCB1)
\item \textbf{Section~G} --- \LiteMem extended artifacts (schema, hierarchy, ranking equations, lifecycle algorithm)
\item \textbf{Section~H} --- Evaluation protocol details (claim checks, root-cause summary, cost)
\end{itemize}

\section{Notation and Symbols}
\label{app:notation}

Table~\ref{tab:notation} summarizes the principal symbols used throughout the report.

\begin{table}[H]
\centering
\caption{Principal notation.}
\label{tab:notation}
\scriptsize
\setlength{\tabcolsep}{3pt}
\renewcommand{\arraystretch}{1.05}
\begin{tabularx}{\textwidth}{@{}l l X@{}}
\toprule
\textbf{Symbol} & \textbf{Introduced} & \textbf{Meaning} \\
\midrule
\multicolumn{3}{@{}l}{\textit{Overview / lifecycle contract (Section~\ref{sec:overview})}} \\
$q_t$ & \S\ref{sec:overview} & User request at turn $t$ \\
$O_t$ & \S\ref{sec:overview} & Typed source-observation pool available at serving time \\
$M_t$ & \S\ref{sec:overview} & Memory state before serving at turn $t$ \\
$E_{q_t}$ & \S\ref{sec:overview} & Evidence selected for request $q_t$ \\
$C_{q_t}$ & \S\ref{sec:overview} & Bounded context assembled for generation \\
$a_t$ & \S\ref{sec:overview} & Generated answer at turn $t$ \\
$M_{t+1}$ & \S\ref{sec:overview} & Memory state after improvement-loop update \\
\midrule
\multicolumn{3}{@{}l}{\textit{Structure / \MemStack (Section~\ref{sec:memorystack})}} \\
$L_{0,t}, L_{1,t}, L_{2,t}, SM_t$ & \S\ref{sec:memorystack} & Layerwise realization of $M_t$: atomic facts, session summaries, profile constraints, and session memory \\
$o_t$ & \S\ref{sec:memorystack} & Admitted observation at turn $t$ \\
$q_t$ & \S\ref{sec:memorystack} & User request at turn $t$ (same as overview $q_t$) \\
$W$ & \S\ref{sec:memorystack} & Layer-aware write operator \\
$R_{\mathrm{sem}}, R_{\mathrm{lex}}, R_{\mathrm{exp}}$ & \S\ref{sec:memorystack} & Semantic, lexical, and expanded retrieval channels \\
$\hat{E}_{q_t}$ & \S\ref{sec:memorystack} & Combined retrieval candidates before budgeting \\
$A$ & \S\ref{sec:memorystack} & Assembly function \\
$B$ & \S\ref{sec:memorystack} & Context budget \\
$U$ & \S\ref{sec:memorystack} & Update operator \\
$c_t$ & \S\ref{sec:memorystack} & Correction arriving after generation \\
\midrule
\multicolumn{3}{@{}l}{\textit{Expansion / Source Layer (Section~\ref{sec:source})}} \\
$O$ & \S\ref{sec:source} & Typed observation universe (text, images, device events) \\
$G$ & \S\ref{sec:source} & Cross-device memory graph \\
$K$ & \S\ref{sec:source} & Image Knowledge Base (IKB) \\
$\lambda$ & \S\ref{sec:source} & Traceability weight \\
$e$ & \S\ref{sec:source} & Individual candidate evidence item \\
$\operatorname{rel}(e,q_t)$ & \S\ref{sec:source} & Relevance of evidence $e$ to query $q_t$ \\
$\operatorname{tok}(e)$ & \S\ref{sec:source} & Token cost of evidence $e$ \\
$\operatorname{prov}(e)$ & \S\ref{sec:source} & Provenance score of evidence $e$ \\
$\mathit{Rel}_j$ & \S\ref{sec:source} & Related L0 facts for image $x_j$ in IKB (avoids overloading $R$) \\
$e_t$ & \S\ref{sec:source} & Incoming event at turn $t$ \\
$\mathcal{N}_t$ & \S\ref{sec:source} & Retrieved neighborhood for fusion \\
\midrule
\multicolumn{3}{@{}l}{\textit{Deployment / \LiteMem (Section~\ref{sec:light})}} \\
$\mathrm{Score}(m,q_t,t)$ & \S\ref{sec:light} & Retrieval scoring function \\
$\lambda_1\text{--}\lambda_4$ & \S\ref{sec:light} & Component weights (lexical, importance, recency, access) \\
$\mathrm{Imp}_t(m)$ & \S\ref{sec:light} & Importance with decay at turn $t$ \\
$\Delta t$ & \S\ref{sec:light} & Interval since record was last accessed (retrieval recency) \\
$\Delta t_m$ & \S\ref{sec:light} & Interval since importance was last assigned (organization decay) \\
$\tau, \tau_m$ & \S\ref{sec:light} & Recency decay timescales ($\tau$ short, $\tau_m$ long) \\
$\eta, \rho$ & \S\ref{sec:light} & Access-boost and skip-penalty coefficients \\
\bottomrule
\end{tabularx}
\end{table}

\section{\MemStack Extended Details}
\label{app:memorystack-extended}

This section provides the \MemStack details used to interpret and reproduce the reported runs: terminology, typed interfaces, retrieval formula, context-budget policy, design principles, and the optional procedural-hook boundary.

\paragraph{Terminology.} Table~\ref{tab:terminology} defines the terms used across the framework. These definitions are shared by all modules; later sections reference them without redefining them.

\begin{table}[H]
\centering
\caption{Terminology.}
\label{tab:terminology}
\scriptsize
\setlength{\tabcolsep}{4pt}
\renewcommand{\arraystretch}{0.92}
\begin{tabularx}{\textwidth}{p{0.22\textwidth}X}
\toprule
\textbf{Term} & \textbf{Definition} \\
\midrule
Memory item & A typed stored unit in \MemStack, including L0 facts, L1 summaries, L2 profile entries, or optional ProcedureEntry payloads. \\
Evidence & Source observations or memory items that support a particular answer, diagnosis, or strategy decision, together with provenance and trace identifiers. \\
Policy & The operational behavior governing ingestion, retrieval, filtering, context construction, and generation support. \\
Strategy artifact & A declarative, versioned policy artifact consumed by \EMEND; it changes parameters, prompts, and routing rules without modifying framework code. \\
Strategy space & The schema-constrained set of admissible strategy artifacts and candidate mutations searched by \EMEND. \\
Human hypothesis & A \DACCI iteration proposal written by an engineer to explain a failure mode and predict how a bounded intervention should affect metrics. \\
Harness & A fixed evaluation environment containing dataset slice, prompts, judge, scoring script, random seeds, and acceptance criteria. \\
Provider & An implementation backend for storage, retrieval, model calling, judging, or source ingestion behind a stable interface. \\
Optional procedural guidance & A typed trigger--procedure payload attached to \MemStack context assembly when enabled. \\
FusionSession & A \MemFuse grouping of cross-device events with three transient zones whose accepted edges are persisted after zone eviction. \\
MemoryPack & A \MemFuse output payload that summarizes an activity-level fused node while retaining links to atomic events, devices, timestamps, and causal/semantic edges. \\
Diagnostic trace & A JSONL-style per-question artifact recording retrieved candidates, filters, ranks, context packing, answer evidence, and failure attribution. \\
Gate & A deterministic or critic-assisted acceptance check that rejects schema-violating, regressive, or non-replayable strategy changes before serving-strategy reload. \\
\bottomrule
\end{tabularx}
\end{table}

\paragraph{Typed interfaces.} Table~\ref{tab:interface-schema} specifies the five payload types that cross module boundaries. Each row states the required fields, producer--consumer pair, current implementation status, and failure-handling policy.

\begin{table}[H]
\centering
\caption{Interface excerpt.}
\label{tab:interface-schema}
\compacttable
\begin{tabularx}{\textwidth}{@{}>{\raggedright\arraybackslash}p{0.14\textwidth}>{\raggedright\arraybackslash}p{0.30\textwidth}>{\raggedright\arraybackslash}p{0.14\textwidth}>{\raggedright\arraybackslash}p{0.15\textwidth}>{\raggedright\arraybackslash}X@{}}
\toprule
\textbf{Payload} & \textbf{Required fields} & \textbf{Producer $\rightarrow$ consumer} & \textbf{Implemented / evidence status} & \textbf{Failure handling} \\
\midrule
\texttt{FusedEvent} & \texttt{id}, \texttt{source\_event\_ids}, \texttt{timestamp}, \texttt{device\_ids}, \texttt{edge\_type}, \texttt{provenance}, \texttt{confidence} & \MemFuse $\rightarrow$ \MemStack & Prototype path used by internal MemFuseBench evidence & Keep atomic events if fusion is low-confidence or contradictory \\
\texttt{PerceptionFact} & \texttt{fact\_id}, \texttt{image\_id}, \texttt{session/date}, \texttt{category/name}, \texttt{caption}, \texttt{confidence}, \texttt{source\_ids} & \MemSense $\rightarrow$ \MemStack & Used by module-level Mem-Gallery run & Fall back to image id when names/categories are uncertain \\
\texttt{ProcedureEntry} & \texttt{entry\_id}, \texttt{trigger}, \texttt{procedure}, \texttt{constraints}, \texttt{validation}, \texttt{status} & \MemStack $\rightarrow$ Context Assembly & Design-only; internal to \MemStack & Operational guidance only \\
\texttt{DiagnosticSignal} & \texttt{question\_id}, \texttt{retrieved\_ids}, \texttt{context\_ids}, \texttt{answer\_evidence}, \texttt{stage\_label}, \texttt{error\_label} & \MemStack $\rightarrow$ \EMEND & Used when evidence annotations exist & Mark stage label uncertain when source ids are incomplete \\
\texttt{StrategyArtifact} & \texttt{artifact\_id}, \texttt{schema\_version}, \texttt{mutation\_paths}, \texttt{guardrails}, \texttt{gate\_record}, \texttt{rollback\_key} & \EMEND $\rightarrow$ \MemStack & Used by the staged \EMEND LoCoMo run & Reject if schema or non-regression gates fail; restore on rollback \\
\bottomrule
\end{tabularx}
\end{table}

\paragraph{Safety and privacy gaps.} Table~\ref{tab:risk-gap} maps known risk surfaces to the mechanism that partially addresses them and the residual gap that remains outside the scope of module-level evidence.

\begin{table}[H]
\centering
\caption{Safety/privacy gaps.}
\label{tab:risk-gap}
\scriptsize
\setlength{\tabcolsep}{3pt}
\renewcommand{\arraystretch}{0.92}
\begin{tabularx}{\textwidth}{p{0.16\textwidth}p{0.28\textwidth}p{0.24\textwidth}X}
\toprule
\textbf{Risk} & \textbf{Mechanism} & \textbf{Evidence} & \textbf{Gap} \\
\midrule
Stale/harmful memory & Provenance, forget/update constraints, invalidation & \MemStack schema + traces & User-facing editing policy pending \\
Cross-device linkage & Source provenance, visibility metadata, typed payloads & \MemFuse interface design & Consent/ACL are platform-level \\
Strategy drift & Schema-constrained artifacts, gate checks, rollback & \EMEND dev-slice trace & Online evolution without labels \\
Conflict/mis-fusion & Atomic-event provenance, fused-node edges & Preliminary MemFuseBench & Conflict arbitration unresolved \\
Local exposure & Markdown/Git boundary, local-file execution & \LiteMem transfer run & OS permissions, encryption \\
\bottomrule
\end{tabularx}
\end{table}

\paragraph{Capability map and interfaces.} Tables~\ref{tab:capability-map} and~\ref{tab:interfaces} trace capabilities from lifecycle stages to modules, then detail the inter-module triggers and payloads.

\begin{table}[H]
\centering
\caption{Capability map.}
\label{tab:capability-map}
\compacttable
\begin{tabularx}{\textwidth}{lXl}
\toprule
\textbf{Stage} & \textbf{Core Capability} & \textbf{Module} \\
\midrule
Sources & Dialogue, device events, visual evidence & \MemFuse + \MemSense \\
Admission and ingestion & Source normalization, segmentation, extraction, consolidation & \MemSense + \MemFuse + \MemStack \\
Representation & L0/L1/L2 facts, optional procedural hooks, fused nodes & \MemStack \\
Retrieval & Hybrid search, rerank, graph expansion & \MemStack \\
Context assembly & Budgeted packing, deduplication, safety constraints & \MemStack \\
Feedback \& update & User correction, decay, invalidation, Dreaming/idle consolidation & \MemStack + \DACCI \\
Governed evolution & Diagnostics, strategy search, gated strategy reload & \DACCI + \EMEND \\
Deployment & Repository-native transfer feasibility, privacy-sensitive mode & \LiteMem \\
\bottomrule
\end{tabularx}
\end{table}

\begin{table}[H]
\centering
\caption{Interfaces and triggers.}
\label{tab:interfaces}
\compacttable
\begin{tabularx}{\textwidth}{ll l l X}
\toprule
\textbf{Source} & \textbf{Target} & \textbf{Interface} & \textbf{Trigger} & \textbf{Payload} \\
\midrule
\MemFuse & \MemStack & \texttt{FusedEvent[]} & Device event processed & Causal events + device provenance \\
\MemSense & \MemStack & \texttt{PerceptionFact[]} & Perception pipeline completed & Structured visual facts + confidence \\
\MemStack\textsuperscript{$\dagger$} & Context Assembly & \texttt{ProcedureEntry[]} & Intent / pattern match & Optional user-specific procedural guidance \\
\MemStack & \EMEND & \texttt{DiagnosticSignal} & Diagnostic logging & Coverage/rank/error traces \\
\EMEND & \MemStack & \texttt{StrategyArtifact} & Gate approval & Updated declarative config \\
\bottomrule
\end{tabularx}
\par\smallskip{\scriptsize $\dagger$\,The optional procedural hook is internal to \MemStack; it is design-only and listed separately because it has a distinct output interface to Context Assembly.}
\end{table}

\paragraph{Three-channel retrieval.}
Given query $q_t$ and memory store $M_t$, the runtime combines semantic vector search, lexical BM25 matching, and LLM-generated subquery expansion. Let $L_c(q_t)$ be the ranked list returned by channel $c\in\{\mathrm{vec},\mathrm{bm25},\mathrm{subq}\}$ and $\operatorname{rank}_c(m)$ the one-based rank of memory $m$ in that list. Candidate lists are fused by weighted Reciprocal Rank Fusion,
\begin{equation}
\mathrm{RRF}(m) = \sum_{c \in \{\mathrm{vec},\mathrm{bm25},\mathrm{subq}\}} \frac{w_c\,\mathbf{1}[m\in L_c(q_t)]}{\kappa + \operatorname{rank}_c(m)}.
\label{eq:app-rrf}
\end{equation}
Unless overridden by a benchmark adapter, $\kappa$ follows the common RRF default of 60, channel weights start uniform, and the final candidate pool is truncated after deduplication and optional reranking. LLM subquery expansion is bounded by a fixed configuration value (three to five subqueries in the reported runs) and must preserve explicit entities, dates, and negations from the original query. The exact prompt is treated as a versioned strategy artifact when evolved by \EMEND; changing it inside framework code is a \DACCI-level change. CombMNZ and Borda are retained as ablation alternatives but are not used for the main reported path. Table~\ref{tab:app-retrieval-params} lists the retrieval parameters exposed as configuration together with their claim boundaries.

\begin{table}[H]
\centering
\caption{Retrieval parameters.}
\label{tab:app-retrieval-params}
\compacttable
\begin{tabularx}{\textwidth}{lXX}
\toprule
\textbf{Parameter} & \textbf{Role} & \textbf{Claim boundary} \\
\midrule
$\kappa$ & Smooths RRF rank contribution; default 60 unless adapter overrides & Not tuned per question \\
$w_{\mathrm{vec}}, w_{\mathrm{bm25}}, w_{\mathrm{subq}}$ & Channel weights for semantic, lexical, and subquery paths & Uniform baseline; evolvable only as declared strategy \\
\texttt{subquery.max\_n} & Caps LLM query decompositions & Small bounded value; prompt changes require artifact versioning \\
\texttt{rerank.top\_n} & Limits expensive reranking after RRF & Configuration knob, recorded in trace \\
\texttt{dedup.threshold} & Prevents near-duplicate L0/L1 evidence from consuming budget & Validated by paired regression gates \\
\bottomrule
\end{tabularx}
\end{table}

\paragraph{Context budget assembly.}
Context assembly reserves profile evidence, treats forget/update constraints as high-priority governance constraints, and fills remaining context by fused rank after deduplication. A typical packing order is shown in Table~\ref{tab:app-context-budget}. The percentages are defaults used to explain the policy, not universal constants: each run records the realized token counts so a regression can be attributed to retrieval, filtering, or packing.

\begin{table}[H]
\centering
\caption{Context budget.}
\label{tab:app-context-budget}
\compacttable
\begin{tabularx}{\textwidth}{lXX}
\toprule
\textbf{Slot} & \textbf{Default policy} & \textbf{Overflow behavior} \\
\midrule
System/task instructions & Fixed prefix & Not counted as memory evidence \\
Forget/update constraints & Mandatory, budget-exempt in priority & Reduce lower-ranked L0/L1 before dropping constraints \\
L2 profile / style & Reserved high-priority slice (typically 10--20\%) & Summarize profile, keep provenance pointer \\
SM current session & Recent unresolved turns and constraints & Window by recency and unresolved state \\
L0/L1 evidence & Pack by fused rank after deduplication & Emit overflow trace with dropped ids \\
Source payloads & IKB / \texttt{FusedEvent} evidence when routed & Exempt from broad cosine pruning; still token-bounded \\
\bottomrule
\end{tabularx}
\end{table}

\paragraph{Assembly and adapters.}
Benchmark-specific logic remains in adapters, while ingestion, storage, retrieval, assembly, and diagnostics stay in the shared kernel.

\paragraph{Representative QA system prompt.}
\label{app:qa-prompt}
This prompt governs answer generation in the QA evaluation path. At the generation stage, it instantiates the audit contract by requiring explicit evidence enumeration from memory layers, provenance-chain resolution, temporal grounding, contradiction arbitration via correction history, and sufficiency verification before committing to a final answer. The structured step format allows diagnostic traces to attribute failures to specific reasoning stages rather than opaque generation.

\begin{promptfile}{prompts/qa\_system.txt}
You are an evidence-governed memory assistant. You answer
questions by retrieving and reasoning over structured
personal memory organized into layers: L0 (atomic facts),
L1 (session/topic summaries), L2 (stable profile and
corrections), and SM (current session context).
Today's date is {today_date}.

# EVIDENCE FIDELITY REQUIREMENTS
1. Preserve all named entities verbatim -- use full
   identifiers (e.g. "Amy's colleague Rob"), never
   generic references
2. Retain exact quantities: numbers, prices, dates,
   times, percentages, frequencies
3. Maintain temporal specificity -- "every Tuesday and
   Thursday", not "twice a week"
4. When multiple records describe similar events,
   use timestamps and layer metadata to distinguish them
5. Perform inference only when evidence from multiple
   layers strongly supports the connection

# STRUCTURED REASONING PATH

## Step 1: EVIDENCE CANDIDATES
Enumerate all memory records (L0 facts, L1 summaries,
L2 profile entries, SM context) that could relate to
the question. Include records with unresolved relative
dates -- do not filter prematurely.

## Step 2: LAYER-AWARE DETAIL EXTRACTION
For each candidate, extract: source layer, record
timestamp, entity names, quantities, and provenance
links to other records.

## Step 3: PROVENANCE CHAIN RESOLUTION
Trace connections across layers and records:
- L0-to-L2 promotion: if an atomic fact establishes
  a stable attribute, link to the profile entry
- Cross-entity inference: properties of linked entities
  (family members, colleagues) may imply attributes
- Collective references: resolve "they/we/together"
  to specific entities via session context

## Step 4: TEMPORAL GROUNDING
Resolve relative time expressions using record metadata:
- "yesterday" from a record dated 2023-08-25 means
  2023-08-24
- Use [recorded:] and [updated:] timestamps for
  ordering; [previously:] marks superseded versions
- Never assume today's date unless the record
  explicitly states "today" or "just now"
- Only use dates explicitly present in retrieved
  evidence for temporal arithmetic

## Step 5: CONTRADICTION ARBITRATION
When records conflict on the same attribute, apply
the correction-history rule: the most recently updated
L2 entry or the latest L0 observation takes precedence.
Flag the conflict source in your reasoning.

## Step 6: EVIDENCE-QUESTION ALIGNMENT
- Re-read the question and identify the specific event,
  time, or context it targets
- For each detail in your answer, verify it originates
  from the SAME event/context the question asks about
- When multiple similar records exist, match question
  constraints to the correct record explicitly

## Step 7: SUFFICIENCY VERIFICATION
If no single record contains the answer, synthesize
across layers. Commit to the strongest supported
inference -- only respond "insufficient evidence" when
no record is even tangentially related.

## FINAL ANSWER
State the answer directly and concisely first, then
add supporting provenance. Do not hedge when evidence
is available.

FINAL ANSWER: <your answer>
\end{promptfile}

\paragraph{Core design principles.}
\label{app:design-principles}
The following principles were refined through the \DACCI development history and guide module-level design decisions. Each principle is an engineering constraint supported by the available paired evaluations; violations surface as regressions in the corresponding module's evaluation. Table~\ref{tab:principles} summarizes each principle together with its concrete embodiment in the system.

\begin{table}[H]
\centering
\caption{Design principles.}
\label{tab:principles}
\compacttable
\begin{tabularx}{\textwidth}{lXX}
\toprule
\textbf{Principle} & \textbf{Meaning} & \textbf{Embodiment} \\
\midrule
Typed boundaries & Modules exchange explicit payloads, not internal state & \texttt{FusedEvent} / \texttt{PerceptionFact} / \texttt{DiagnosticSignal} \\
Decoupled evolution & Stable framework locked, variable strategies declarative & Framework/Strategy separation \\
Progressive disclosure & On-demand expansion, no full-history preload & L0/L1/L2 + optional hooks + lazy cascade \\
Causal multi-source & Causal evidence complements semantic similarity & \MemFuse dual-layer graph \\
Evidence-governed improvement & Every change needs paired evidence and bounded regressions & Diagnostic trace + gate decision \\
Edge transferability & Core memory principles can be simplified for local deployment & \LiteMem transfer evidence \\
\bottomrule
\end{tabularx}
\end{table}

\subsection{Optional procedural hooks inside \MemStack}
\label{app:memorystack-hooks}

\MemStack can store a compact procedural guidance entry when repeated user feedback shows that factual recall alone is not enough. The hook was introduced during \DACCI iterations after repeated errors exposed a distinct failure pattern: the system could remember the relevant facts but still fail to adapt its response procedure to a user's recurring feedback, routines, or conversational expectations. These failures motivate a memory object that stores \emph{how the assistant should interact with this user in a recurring situation}, not only what the user said. Unlike general-purpose agent skill memory or tool-skill libraries, the hook is scoped to personalized conversational behavior and remains internal to \MemStack.

A procedural entry is represented as a compact contract:
\begin{equation}
\mathrm{ProcedureEntry} = (\mathrm{trig}, p, c, v),
\end{equation}
where $\mathrm{trig}$ is the trigger condition, $p$ is an ordered conversational procedure, $c$ is the constraint set, and $v$ is the validation rule applied before the final answer or experiment report. Triggers decide when the procedural entry is relevant, procedures encode reusable user-specific response steps, constraints record what must not be violated, and validation rules convert the entry into an executable checklist. This differs from preference memory: a preference states what the user wants, while the procedural entry states how the assistant should reliably respond to that user's recurring task or situation.

At runtime, procedural entries are retrieved only after the topic, entity, temporal, and confidence gates pass. The gated entry is injected as operational guidance only, keeping procedural rules separate from answer content so diagnostics can distinguish factual failures from interaction-procedure failures. The hook is therefore framed as a design outcome of human-guided diagnostic iteration, not a public benchmark track. Limited internal diagnostic cases are used only to check the mechanism; large-scale cross-domain evaluation and ablation are left to future work.

\section{Extended Evidence Admission Artifacts}
\label{app:source-breadth-extended}

\begin{figure}[H]
\centering
\begin{minipage}[t]{0.48\textwidth}
\centering
\includegraphics[width=\textwidth]{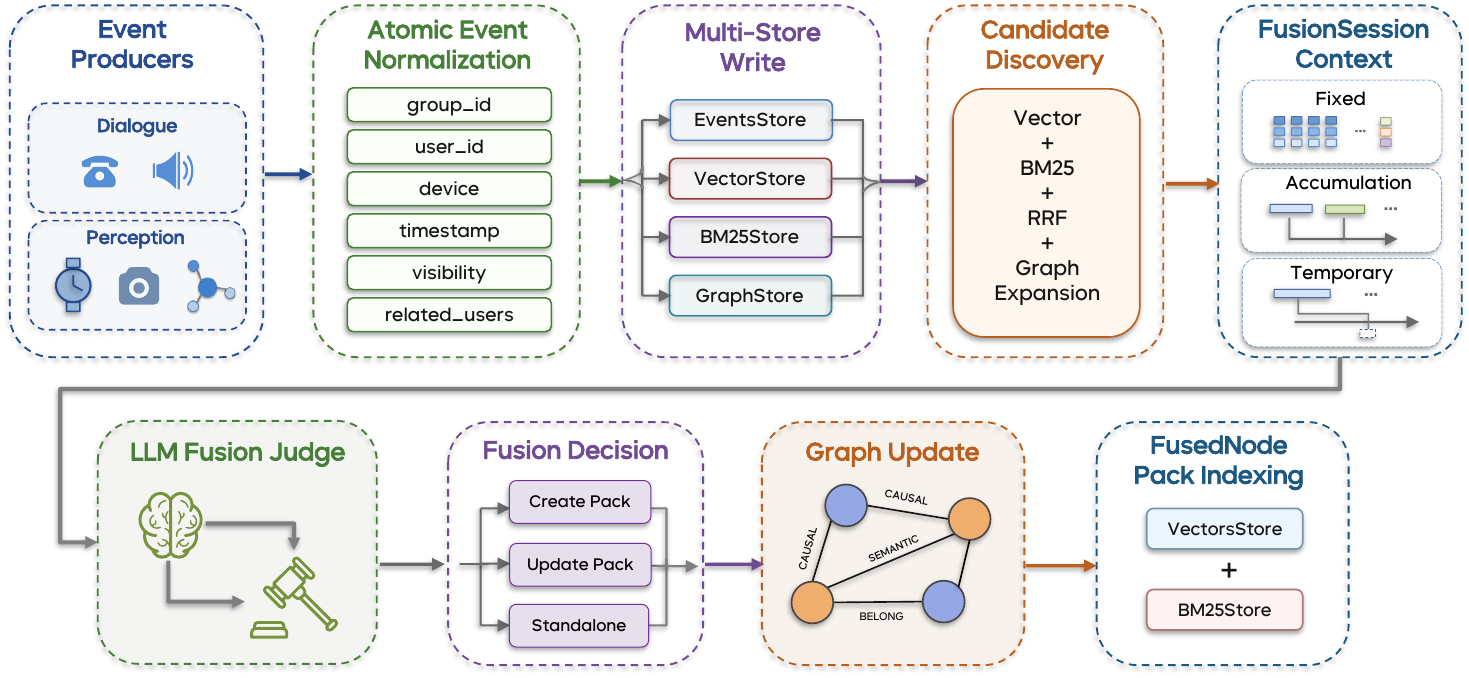}
\caption{\MemFuse ingestion path.}
\label{fig:memfuse-ingestion}
\end{minipage}\hfill
\begin{minipage}[t]{0.48\textwidth}
\centering
\includegraphics[width=\textwidth]{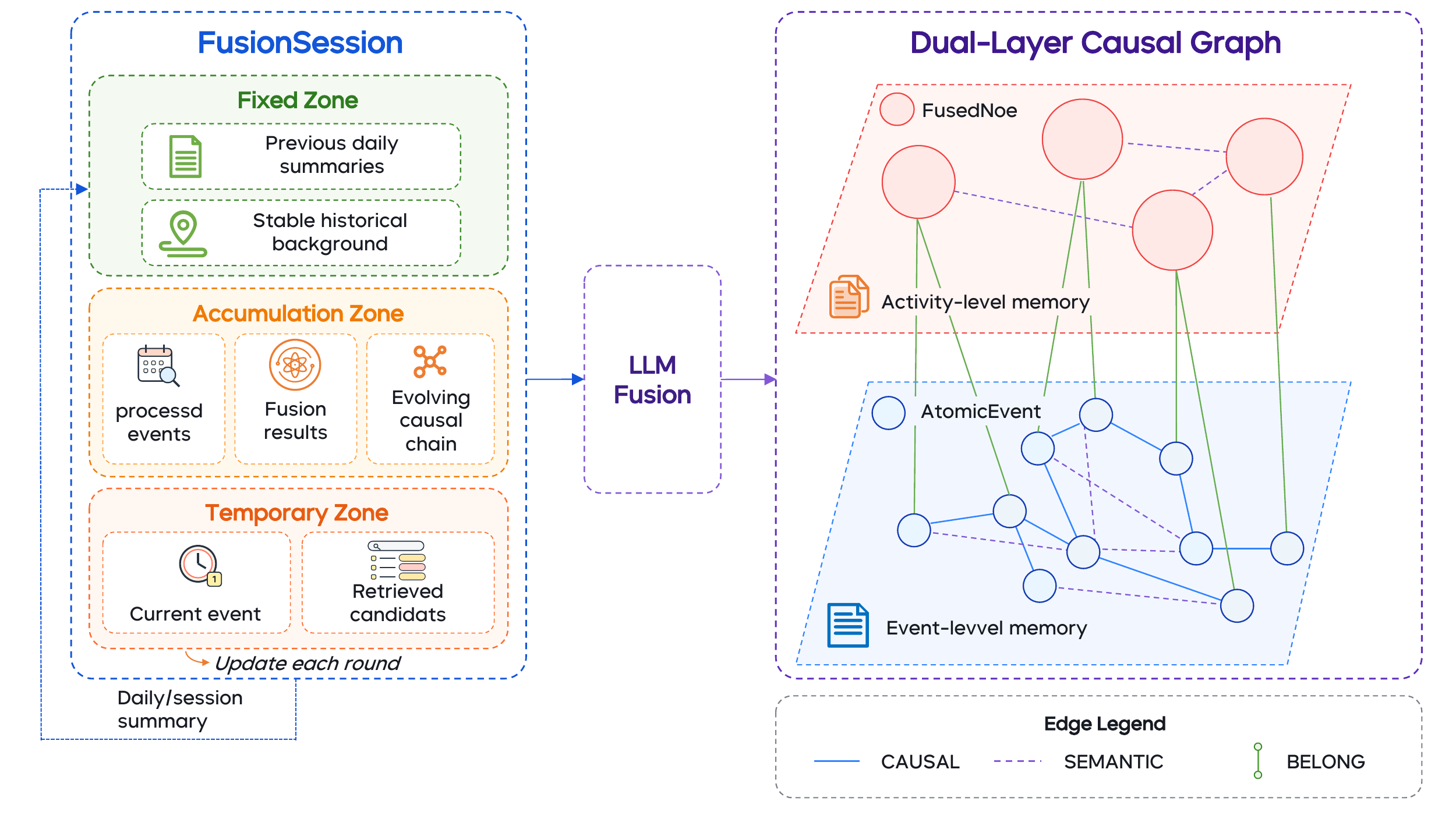}
\caption{\MemFuse session graph.}
\label{fig:memfuse-session-graph}
\end{minipage}
\end{figure}

This section adds implementation-level \MemFuse diagrams and FusionSession mechanics to the Evidence Admission chapter. Figure~\ref{fig:memfuse-ingestion} illustrates device-specific payload normalization before fusion, so later retrieval can operate over fused nodes rather than raw sensor formats alone. Figure~\ref{fig:memfuse-session-graph} motivates the dual representation in \MemFuse: atomic nodes preserve provenance, while fused nodes provide the activity-level memory used by downstream retrieval.

\section{Extended \MemSense Artifacts}
\label{app:mm-extended}

This section adds the IKB construction pipeline (Table~\ref{tab:app-ikb-passes}), the IKB-first answer algorithm (Algorithm~\ref{alg:app-mm-answer}), and implementation status with remaining gaps (Table~\ref{tab:app-mm-progress}) for the multimodal-memory branch. The five-pass construction in Table~\ref{tab:app-ikb-passes} allows Algorithm~\ref{alg:app-mm-answer} to prefer symbolic image knowledge over top-$k$ visual retrieval.

\begin{table}[htbp]
\centering
\caption{IKB build passes.}
\label{tab:app-ikb-passes}
\compacttable
\begin{tabularx}{\textwidth}{lXX}
\toprule
\textbf{Pass} & \textbf{Operation} & \textbf{Failure Addressed} \\
\midrule
1 & Per-image conversation-name extraction with preceding turn context & Resolves references such as ``another one'' or assistant-provided species names \\
2 & Session-level batch naming for unnamed images & Recovers images whose local turn lacks enough context \\
3 & Synonym / variant normalization & Merges names such as ``frangipani'' and ``plumeria'' when they denote the same object \\
4 & Similarity-check split for large groups & Over-merged categories that inflate counts \\
5 & Registry backfill & Missing fallback IKB entries for registered images \\
\bottomrule
\end{tabularx}
\end{table}

The staged build directly affects visual counting and enumeration questions, where a missing registry entry or over-merged synonym can change the final answer.

Algorithm~\ref{alg:app-mm-answer} specifies how the constructed IKB is used at answer time. The algorithm functions as a routing rule: standard text memory remains available, but VR, VS, and TTL cases receive deterministic image-knowledge injections before generation.

\begin{algorithm}[htbp]
\DontPrintSemicolon
\SetAlgoLined
\KwIn{Question $q$, optional question image $x_q$, mem0 store $M$, RAGAnything store $R$, IKB $K$}
\KwOut{Answer $a$}
\BlankLine
Classify $q$ into intent and retrieval strategy\;
Retrieve text memories from $M$ using the selected strategy\;
Run session filter/rerank over Session Memory candidates\;
Retrieve graph chunks and image candidates from $R$ using text--text, text--image, and image--image paths\;
Fuse L0/L1/L2, session context, graph chunks, entities, relations, and reasoning paths\;
\If{$q$ is VR counting and $K$ matches a category}{
  inject count and image ID evidence from $K$ when available; constrain output format to number only\;
}
\ElseIf{$q$ is VS enumeration or time-filtered selection}{
  inject candidate image list from $K$ using category/session/temporal indexes\;
}
\ElseIf{$q$ is TTL with question image}{
  match $x_q$ against conversation images; inject top conversation names from $K$
}
Generate final answer with text context and at most eight retrieved images\;
\caption{IKB-first answer path.}
\label{alg:app-mm-answer}
\end{algorithm}

This algorithm links construction to evaluation: the remaining errors in Table~\ref{tab:app-mm-progress} are mostly trigger, normalization, and VLM-override issues, not missing support for multimodal retrieval itself.

Table~\ref{tab:app-mm-progress} summarizes implementation status so that remaining engineering gaps are not confused with completed benchmark claims.

\begin{table}[htbp]
\centering
\caption{Multimodal progress.}
\label{tab:app-mm-progress}
\compacttable
\begin{tabularx}{\textwidth}{lXX}
\toprule
\textbf{Component} & \textbf{Implementation Status} & \textbf{Open Gap} \\
\midrule
IKB build & Five-pass construction, temporal index, related-fact binding & Synonym normalization can over-merge fine-grained categories \\
VR counting & IKB count query with format override and registry cross-check & Fuzzy aggregation can inflate counts; needs stricter category confidence \\
VS enumeration & IKB-based candidate-list path exists for ``find all'' patterns & Triggering remains sensitive; some VS questions still rely on image-vector top-$k$ \\
TTL naming & Candidate conversation-name injection exists; image-match helper exists & Deterministic image matching is not yet first-class in the main TTL path \\
RAGAnything retrieval & Text--text, text--image, image--image, graph PPR, reasoning paths & Cross-session visual recall can truncate before IKB enumeration is applied \\
Answer formatting & Category-specific format instructions and JSON/no-JSON routing & VLM can override textual constraints for visually ambiguous images \\
Knowledge filter & Optional evidence filter and AR (Association/Reasoning) post-processing & Disabled by default; AR is a local error-analysis tag for associative or reasoning hallucinations, not a fourth visual category alongside VR/VS/TTL \\
\bottomrule
\end{tabularx}
\end{table}

The table pairs each implemented component with its remaining gap, keeping the appendix evidence conservative: it documents what is already implemented in the branch and what still needs robustness work.

\section{Extended \DACCI Artifacts}
\label{app:dacci-extended}

This section summarizes the \DACCI evidence trail: paired comparison sets, failure-case traces, root-cause labels, patch diffs, statistical reports, experiment logs, and representative accepted/rejected rounds. It provides methodology evidence, not a benchmark result.

\paragraph{Diagnostic trace and constrained optimization.}
Let $\mathcal{S}=\{s_i\}_{i=1}^N$ be an evaluation set and $\mathcal{P}_\theta$ a memory configuration. For each sample, the shared kernel emits a diagnostic trace
\begin{equation}
d_i = \big(e_i^{\text{raw}}, r_i^{\text{top-k}}, f_i^{\text{filter}}, c_i^{\text{ctx}}, g_i^{\text{gen}}, j_i\big),
\label{eq:diagnostic-trace}
\end{equation}
covering ingestion, retrieval, filtering, context assembly, generation, and judging. The objective is constrained improvement:
\begin{equation}
\theta^* = \arg\max_{\theta \in \Theta} \mathrm{Acc}(\theta)
\quad \text{s.t.}\quad
\forall c \in \mathcal{C}: \Delta_c(\theta,\theta_{\text{base}}) \geq -\epsilon_c,
\label{eq:constrained-opt}
\end{equation}
where $\mathcal{C}$ is the category set and $\epsilon_c$ bounds acceptable regression.

\paragraph{Evidence-preservation ladder.}
Stage attribution follows the earliest point where ground-truth evidence $e_i^*$ is lost:
\begin{equation}
\mathrm{FailType}(s_i) = \begin{cases}
\texttt{ingestion\_gap} & \text{if } e_i^* \notin \mathrm{MemStore}(\theta) \\
\texttt{retrieval\_gap} & \text{if } e_i^* \in \mathrm{MemStore} \setminus r_i^{\text{top-k}} \\
\texttt{kf\_filtered} & \text{if } e_i^* \in r_i^{\text{top-k}} \setminus c_i^{\text{ctx}} \\
\texttt{generation\_error} & \text{if } e_i^* \in c_i^{\text{ctx}} \text{ but } J(\hat{y}_i,y_i^*)=0.
\end{cases}
\label{eq:failtype}
\end{equation}
The \texttt{kf\_filtered} label refers to evidence removed by the knowledge filter during context assembly.

\paragraph{Representative iteration rounds.}
Table~\ref{tab:dacci-rounds} provides the roadmap for this appendix: it lists which iterations are discussed later, why each change was accepted or rejected, and how the diagnostic labels connect the raw traces to the reported score movement.

\begin{table}[htbp]
\centering
\caption{\DACCI rounds.}
\label{tab:dacci-rounds}
\compacttable
\begin{tabularx}{\textwidth}{l X l X r r l}
\toprule
\textbf{Round} & \textbf{Hypothesis / intervention} & \textbf{Root cause} & \textbf{Evidence gate} & \textbf{Score} & \textbf{$\Delta$} & \textbf{Decision} \\
\midrule
$t_0$ & Mem0-style pipeline & --- & establishes aligned baseline & 83.07 & --- & baseline \\
$t_1$ & CoT 7-step + time-aware prompt & generation & paired judge gain without category regression & 88.83 & +5.76 & accept \\
$t_2$ & Ingestion ordering + update fix & ingestion & missing-evidence traces recovered & 90.70 & +1.87 & accept \\
$t_3$ & Dedup threshold 0.75 & over-merge & regression exceeds non-regression gate & 88.25 & $-$2.45 & revert \\
$t_4$ & Threshold 0.92 + topic segmentation & ingestion & improved coverage with bounded merge errors & 92.73 & +4.48 & accept \\
$t_5$ & Option-aware KF (PersonaMem) & filter & cross-benchmark regression signal & $-$1pp & $-$1.0 & diagnostic flag \\
$t_6$ & BM25/RRF + Agentic + unified core & retrieval & shared-core gain across diagnostics & \textbf{93.59} & +0.86 & accept \\
\bottomrule
\end{tabularx}
\end{table}

The table highlights that most accepted gains came from fixing a localized failure surface rather than adding broad prompt complexity; the rejected rows are included because they became negative priors for later automated search.

The full internal development log contains more than 30 iterations. Table~\ref{tab:dacci-rounds} lists only representative rounds with clear evidence artifacts; unlisted rounds are not used as independent empirical claims. Operationally, the timeline alternated among three states: accepted fixes that moved the shared kernel forward, rejected or reverted probes that became negative priors, and pending hypotheses that lacked enough paired evidence for acceptance. This distinction avoids reading the iteration history as monotonic hill-climbing or as evidence that every diagnostic idea worked. A visual timeline is omitted because the discrete three-state model (accept/revert/pending) and the table's root-cause labels provide an auditable record that a linear diagram would flatten into a misleading monotonic curve.

\subsection{Verification Infrastructure}

\paragraph{Paired comparison algebra.}
All acceptance decisions operate on aligned question keys rather than aggregate scores. Given baseline output set $Y_{\text{base}}$ and candidate output set $Y_{\text{cand}}$ aligned by identifier, the analyzer computes:
\begin{align}
I &= \{i : J_{\text{base}}(i) = 0 \wedge J_{\text{cand}}(i) = 1\} && \text{(improved)} \\
R &= \{i : J_{\text{base}}(i) = 1 \wedge J_{\text{cand}}(i) = 0\} && \text{(regressed)} \\
B_w &= \{i : J_{\text{base}}(i) = 0 \wedge J_{\text{cand}}(i) = 0\} && \text{(both-wrong)} \\
B_c &= \{i : J_{\text{base}}(i) = 1 \wedge J_{\text{cand}}(i) = 1\} && \text{(both-correct)}
\end{align}
A candidate is accepted only if $|I| > |R|$ and the applicable gate is satisfied. When statistical artifacts are reported, this includes a significance check; otherwise the result remains descriptive.

\paragraph{Statistical gates.}
The two-stage gate structure is intended to reduce false-positive acceptance while maintaining iteration speed:
\begin{itemize}
    \item \textbf{Stage-1 (probe):} Run on $|\mathcal{S}_{\text{probe}}| = 50$--100 samples. Reject if $\mathrm{Acc}(\theta') < \mathrm{Acc}(\theta_t) - \delta_1$ (typically $\delta_1 = 2\text{pp}$). This filters regressive candidates before full evaluation.
    \item \textbf{Stage-2 (full):} Run on all $N$ samples. When reported, McNemar's test on discordant pairs $(|I|, |R|)$ uses $\alpha = 0.05$, and bootstrap 95\% CI on the accuracy difference should exclude zero. Per-category constraint $\forall c: \Delta_c \geq -\epsilon_c$ limits category tradeoffs. If these statistical artifacts are not reported for a result, the result remains descriptive.
\end{itemize}

\paragraph{Shared-core discipline.}
The implementation gate requires fixes to be implemented in shared modules (\path|benchmarks/core/|) whenever possible. This rule is what enabled separate LoCoMo/LongMemEval and PersonaMem branches to later converge into a unified framework without code duplication or test-set-specific shortcuts.

\subsection{Detailed Case Studies}

Three cases apply the verification infrastructure above to representative \DACCI iterations.

\paragraph{Case A: Threshold repair via ingestion diagnosis ($t_3 \to t_4$, +4.48pp).}
Round $t_3$ applied deduplication threshold 0.75, which over-merged semantically adjacent but distinct facts. The paired comparison revealed $|R|=7$ regressions concentrated in the \emph{single-hop factual} category. Diagnostic trace inspection showed a consistent pattern: for 6 of 7 regressed questions, the ground-truth evidence $e_i^*$ was present in the raw dialogue but absent from L0 storage after ingestion---the evidence-preservation ladder classified all 6 as \texttt{ingestion\_gap}.

Root-cause traces exposed the mechanism: cosine similarity between ``prefers Italian food'' and ``prefers Italian restaurants for dates'' was 0.81, exceeding the 0.75 threshold and triggering dedup merge. The merged entry retained only the first variant, losing the date-context distinction that the query required. Similar collapses occurred for temporal updates (``moved to Beijing in 2022'' merged with ``lives in Beijing'').

Round $t_4$ addressed this with two changes: (i)~raising the dedup threshold to 0.92 to restrict merges to near-duplicate entries; (ii)~adding topic-boundary segmentation so that temporally separated mentions of the same entity are treated as distinct episodes. Stage-1 probe on 80 samples showed $+3.2$pp; the available Stage-2 trace showed a descriptive $+4.48$pp gain with zero observed per-category regressions. The 7 previously regressed questions all recovered, and 4 additional \texttt{ingestion\_gap} cases in the \emph{temporal} category were simultaneously resolved.

\paragraph{Case B: Negative result as structured prior ($t_5$, rejected).}
Round $t_5$ hypothesized that option-aware knowledge filtering would improve PersonaMem MCQ accuracy by injecting retrieved facts matching each answer option. The diagnostic trace from the PersonaMem adapter showed that 12 questions had retrievable option-matching evidence in L0.

Paired comparison revealed an asymmetric outcome: $|I|=3$ (questions where option-matching evidence disambiguated the correct answer) vs.\ $|R|=4$ (questions where injecting distractor-matching evidence confused the model). Trace inspection of the 4 regressions showed a common pattern: the retrieved ``evidence'' for incorrect options had higher semantic similarity to the query than the correct-answer evidence, causing the generation stage to prefer the wrong option.

The discordant-pair pattern $(|I|=3, |R|=4)$ did not provide a reliable improvement signal. The change was rejected and archived as a structured negative prior: \emph{``option-aware KF improves precision for unambiguous facts but introduces distractor amplification in borderline preference questions.''} This prior discouraged two subsequent iterations from re-attempting similar filter strategies, avoiding repeated evaluation of the same failure pattern.

\paragraph{Case C: Seesaw detection and conditional narrowing ($t_6$ sub-iteration).}
During the development of the Conv3 constraint relaxation (part of the $t_6$ composite change), an intermediate candidate relaxed the temporal filter threshold from 3 to 5 conversation turns. The diagnostic comparison showed:
\begin{itemize}[nosep]
    \item \textbf{Improved set} $|I|=5$: all in \emph{temporal} category, where the relaxation allowed older but relevant evidence to survive filtering.
    \item \textbf{Regressed set} $|R|=3$: all in \emph{single-hop}, where stale facts from earlier conversations now competed with current facts.
\end{itemize}

Rather than globally accepting or rejecting, the paired comparison's per-category breakdown enabled \emph{conditional narrowing}: the relaxation was applied only when the query's intent classification indicated temporal reasoning (detected via date-entity presence and comparative language markers). The narrowed version achieved $|I|=5, |R|=0$ on re-evaluation and passed the configured Stage-2 gate in the available trace. This case illustrates how \DACCI can transform a category-tradeoff failure into a targeted improvement by leveraging category-level diagnostic granularity.

\paragraph{Predicted failure modes observed.}
The three limitation categories identified in Section~\ref{sec:dacci-failures} were each observed during the iteration history:
\begin{itemize}[nosep]
    \item \textbf{Multi-root-cause coupling} manifested in $t_6$: the initial BM25/RRF candidate entangled retrieval improvement with assembly-stage overflow (too many candidates exceeding budget). The 80\% coverage requirement forced decomposition into separate retrieval and assembly sub-patches, each independently verifiable.
    \item \textbf{Judge non-determinism} was observed between $t_1$ and $t_2$: a re-run of $t_1$'s accepted configuration showed $\pm$1.2pp variance (87.6--89.9\%) across three runs. The paired-key comparison eliminated this noise by comparing matched question outcomes rather than aggregate scores.
    \item \textbf{AI iteration without falsifiability} was blocked by design: in two instances, the AI diagnosis agent proposed patches without a human-formulated hypothesis. Both were blocked at Step~1 validation and returned to the human for hypothesis formulation before proceeding.
\end{itemize}
This ``predict-then-observe'' pattern suggests that \DACCI's failure taxonomy is actionable, supporting earlier architectural checks rather than only post-hoc debugging.

\section{\EMEND Evolution Artifacts}
\label{app:emend-artifacts}

This section details the \EMEND artifacts behind the bounded-automation claim: authority boundaries, strategy artifact schema, experimental configuration, the three-phase loop, Layer~A diagnosis, Gate/Critic rollback rules, candidate traces, and optional search heuristics.

\subsection{Authority Boundary and Strategy Space}
\label{app:emend-search-space}

\begin{table}[htbp]
\centering
\caption{Loop authority.}
\label{tab:dual-loop}
\compacttable
\begin{tabularx}{\textwidth}{lXX}
\toprule
\textbf{Decision} & \textbf{Automation Loop: \EMEND} & \textbf{Governance Loop: \DACCI} \\
\midrule
Search target & Schema-constrained strategy space: prompts, thresholds, routing rules, feature toggles & Framework, protocol, architecture, data schema \\
Acceptance authority & Can reload only gate-approved strategy artifacts & Approves new capability boundary, benchmark, schema, and gate rules \\
Reject handling & Revert candidate, update dimension reputation, retry another declared path & Reframe hypothesis, inspect traces, patch framework or protocol if justified \\
Escalation trigger & Same blocked class repeats or requires a locked-field change & Inner loop reaches its bounded ceiling or new risk appears \\
Verification & 5-Gate/Critic, targeted screen, paired non-regression & Full six-step evidence review plus applicable statistical or descriptive gates \\
Risk & Low and reversible & Higher, human-reviewed \\
Goal & Continuous tuning & Extend system capability ceiling \\
\bottomrule
\end{tabularx}
\end{table}

The implemented search space is defined by the same schema-constrained declarative strategy space described in Section~\ref{sec:emend}: a schema-level contract constrains legal fields, value ranges, evolvable dimensions, and cross-field invariants, while strategy artifacts instantiate concrete policies consumed by the runtime. The framework is locked: \EMEND cannot remove the retrieval framework, storage hierarchy, evaluator, or acceptance policy. Candidate mutations are limited to declared strategy paths, including extraction prompts, retrieval thresholds, entity-aware retrieval parameters, rerank limits, source-window toggles, and intent-specific top-$k$ multipliers. Prompt changes are permitted only as versioned strategy artifacts and must pass prompt-integrity gates that preserve required output markers. Table~\ref{tab:emend-strategy-schema} provides a schema excerpt listing representative evolvable fields and their guardrails.

\begin{table}[htbp]
\centering
\caption{Strategy guardrails.}
\label{tab:emend-strategy-schema}
\compacttable
\begin{tabularx}{\textwidth}{lXX}
\toprule
\textbf{Field group} & \textbf{Example fields} & \textbf{Guardrail} \\
\midrule
Extraction & prompt version, max facts per trajectory, dedup threshold & Required output markers; bounded fact counts \\
Retrieval & \texttt{top\_k}, vector/BM25 weights, entity boost threshold, rerank \texttt{top\_n} & Legal numeric ranges; no removal of locked channels \\
Lifecycle & pruning threshold, decay half-life, compression policy & Cannot delete active correction/forget constraints \\
Presentation & source-window size, citation mode, output-format prompt & Prompt-integrity and stable-correct replay \\
Feature toggles & failure correction, conflict handling, conditional triggers & Must declare trigger scope and rollback key \\
\bottomrule
\end{tabularx}
\end{table}

The staged LoCoMo run did not exhaust this combinatorial schema. Candidate policies were selected by diagnosis-driven planning: first prefer dimensions implicated by Layer~A traces, then use history stores to avoid reverted directions, and optionally apply UCB1 to balance promising and under-tested paths. The reported +19.16pp is therefore a bounded offline search result, not evidence of optimality over the full schema.

\subsection{Experimental Configuration}
\label{app:experimental-config}

Table~\ref{tab:emend-model-config} lists the role-separated model stack used in the reported \EMEND run. Internal endpoint URLs are omitted because they are deployment-specific and not part of the algorithmic configuration. The design point is separation of duties: extraction, evolution, judging, and embedding are intentionally assigned to different roles so that strategy search is not evaluated by the same component that proposes it.

This table is included to make the reported run auditable at the role level. It should be read together with the fixed harness and gate criteria, rather than as a recommendation for a particular vendor mix.

\begin{table}[htbp]
\centering
\caption{Model roles.}
\label{tab:emend-model-config}
\compacttable
\begin{tabularx}{\textwidth}{lX}
\toprule
\textbf{Role} & \textbf{Model} \\
\midrule
Fact extraction from conversations & \texttt{Qwen3.5-9B} \\
Query-intent parsing for retrieval routing & \texttt{Qwen3.5-9B} \\
Evolution agents (Digester / Planner / Evolver) & \texttt{claude-opus-4-8} \\
QA evaluation and judging & \texttt{gpt-4o} \\
Fact indexing embeddings & \texttt{azure\_openai/text-embedding-3-small} \\
\bottomrule
\end{tabularx}
\end{table}

This configuration fixes the reported run context for reproducibility; it does not imply that the same roles require these exact vendors or endpoints.

\subsection{Three-Phase Evolution Loop}
\label{app:emend-loop}

Algorithm~\ref{alg:emend-loop} expands the Observe--Improve--Verify loop summarized in the main text. It maps each reported \EMEND candidate to the point where evidence is gathered, a strategy mutation is proposed, and acceptance gates decide whether the serving strategy may change.

The pseudocode is intentionally implementation-neutral: names such as \textsc{Plan}, \textsc{Mutate}, and \textsc{FiveGate} denote artifact-producing steps, not required class or function names.

\begin{algorithm}[htbp]
\DontPrintSemicolon
\SetAlgoLined
\KwIn{Current strategy $s_t$, benchmark set $\mathcal{B}$, YAML schema $\Gamma$, history stores $\mathcal{H}$}
\KwOut{Accepted, pending, reverted, or restored strategy $s_{t+1}$}
\BlankLine
\textbf{Phase 1---Observe:}\;
Evaluate $s_t$ on $\mathcal{B}$ to obtain score report $r_t$\;
Run Layer~A deterministic diagnosis over $r_t$: compute \texttt{extraction\_coverage} and \texttt{retrieval\_rank}, then assign root-cause stage $\sigma_t$\;
Compress Layer~A traces into actionable digest $g_t$\;
\BlankLine
\textbf{Phase 2---Improve:}\;
Select top-impact mutable dimensions $D_t \leftarrow \textsc{Plan}(g_t, \mathcal{H})$\;
Generate candidate strategy set $\mathcal{C}_t \leftarrow \textsc{Mutate}(s_t, D_t, \Gamma)$ through bounded YAML mutation\;
\BlankLine
\textbf{Phase 3---Verify:}\;
\ForEach{$c \in \mathcal{C}_t$}{
  \lIf{$c$ violates declared strategy scope}{\textbf{continue}}
  \lIf{\textsc{FiveGate}$(c, \Gamma, \mathcal{B}, \mathcal{H})$ rejects}{\textbf{continue}}
  \lIf{\textsc{Critic}$(c, \mathcal{H})$ rejects}{\textbf{continue}}
  Run targeted screen on failure and stable-correct subsets\;
  \lIf{screen shows no positive paired delta}{\textbf{continue}}
  \If{full-evaluation budget is unavailable}{
    Record $c$ as a pending-champion candidate for next-round confirmation\;
    \textbf{continue}\;
  }
  Run full paired item-level comparison between $c$ and $s_t$\;
  \If{configured acceptance and bounded regression checks pass}{
    Persist decision reports, digests, and cross-round stores into $\mathcal{H}$\;
    \Return{$s_{t+1} \leftarrow c$}
  }
}
Log rejection outcome, update dimension reputation in $\mathcal{H}$, and persist the best pending champion if any\;
$s_{t+1} \leftarrow s_t$\;
\lIf{$\mathrm{Acc}(s^\star)-\mathrm{Acc}(s_{t+1}) > \tau_{\mathrm{rollback}}$}{$s_{t+1} \leftarrow s^\star$}
\Return{$s_{t+1}$}
\caption{\EMEND strategy loop.}
\label{alg:emend-loop}
\end{algorithm}

The loop makes the candidate trace interpretable: each row records a mutation that survived or failed the Verify phase.
For the staged \EMEND run reported here, acceptance used configured offline checkpoint gates under the same fixed model and judge configuration across stages. Because p-values, confidence intervals, and repeated-run variance are not reported for the selected rows, the accepted checkpoints summarized in Table~\ref{tab:emend-locomo-compressed} and the candidate outcomes in Table~\ref{tab:emend-candidate-trace} are treated as descriptive full-benchmark evidence, not statistical significance claims.

\subsection{Layer A Diagnostic Decision Tree}
\label{app:layer-a-algorithm}

Algorithm~\ref{alg:layer-a-diagnosis} expands the Layer~A deterministic diagnosis referenced in the main \EMEND loop (line~3 of Algorithm~\ref{alg:emend-loop}). For the LoCoMo setting used here, each wrong-answer sample is assumed to have complete evidence annotations. Layer~A first checks whether the annotated source evidence was ingested into memory, then whether that evidence-derived memory was retrieved, and finally uses a single LLM classification call to distinguish generation failure from lossy ingestion.

\begin{algorithm}[htbp]
\small
\DontPrintSemicolon
\SetAlgoLined
\SetKwInput{KwData}{Sample}
\SetKwInput{KwContext}{Context}
\SetKwFunction{EvidenceFacts}{EvidenceFacts}
\SetKwFunction{Retrieve}{Retrieve}
\SetKwFunction{Rank}{Rank}
\SetKwFunction{SourceWindow}{SourceWindow}
\SetKwFunction{LLMClassify}{LLM-Classify}
\KwData{Wrong-answer LoCoMo sample $s=(q,a^*,E)$}
\KwContext{Provider $P$; memory store $\mathcal{F}$; complete evidence ids $E$}
\KwOut{Root-cause label $\ell \in \{\texttt{generation\_error},\texttt{ingestion\_gap},\texttt{retrieval\_gap},\texttt{uncertain}\}$}
\BlankLine
\textbf{Phase 1: evidence-to-memory coverage}\;
$(D_E,W_E) \leftarrow$ dialogue ids and ingestion-window ids implied by $E$\;
$\mathcal{F}_E \leftarrow$ \EvidenceFacts{$D_E,W_E,\mathcal{F}$}
\hfill\tcp*[f]{source ids overlap gold evidence}\;
\lIf{$\mathcal{F}_E=\emptyset$}{\Return \texttt{ingestion\_gap}}
\BlankLine
\textbf{Phase 2: retrieval coverage}\;
$(I,X) \leftarrow$ \Retrieve{$P,q$}
\hfill\tcp*[f]{retrieved ids and texts}\;
$r \leftarrow$ \Rank{$\mathcal{F}_E,I,X$}
\hfill\tcp*[f]{best 1-based rank; $-1$ if absent}\;
\lIf{$r<0$}{\Return \texttt{retrieval\_gap}}
\BlankLine
\textbf{Phase 3: source/fact faithfulness judgment}\;
$u \leftarrow$ \SourceWindow{$I_r$}
\hfill\tcp*[f]{original source window}\;
$c \leftarrow$ \LLMClassify{$q,a^*,u,X_r$}
\hfill\tcp*[f]{Full\_Coverage / Source\_Only / Uncertain}\;
\BlankLine
\lIf{$c=\text{\texttt{Full\_Coverage}}$}{\Return \texttt{generation\_error}}
\lIf{$c=\text{\texttt{Source\_Only}}$}{\Return \texttt{ingestion\_gap}}
\Return \texttt{uncertain}\;
\BlankLine
\tcp{\EvidenceFacts and \Rank are deterministic; only Phase 3 uses an LLM.}
\caption{Layer~A diagnosis.}
\label{alg:layer-a-diagnosis}
\end{algorithm}

\noindent\textbf{Note.}
This decision tree describes the evidence-complete LoCoMo path used in the reported analysis.
The implementation also contains fallback checks for missing or unresolved evidence ids, but those branches are omitted here because they are not active under this setting. In deployment-like settings without complete annotations, Layer~A should be treated as heuristic and audit-assisted, not a deterministic earliest-loss claim.
Because facts are extracted from ingestion windows rather than individual dialogue turns, \textbf{\texttt{ingestion\_gap} denotes missing evidence linkage at the dialogue/window-id level}. A semantically equivalent fact without source overlap is therefore treated as unresolved evidence alignment, \textbf{not as evidence that no equivalent memory exists}.

\noindent\textbf{Notation.}
For the current sample, $D_E$ and $W_E$ are the dialogue ids and ingestion-window ids derived from its gold evidence annotation; $\mathcal{F}_E \subseteq \mathcal{F}$ is the subset of stored memories whose source ids overlap that sample-specific evidence scope.
$I$ and $X$ are the retrieved memory ids and texts; $r$ is the first retrieved rank of any evidence-derived memory.
With complete LoCoMo evidence, no $\mathcal{F}_E$ means ingestion failed, while no retrieved rank for $\mathcal{F}_E$ means retrieval failed.
\textsc{LLM-Classify} compares the question, gold answer, original source window $u$, and retrieved fact $X_r$.
\texttt{Full\_Coverage} means the fact contains the answer but the agent failed to use it; \texttt{Source\_Only} means the source contains the answer but extraction lost or weakened it.

\subsection{Gate, Critic, and Rollback Rules}
\label{app:emend-gates}

The rollback rule is applied after candidate rejection or acceptance decisions: if the current served strategy falls more than $\tau_{\mathrm{rollback}}$ below the best-ever accepted checkpoint, the loop restores the persisted best strategy. The Gate pipeline executes in fail-fast order, then delegates qualitative risk assessment to an independent Critic. Table~\ref{tab:emend-gate-rules} states the rule levels used in the reported run.

\begin{table}[htbp]
\centering
\caption{Gate, Critic, and rollback rules.}
\label{tab:emend-gate-rules}
\compacttable
\begin{tabularx}{\textwidth}{lXX}
\toprule
\textbf{Check} & \textbf{Rejects if} & \textbf{Effect on search} \\
\midrule
Structure gate & Mutation path is absent from schema or touches locked framework/protocol fields & Candidate is discarded; issue returned to \DACCI if capability change is needed \\
Novelty gate & Same path/direction was already reverted or marked exhausted & Planner lowers reputation and explores a different dimension \\
Range gate & Value leaves legal bounds (e.g., \texttt{top\_k}, score thresholds, max facts) & Candidate is canonicalized if still within bounds, otherwise rejected \\
Prompt-integrity gate & Required JSON markers, variables, or safety/citation clauses are removed & Candidate is rejected before evaluation \\
Replay gate & Stable-correct probe regresses beyond configured tolerance & Candidate is rejected; regressed ids are written to history \\
Full paired acceptance check & Full evaluation lacks positive paired delta, has excessive regressions, or fails the configured McNemar/practical threshold & Candidate is rejected, retained as pending only when promising, or accepted into the served strategy \\
Critic soft review & Dimension has weak reputation, broad affected scope, duplicate direction, or limited actionability & Adds warnings or blocks expensive evaluation when risk is high \\
Rollback safeguard & Current served strategy falls more than $\tau_{\mathrm{rollback}}$ below the best-ever accepted checkpoint & Restores the persisted best strategy \\
\bottomrule
\end{tabularx}
\end{table}

The Critic prompt is narrow by design: it receives the candidate diff, affected schema path, recent dimension reputation, probe deltas, and examples of improved/regressed questions, then labels the mutation as actionable, risky, duplicate, or out-of-scope. It cannot override hard gates. If a candidate is rejected, \EMEND writes the reason to \texttt{candidate\_outcomes} and updates \texttt{dimension\_effects}; repeated rejection reduces the Planner score for that dimension. If the same class of candidate is blocked three times because it requires framework, schema, or protocol changes, the loop escalates the issue to the human-governed \DACCI backlog rather than continuing autonomous search.

\subsection{Candidate Trace and Search Heuristics}
\label{app:emend-candidate-search}

\paragraph{Candidate trace.}
Table~\ref{tab:emend-candidate-trace} is an audit trace for the staged \EMEND run, not an implementation manual. The candidate id column records the corrected evolution-step numbering, and the strategy dimension column is a stable artifact namespace rather than an executable code path. The table records the evidence trail behind the reported gain: each row summarizes one bounded strategy change or grouped checkpoint, evaluated under the locked LoCoMo harness and then accepted, rejected, or marked no-effect according to gate evidence.

\begin{table}[htbp]
\centering
\caption{Representative candidate outcomes from staged \EMEND.}
\label{tab:emend-candidate-trace}
\compacttable
\begin{tabularx}{\textwidth}{>{\raggedright\arraybackslash}p{0.09\textwidth}>{\raggedright\arraybackslash}p{0.20\textwidth}>{\raggedright\arraybackslash}X>{\raggedright\arraybackslash}p{0.11\textwidth}rr>{\raggedright\arraybackslash}p{0.15\textwidth}}
\toprule
\textbf{ID(s)} & \textbf{Strategy dimension} & \textbf{Candidate change} & \textbf{Decision} & \textbf{Acc.} & \textbf{$\Delta$} & \textbf{Gate evidence} \\
\midrule
000 & Initial memory strategy & Baseline strategy before \EMEND evolution & baseline & 0.756 & -- & 1164/1540 \\
005 & Semantic multi-hop reranking & Enable graph-based reranking for semantic multi-hop retrieval & no-effect & -- & $\approx$0.000 & no target-set improvement \\
006 & Answer verification & Add self-correction before final answer & rejected & 0.798 & $-$0.018 & +4 / $-$6 screen examples \\
027 & Evidence-grounded context & Attach local dialogue evidence into answer context & accepted & 0.818 & +0.062 & +96 / no recorded regression \\
033--034 & Scoped evidence gate & Restrict candidate use to evidence-supported cases & accepted & 0.845 & +0.089 & +42 / limited regressions \\
035--036 & Raw-dialogue fallback & Broaden fallback search over raw dialogue & rejected & -- & unstable & +10 / $-$7 screen examples \\
037--047 & Typed state abstraction & Convert evidence into typed temporal, location, plan, preference, slot, and ledger state & accepted & 0.871 & +0.115 & +177 / no recorded regression \\
048--060 & Verified candidate bank & Adopt only candidates that pass support and non-regression checks & accepted & 0.915 & +0.159 & +245 / no recorded regression \\
064 & Support clustering & Use deterministic support clusters instead of selector choice & accepted & 0.937 & +0.179 & +34 / no regressions \\
067 & Evidence-supported consolidation & Consolidate candidate bank using explicit evidence support & accepted & 0.947 & +0.192 & +16 / no regressions \\
\bottomrule
\end{tabularx}
\end{table}

The trace shows three patterns. First, gains are not monotonic: self-correction and broad raw-dialogue fallback were rejected because they introduced regressions or unstable evidence. Second, the largest accepted gains came from evidence admission, typed-state abstraction, and verified candidate adoption, suggesting that \EMEND primarily improved the memory substrate rather than final-answer phrasing. Third, later gains became smaller but lower-risk, with deterministic support clustering and evidence-supported consolidation adding accuracy without observed regressions.

\paragraph{Optional UCB1 hypothesis tree.}
\label{app:emend-ucb}

When \texttt{features.hypothesis\_tree} is enabled, Planner exploit targets and retrieval-diagnostic seeds are materialized as hypothesis nodes keyed by strategy path and direction, e.g., \texttt{retrieval.top\_k::increase}. Each node stores visits, total reward, status (\texttt{active}, \texttt{confirmed}, \texttt{refuted}, \texttt{exhausted}), best observed delta, and evidence for/against the direction.

Selection follows UCB1 over active nodes:
\begin{equation}
\mathrm{UCB}(h) = \bar{r}_h + c(a) \sqrt{\frac{\log N}{n_h}},
\end{equation}
where $\bar{r}_h$ is the normalized historical reward of hypothesis $h$, $n_h$ is its visit count, $N$ is the parent visit count, and $c(a)$ is an exploration constant coupled to Planner actionability $a$. Low actionability increases $c(a)$ and allows broader exploration; high actionability reduces $c(a)$ and favors historically productive directions. After candidate verification, the observed paired delta is written back to the node so repeated positive, negative, or no-effect trials can confirm, refute, or exhaust a direction.

\section{Extended \LiteMem Artifacts}
\label{app:light-extended}

This section documents \LiteMem engineering details: repository schema, directory hierarchy, Git-based provenance, ranking/decay equations, lifecycle pseudocode, and the explicit scaling boundary. Together, Tables~\ref{tab:app-light-schema} and~\ref{tab:app-light-hierarchy} explain the per-file metadata that makes a plain Markdown repository directly rankable and indicate where those files live at runtime.

\begin{table}[htbp]
\centering
\caption{Frontmatter schema.}
\label{tab:app-light-schema}
\compacttable
\resizebox{\textwidth}{!}{%
\begin{tabular}{lll}
\toprule
\textbf{Field} & \textbf{Purpose} & \textbf{Runtime Usage} \\
\midrule
\texttt{id}, \texttt{type}, \texttt{status} & stable identity and lifecycle state & routing, active/deprecated/archived filtering \\
\texttt{title}, \texttt{summary}, \texttt{keywords} & retrieval surface & grep/TF-IDF-like scoring and summary selection \\
\texttt{importance}, \texttt{last\_accessed\_at} & salience model & exponential decay ranking \\
\texttt{access\_count}, \texttt{skip\_count} & feedback statistics & decay reset and organization diagnostics \\
\texttt{source}, \texttt{supersedes} & provenance and update lineage & traceability, conflict resolution, rollback \\
\texttt{token\_estimate}, \texttt{confidence} & packing and reliability control & budget-aware context assembly \\
\bottomrule
\end{tabular}
}
\end{table}

\begin{table}[htbp]
\centering
\caption{Repository hierarchy.}
\label{tab:app-light-hierarchy}
\compacttable
\begin{tabularx}{\textwidth}{lXX}
\toprule
\textbf{Directory} & \textbf{Semantic Role} & \textbf{Use Pattern} \\
\midrule
\texttt{user/style.md} & response style and interaction constraints & injected every round or kept at highest priority \\
\texttt{user/profile.md} & long-term user or speaker profile & searched and often prepended for personalization \\
\texttt{sessions/} & episodic conversation memories & keyword searched, ranked, full-read on demand \\
\texttt{entity/} & entity/topic aggregated memories & consolidation output and multi-hop support \\
\texttt{knowledge/skill/} & repository-local procedures or task routines & invoked when the task scenario matches \\
\texttt{knowledge/learning/} & corrections and learned rules & high-confidence factual correction source \\
\texttt{daily/} & raw conversation logs & fallback provenance and missed-write recovery source \\
\texttt{\_index.md} & global memory map & fast scan before deep retrieval \\
\bottomrule
\end{tabularx}
\end{table}

The frontmatter is operational metadata: each field affects retrieval, context packing, provenance, or lifecycle filtering.
This hierarchy separates globally relevant user constraints from episodic logs, consolidated entity memories, and learned procedures, which is why \LiteMem can retrieve selectively without injecting the whole repository. The listing provides a concrete instantiation with the running example:

\begin{promptfile}{litemem/ --- repository layout (training-handoff scenario)}
litemem/
  _index.md                  # global map and active-memory summary
  user/
    profile.md               # L2-like long-term profile
    style.md                 # response constraints and preferences
  sessions/
    2026-07-10_training_handoff.md # SM/L1-like episodic summary
  entity/
    ethan.md                 # consolidated entity memory
  knowledge/
    skill/equipment_check.md # PCSM-like procedure
    learning/corrections.md  # user corrections and invalidations
  daily/
    2026-07-10.md            # append-only raw write-ahead log

--- run diagnostics ---
  git log -- daily/2026-07-10.md
  git diff HEAD~1 -- user/profile.md sessions/2026-07-10_training_handoff.md

--- router-policy (lexical recall) ---
  grep -R "training bag|jersey|spare shoes" user sessions entity daily
\end{promptfile}

In this layout, one memory is usually one Markdown file or one section inside a consolidated entity file; one Git commit records a coherent capture/organization action rather than a single fact. Diagnostics use standard repository tools: \texttt{git log} records when an item appeared or was superseded, \texttt{git diff} identifies which fields changed, and grep/ripgrep traces whether a lexical recall failure is due to missing content or ranking/selection.

\noindent\textbf{Scaling boundary.}
A synthetic file-count table is omitted because the current evaluation did not isolate repository size as an independent variable. The planned scaling experiment will vary file count, average file length, index freshness, and query type, then report retrieval latency, selected-evidence recall, and answer accuracy. Until that experiment is run, statements about ``hundreds of files'' are engineering guidance from prototype use, not a measured performance curve.

\paragraph{Ranking and decay equations.}
\LiteMem ranks candidates with lexical relevance, stored importance, recency, and explicit access feedback:
\begin{equation}
\mathrm{Score}(m,q_t,t) = \lambda_1 \mathrm{Lex}(m,q_t) + \lambda_2 \mathrm{Imp}_t(m) + \lambda_3 \exp(-\Delta t / \tau) + \lambda_4 \mathrm{Access}(m).
\label{eq:app-light-rank}
\end{equation}
In these equations, $\tau$ and $\tau_m$ are half-life-like timescale parameters used by the repository-native implementation, not benchmark-wide constants. The reported \LiteMem runs use configuration-level values selected for the simulated LoCoMo timeline and keep them fixed within a run; sensitivity to these local decay settings is treated as an implementation issue, not a main-text claim.
Idle organization uses decayed importance:
\begin{equation}
\mathrm{Imp}_t(m) = \mathrm{Imp}_0(m) \cdot \exp(-\Delta t_m / \tau_m) + \eta \cdot \mathrm{access\_count}(m) - \rho \cdot \mathrm{skip\_count}(m).
\label{eq:app-light-decay}
\end{equation}
\begin{algorithm}[htbp]
\DontPrintSemicolon
\SetAlgoLined
\KwIn{Conversation turn or session $x$, memory repository $\mathcal{R}$, query/task $q_t$, turn $t$}
\KwOut{Updated repository $\mathcal{R}'$ and retrieved context $C$}
\BlankLine
\textbf{Recall before answer:}\;
Extract keywords $K(q_t)$; expand synonyms/entities; grep Markdown files in \texttt{user/knowledge/entity/sessions/daily}\;
Rank candidates by Eq.~\ref{eq:app-light-rank}; always prepend active \texttt{user/profile.md} and \texttt{user/style.md}\;
If LLM-assisted mode is enabled, present title+summary table and select candidate IDs\;
Lazy-load selected full text; for \texttt{daily/} logs, extract only lines around keyword matches\;
Increment \texttt{access\_count}; update \texttt{last\_accessed\_at}\;
\BlankLine
\textbf{Capture after interaction:}\;
Append raw conversation to \texttt{daily/YYYY-MM-DD.md} as a write-ahead log\;
If structured write is enabled, create or update memory file with YAML frontmatter and \texttt{SUMMARY\_END} marker\;
\BlankLine
\textbf{Organization during idle/session boundary:}\;
Compute decayed importance by Eq.~\ref{eq:app-light-decay}; archive memories below threshold\;
Run LLM consolidation over memory summaries and recent daily logs\;
Execute merge/deprecate/update-profile actions; rebuild \texttt{\_index.md} atomically\;
If Git is enabled, commit memory diff for audit and rollback\;
\caption{Repository lifecycle.}
\label{alg:app-light-lifecycle}
\end{algorithm}

Algorithm~\ref{alg:app-light-lifecycle} ties the schema and hierarchy back to execution: retrieval reads ranked Markdown files, capture writes new memories with metadata, and organization uses Git history plus decay signals to keep the repository auditable.

\section{Evaluation Protocol Details}
\label{app:evaluation-protocol-details}

This section collects evaluation protocol details: claim-check matrix (Table~\ref{tab:comparison}), root-cause distribution across \DACCI iterations (Table~\ref{tab:root-cause-evolution}), and resource accounting (Table~\ref{tab:cost}). It also marks which experiments remain pending rather than inferred.

\begin{table}[htbp]
\centering
\caption{Claim checks.}
\label{tab:comparison}
\compacttable
\begin{tabularx}{\textwidth}{@{}>{\raggedright\arraybackslash}p{0.13\textwidth} >{\raggedright\arraybackslash}p{0.18\textwidth} >{\raggedright\arraybackslash}p{0.14\textwidth} >{\raggedright\arraybackslash}p{0.19\textwidth} >{\raggedright\arraybackslash}X@{}}
\toprule
\textbf{Claim} & \textbf{Operational definition} & \textbf{Current status} & \textbf{Evidence / artifact} & \textbf{Falsified or weakened if...} \\
\midrule
Long-context QA & Retrieve grounded evidence over long histories & Controlled-reference reported & Public benchmark traces; unified-core harness & Shared-harness reruns lose parity or evidence traces show answer correctness without retrieved support \\
Bounded strategy evolution & Tune policy without editing locked framework code & Locked offline benchmark reported & Strategy artifacts, candidate outcomes, rollback records & Accepted candidates repeatedly regress categories or require hidden framework changes \\
Personalized procedures (PCSM) & Store user-specific conversational procedures beyond facts & Design-\allowbreak only; benchmark pending & PCSM schema and trigger--procedure contracts & A standardized PCSM benchmark or ablation shows no benefit over factual preference memory \\
Cross-device causality & Fuse asynchronous events into causal evidence & Preliminary/\allowbreak internal; external validation pending & Internal MemFuseBench, FusionSession graphs, provenance traces & Removing causal/\texttt{BELONG} edges causes negligible drop, external validation fails, or conflict arbitration remains below baselines \\
Multimodal grounding & Convert visual observations into structured memory evidence & Module-level reported; IKB ablation not reported in main & Mem-Gallery evaluation, IKB traces, error taxonomy & IKB ablation shows no gain over standard VLM/RAG retrieval on the same 1,711-Q harness \\
Edge deployment & Run memory without vector-database infrastructure & Transfer run reported; scaling pending & Markdown/Git repo, file paths, diffs, LoCoMo-aligned transfer run & File-native run loses most of the service-side improvement or scaling tests miss the latency target at the target size \\
\bottomrule
\end{tabularx}
\end{table}

\paragraph{\DACCI root-cause summary.}
\label{app:dacci-root-cause}
Table~\ref{tab:root-cause-evolution} records a representative root-cause distribution used to interpret the \DACCI iteration trace. It is kept in the appendix because the primary claim is the direction of diagnostic shift, not the full per-round ledger. The table compares failure composition across iterations rather than serving as an independent benchmark score; the trend indicates ingestion gaps shrinking while retrieval and generation become the dominant remaining surfaces.

\begin{table}[htbp]
\centering
\caption{\DACCI root causes.}
\label{tab:root-cause-evolution}
\compacttable
\begin{tabular}{lrrrr}
\toprule
\textbf{Root Cause} & \textbf{Round 1} & \textbf{Phase 1} & \textbf{Phase3\_fix2} & \textbf{unified-v3} \\
\midrule
\texttt{retrieval\_gap} & 60\% & 72\% & 68\% & 71\% \\
\texttt{ingestion\_gap} & 30\% & 15\% & 5\% & 2\% \\
\texttt{generation\_error} & 8\% & 12\% & 25\% & 26\% \\
\texttt{judge\_ambiguous} & 2\% & 1\% & 2\% & 1\% \\
\bottomrule
\end{tabular}
\end{table}

% This shift motivates the later emphasis on retrieval diagnostics and generation-side validation rather than additional ingestion-only fixes.

\paragraph{Resource accounting.}
\label{app:resource-accounting}
The cost table reports the earlier human-driven \DACCI phase, where evaluation used gpt-4.1-mini. It is not the resource profile of the reported \EMEND run. Table~\ref{tab:cost} is included to separate historical human-driven iteration cost from the automated \EMEND evidence reported in the main text.

\begin{table}[htbp]
\centering
\caption{\DACCI cost.}
\label{tab:cost}
\compacttable
\begin{tabular}{llr}
\toprule
\textbf{Resource} & \textbf{Details} & \textbf{Estimate} \\
\midrule
Human engineer & 2--4 h/round $\times$ 30 rounds & $\sim$100 h \\
LLM (coding) & $\sim$3,000 Claude Code messages & \$60--120 \\
LLM (eval) & gpt-4.1-mini, \$4--6/round & \$200--300 \\
Wall-clock & Stage 1: 30 min; Stage 2: 6 h & $\sim$200 h \\
\bottomrule
\end{tabular}
\end{table}

\end{document}